\documentclass[twoside,11pt]{article}

\usepackage{jmlr2e}
\usepackage[symbol]{footmisc}
\usepackage{amsmath,bm}
\usepackage{array}
\usepackage{multirow}
\usepackage[title,toc,titletoc,page]{appendix}
\usepackage[noend]{algpseudocode}
\usepackage{graphicx}
\usepackage[linesnumbered,ruled]{algorithm2e}
\SetKwInput{KwInput}{Input} 
\SetKwInput{KwOutput}{Output}  
\usepackage{graphicx}
\graphicspath{ {./images/} }
\usepackage{float}

\pdfoutput=1

\ShortHeadings{Constructing Prediction Intervals with Neural Networks}{Contarino et al.}
\firstpageno{1}

\begin{document}

\title{Constructing Prediction Intervals with Neural Networks: \\ An Empirical Evaluation of Bootstrapping and Conformal Inference Methods}

\author{\name Alex Contarino
\email alexander.contarino@afacademy.af.edu \\
    \addr Department of Mathematical Sciences \\ 
    United States Air Force Academy\\
       \\
\name Christine Schubert Kabban
\email christine.schubertkabban@afit.edu \\
\name Chancellor Johnstone
\email chancellor.johnstone@afit.edu \\
       \addr Department of Mathematics and Statistics\\
       Air Force Institute of Technology\\
       Wright-Patterson AFB, OH 45433, USA\\
\\
\name Fairul Mohd-Zaid 
\email fairul.mohd-zaid@us.af.mil \\
\addr 
711th Human Performance Wing \\
Machine Learning Section, Mission Analytics\\Wright-Patterson AFB, OH 45433, USA}

\editor{}

\maketitle

\begin{abstract}
Artificial neural networks (ANNs) are popular tools for accomplishing many machine learning tasks, including predicting continuous outcomes. However, the general lack of confidence measures provided with ANN predictions limit their applicability. Supplementing point predictions with prediction intervals (PIs) is common for other learning algorithms, but the complex structure and training of ANNs renders constructing PIs difficult. This work provides the network design choices and inferential methods for creating better performing PIs with ANNs. A two-step experiment is executed across 11 data sets, including an imaged-based data set. Two distribution-free methods for constructing PIs, bootstrapping and conformal inference, are considered. The results of the first experimental step reveal that the choices inherent to building an ANN affect PI performance. Guidance is provided for optimizing PI performance with respect to each network feature and PI method. In the second step, 20 algorithms for constructing PIs—each using the principles of bootstrapping or conformal inference—are implemented to determine which provides the best performance while maintaining reasonable computational burden. In general, this trade-off is optimized when implementing the cross-conformal method, which maintained interval coverage and efficiency with decreased computational burden.  
\end{abstract}


\footnotetext{The views expressed in this document are those of the authors and should not be interpreted as representing the official policy or position of the U.S. Air Force, the U.S. Department of Defense, or the U.S. Government.}

\section{Introduction}
 
Neural networks, specifically multi-layer perceptrons (MLPs) and convolutional neural networks (CNNs), are popular tools used for many machine learning tasks. Comprised of connected nodes, neural networks aim to create a flexible means to mathematically learn representations of data. Within each node, vector-to-scalar and activation operations are performed. Each activation function $\phi(\cdot)$, e.g., the sigmoid (“logistic”) function, serves to map a scalar value to a desired range. The collection of nodes acting in parallel represent an automated method for mapping a set of features \bm{$X$} to a target outcome \bm{$y$}~\citep{chollet:2017}. Neural networks have proved useful for a variety of tasks; one specific setting is natural language processing where manually engineering a mapping between \bm{$X$} and \bm{$y$} is difficult \citep{goodfellow:2013}. Other regression-based tasks in which neural networks have provided state-of-the-art performance include age prediction \citep{rothe-et-al:2018}, orientation estimation \citep{berg-et-al:2020} and microscopy image processing \citep{xie-et-al:2018}.

MLPs and CNNs are “feed-forward” neural networks, with nodes  arranged hierarchically in successive hidden layers \citep{goodfellow:2013}. The key difference between MLPs and CNNs is the choice of the vector-to-scalar operator $\Sigma(\cdot)$, with the former using vector multiplication. In contrast, CNNs are so-called because a convolution operation is used in the nodes of one or more hidden layers. Convolution effectively induces sparse connectivity among the nodes (generally called “filters” or “feature maps”), with favorable results when \bm{$X$} has high dimensionality, e.g., a set of images. 

In regression, \bm{$y$} is continuous-valued variable, which can pose challenges when using neural networks for inference. For instance, suppose that an MLP is trained on \bm{$X$} and \bm{$y$} under mean squared error (MSE) loss, and a new instance of \bm{$X$}, $\bm{x}_{test}$, is observed. Suppose further that we wish to predict the value of the associated, unobserved target value, $y_{test}$. The trained network will provide a prediction for $y_{test}$ given $\bm{x}_{test}$; denote this value as $\hat{f}(\bm{x}_{test})$. However, with \bm{$y$} being real-valued, the probability that $\hat{f}(\bm{x}_{test})=y_{test}$ is zero. Thus, associating $\hat{f}(\bm{x}_{test})$ with some measure of uncertainty is useful for enhancing the usability of a trained network. A prediction interval (PI) is one such measure, placing probabilistic bounds on the value of $y_{test}$ ~\citep{casella:2002}. A $1-\alpha$ PI is a set of target values $\Gamma \equiv \Gamma(\bm{x}_{test})$ such that
\\
\begin{equation} \label{PI_def}
\text{P}\left(y_{test} \in \Gamma \right)=1-\alpha.
\end{equation}

Prediction intervals must account for two sources of error inherent to training a regression
model, including model-fitting error (or “prediction variance”) and irreducible error (or “data noise”) \citep{gareth:2013}. The performance of a prediction interval is generally evaluated according to two metrics: validity and efficiency \citep{fraser-guttman:1956}. Validity is a function of the coverage of the PI, or the rate at which the PI correctly captures the estimated target value, and whether or not this rate matches the nominal confidence level $1-\alpha$. The second metric of performance is efficiency which measures the width of the PI. Narrow prediction intervals give more information about $y_{test}$ and are preferred, though not at the expense of coverage. Thus, the optimal PI for an estimated target $y_{test}$ is one having the smallest average width while still satisfying (\ref{PI_def}). 

Due to their complex structures and iterative training process, constructing PIs for neural networks is generally a difficult task. Factors such as performance and computational time must be considered. The design choices inherent to constructing a neural network, such as the number of layers and nodes, have sizeable impacts on the network’s predictive ability, yet little is known regarding how such factors impact PI performance. Moreover, each method for constructing a PI has its own advantages and disadvantages. For instance, parametric approaches, such as maximum likelihood and approximate Bayes, estimate uncertainty by computing covariances among the network parameters \citep{papadopoulos-et-al:2001}. To do so, gradient matrices are computed at each training iteration, both complicating the optimization of the network and introducing a large computational burden \citep{khosravi-et-al:2015}. 

In contrast to the uncertainty quantification methods mentioned above, methods leveraging bootstrap resampling are more easily implemented and reduce dependency on assumptions related to the distributions of parameter estimates. In a bootstrap method, an ensemble of $B$ neural networks is trained, each to a different bootstrap resample of the data tuple $(\bm{X}, \bm{y})$. For a given instance \bm{$x_i$} denote $\hat{f}_b(\bm{x_i})$ as the predicted value of the target $y_i$ from the network trained on the $b$\textsuperscript{th} bootstrap resample. The empirical distribution of values $\hat{f}_b(\bm{x_i})$, $b=1,…,B$, is then used to calculate the statistics needed to construct a PI \citep{efron:1993}. As a flexible approach, both in assumptions and practical implementation, bootstrapping has been used extensively with neural networks (\citealt{papadopoulos-et-al:2001}; \citealt{khosravi-et-al:2015}; \citealt{anirudh:2017}). However, the time and resources required to train and store modern MLPs and CNNs make bootstrapping methods increasingly difficult to implement. Moreover, to our knowledge, no experiment has examined how PI performance changes as neural network hyperparameters are tuned. 

Conformal inference methods offer an intriguing alternative to bootstrapping. These approaches have many of the same advantages of bootstrapping but, for the most part, are less computationally expensive. Conformal inference determines whether a candidate value $q$ should belong in $\Gamma$ by evaluating how well the data pair $(\bm{x}_{test},q)$ conforms to the prior knowledge derived from \bm{$X$} and \bm{$y$} \citep{gammerman-et-al:1998}. While there has been research evaluating the efficacy of conformal inference approaches in deep learning settings (\citealt{papadopoulos:2008}; \citealt{angelopoulos-et-al:2020}), these experiments have not, for example, compared the various conformal inference methods on differing data sets. 

\subsection{Contribution and Document Overview}

The contribution of our research is two-fold. First, we show how the choice of hyperparameter settings in a MLP or CNN can affect the performance of PIs in terms of validity and efficiency using bootstrap and conformal inference methods. We specifically design MLPs and CNNs with varying hyperparameter settings, with each network trained to a common training set. PIs are then constructed for responses associated with pre-sequestered test observations. Differences in PI performance among the different hyperparameter settings are then identified using an analysis of variance (ANOVA) approach. These results are used to quantify the sensitivities of the bootstrap and conformal inference PI methods to various hyperparameter settings in a neural network. 

The second contribution is a thorough evaluation of bootstrap and conformal inference methods to determine which provide the best performing PIs, while maintaining a reasonable computational burden. Within this analysis, we also conduct a cursory analysis of conditional coverage. We use the findings of these analyses to provide guidance regarding the optimal PI method given varying time and resource constraints. Experiments are executed across eleven commonly-used data sets for regression tasks, including the image-based Rotated MNIST (“RotNIST”) data set.

The remainder of this document is organized as follows. Section 2 provides comprehensive overview of the methods examined in this research. Section 3 describes the two-step experiment conducted in order to evaluate PI method and performance. In Section 4, the empirical results of the two-step experiment are presented and analyzed. Section 5 concludes the paper with key findings. 

\section{Description of Methods}

The following section describes the prediction interval methods examined in this research. These approaches fall into two classes: bootstrapping and conformal inference. These methods are discussed in terms of their theoretical underpinnings, practical implementation, as well as their advantages and shortcomings in various settings. 

\subsection{Bootstrapping}

Bootstrapping is a resampling-based method for inference, used in computing measures of accuracy for statistics, such as bias and variance \citep{gentle:2009}. Specifically, suppose a random sample of size $n$, $\bm{z}=\{z_1,z_2,…,z_n\}$, is drawn from an unknown distribution and it is desired to use \bm{$z$} to calculate some statistic of interest, $T=g(\bm{z})$. Define {$\hat{F}$} as the empirical distribution of the observed data, where each $z_i$ in \bm{$z$} occurs with probability $1/n$. Then, a “bootstrap resample” is a random sample of size $n$ drawn (with replacement) from {$\hat{F}$} \citep{efron:1993}. Each of the $B$ resamples collected are generally denoted as $\bm{z}_b^*$, $b=1,…,B$. The sampling distribution of $T$ is then estimated by the empirical distribution of $T_b = g(\bm{z}_b)$, $b=1,…,B$, calculated on each bootstrapped resample $\bm{z}_b^*$.

Bootstrapping is regularly applied in the construction of prediction intervals, with several different strategies available for constructing PIs. The two bootstrapping variations examined in this research are the pivot and percentile methods.

\subsubsection{Pivot Bootstrap Method}

A popular bootstrapping approach for constructing PIs is a pivotal quantity method where the distribution of the unknown target $y_{test}$, conditioned on the observed instance $\bm{x}_{test}$, is assumed to be approximately normal. In this approach, the half-width of the PI is a function of the estimates for the model-fitting and irreducible errors, with the interval centered at the bootstrap aggregated prediction for $y_{test}$. This prediction, denoted as $\hat{f}(\bm{x}_{test})$, is found through the simple averaging of each resampled model: 
\\
\begin{equation} \label{bagged_pred}
\hat{f}(\bm{x}_{test})=\frac{1}{B}\sum_{b=1}^{B} \hat{f}_b(\bm{x}_{test}).
\end{equation}
The pivot bootstrap PI, $\Gamma_{pivot}$, is the interval 
\\
\begin{equation} \label{pivot_PI}
\Gamma_{pivot}=\left[\hat{f}(\bm{x}_{test}) - z_{1-\alpha/2}\sqrt{\hat{\sigma}^2(\bm{x}_{test})+\hat{\sigma}^2_{\epsilon}}, \text{ } \hat{f}(\bm{x}_{test}) + z_{1-\alpha/2}\sqrt{\hat{\sigma}^2(\bm{x}_{test})+\hat{\sigma}^2_{\epsilon}} \right],
\end{equation}
where $\hat{\sigma}^2(\bm{x}_{test})$ is the estimated prediction variance, $\hat{\sigma}^2_{\epsilon}$ is the estimated irreducible error, and $z_{1-\alpha/2}$ is the $(1-\alpha/2)$-th percentile of the standard normal distribution. In general, using the corresponding percentile from the Student’s $t$ distribution is a more mathematically rigorous method to capture the variability of the estimated target value. However, determining the degrees of freedom when modeling with a neural network is not a straightforward process \citep{gao-jojic:2016}. The standard normal distribution is used here for simplicity.  

Suppose that the feature and target data, \bm{$X$} and \bm{$y$}, are observed. $B$ bootstrap resamples are collected, and a neural network is trained on each to produce an estimated regression function $\hat{f}_b$, $b=1,…,B$. Now, suppose a new feature vector, $\bm{x}_{test}$, is observed and it is desired to provide a point estimate and PI for its unknown target value, $y_{test}$. After calculating $\hat{f}(\bm{x}_{test})$, as in (\ref{bagged_pred}), constructing the PI is a matter of estimating the component model-fitting and irreducible errors.
The latter prediction variance estimate for $\bm{x}_{test}$, $\hat{\sigma}^2(\bm{x}_{test})$, can be calculated from the distribution of predicted target values calculated from the bootstrap resamples, 

\begin{equation} \label{pivot_pred_var}
\hat{\sigma}^2(\bm{x}_{test})=\frac{1}{B-1}\sum_{b=1}^{B} \Big[\hat{f}_b(\bm{x}_{test})-\hat{f}(\bm{x}_{test}) \Big]^2.
\end{equation}
The irreducible error $\hat{\sigma}^2_{\epsilon}$ can be similarly estimated from the residuals of the out-of-sample predictions. In bootstrapping, out-of-sample observations are conveniently provided in the form of the “out-of-bag” sets. For a bootstrap resample $\bm{z}_b^*$, the corresponding set of out-of-bag observations is $\{z_i \in \bm{z} | z_i \notin \bm{z}_b^*\}$ and is denoted here as $\Tilde{\bm{z}}_b^*$. Then, for each bootstrap resample, $\hat{\sigma}^2_{\epsilon(b)}$ is: 
\\
\begin{equation} \label{resampled_data_noise}
\hat{\sigma}^2_{\epsilon(b)}=\frac{\sum_{\bm{x^{(b)}_i \in \Tilde{\bm{X}}_b^*}}^{} \Big[y^{(b)}_i-\hat{f}_b(\bm{x}^{(b)}_i) \Big]^2}{|\Tilde{n}_b-1|},
\end{equation}
where: \\
\indent \textbullet{\indent $\Tilde{n}_b$ is the number of observations in the out-of-bag set $\Tilde{\bm{X}}_b^*$} \\
\indent \textbullet{\indent $\bm{x}^{(b)}_i$ is the $i$\textsuperscript{th} observation in  $\Tilde{\bm{X}}_b^*$} \\
\indent \textbullet{\indent $y^{(b)}_i$ is the $i$\textsuperscript{th} observation in the out-of-bag set $\Tilde{\bm{y}}_b^*$ and the target value of $\bm{x}_i^{(b)}$} \\
\\
The irreducible error estimate for the entire data set then is:
\\
\begin{equation} \label{pivot_data_noise}
\hat{\sigma}^2_{\epsilon}=\frac{1}{B}\sum_{b=1}^{B} \hat{\sigma}^2_{\epsilon(b)}.
\end{equation}

Algorithm \ref{alg:pivot_bootstrap} provides pseudocode for calculating each component of the pivot bootstrap PI. While the approach described in Algorithm \ref{alg:pivot_bootstrap} is readily applied, the performance of the PIs constructed with the pivot bootstrap method have their own limitations. In particular, confidence intervals constructed in this manner tend be conservative \citep{papadopoulos-et-al:2001}, indicating the desired coverage could be achieved with a smaller prediction region. Further note that the performance of the PIs is dependent upon the sampling distribution of $\hat{f}(\bm{x}_{test})$; if it is skewed or biased, the validity and efficiency of the resulting PIs will deteriorate. \\

\begin{algorithm}[H]
\caption{Pivot Bootstrap Method}\label{alg:pivot_bootstrap}
\KwInput{training data \bm{$X$} and \bm{$y$}, test observation $\bm{x}_{test}$, learning algorithm $L$, desired number of bootstrap resamples $B$, and desired coverage probability $1-\alpha$}
\For{$b=1$ to $B$}
    {Generate bootstrap resamples of \bm{$X$} and \bm{$y$}; denote them as $\bm{X}^*_b$ and $\bm{y}^*_b$ \\
    Find the out-of-bag sets and denote them as $\Tilde{\bm{X}}^*_b$ and $\Tilde{\bm{y}}^*_b$ \\
    Train learning algorithm $L$ on $\bm{X}^*_b$ and $\bm{y}^*_b$; denote the trained regressor as $\hat{f}_b$ \\
    Calculate $\hat{f}_b(\bm{x}_{test})$ and $\hat{f}_b(\bm{X}^*_b)$ \\
    Calculate $\hat{\sigma}^2_{\epsilon(b)}$ as in (\ref{resampled_data_noise})}
{Calculate $\hat{f}(\bm{x}_{test})$ as in (\ref{bagged_pred})} \\
{Calculate $\hat{\sigma}^2(\bm{x}_{test})$ as in (\ref{pivot_pred_var})}\\
{Calculate $\hat{\sigma}^2_{\epsilon}$ as in (\ref{pivot_data_noise})}\\
\KwOutput{a $1-\alpha$ prediction interval, as constructed in (\ref{pivot_PI})}
\end{algorithm}

\subsubsection{Percentile Bootstrap Method}

A $1-\alpha$ PI constructed using the percentile bootstrap method leverages the empirical distribution of the predicted values $\hat{f}_b$, $b=1,…,B$ and a random sampling of the error terms. This procedure weakens assumptions regarding the sampling distribution of  $\hat{f}$ and does not require the explicit computation of component error terms. 

The collection of values $\hat{f}_b(\bm{x}_{test})$ for $b=1,…,B$ yield an empirical cumulative distribution function (ECDF), $\hat{F}$. Then, the percentile bootstrap algorithm computes the $\alpha⁄2$ and $1-\alpha⁄2$ percentiles from $\hat{F}$, $\hat{F}_{(\alpha⁄2)}$  and $\hat{F}_{(1-\alpha⁄2)}$, respectively. Note that
\\
\begin{equation} \label{percentile_property}
\text{P} \left(\hat{F}_{(\alpha⁄2)} < \hat{f}(\bm{x}_{test}) < \hat{F}_{(1-\alpha⁄2)} \right) \approx 1-\alpha.
\end{equation}
To accurately estimate the tail percentiles from sampling distributions requires specifying $B$ to be relatively large (i.e., $B>1,000$) \citep{efron:1993}. 

A valid confidence interval for the value $f(\bm{x}_{test})$, or the expected value of $y_{test}$ given $\bm{x}_{test}$, can be constructed by inverting (\ref{percentile_property}). In contrast, a valid PI for $y_{test}$ must also account for the irreducible error. To do so, the values of $\hat{F}$ are adjusted with a random error, sampled from the prediction errors of an out-of-sample data set \citep{davidson:1997}. When employing bootstrap resampling, such sets come in the form of the out-of-bag sets, $\Tilde{\bm{X}}_b^*$ and $\Tilde{\bm{y}}_b^*$, of each $b$\textsuperscript{th} resample of training data, $\bm{X}_b^*$ and $\bm{y}_b^*$, respectively. Therefore, to construct the PI, prediction errors for each fitted regressor $\hat{f}_b$ need to be calculated:
\\
\begin{equation} \label{random_resid_sample}
\bm{r}_b^* = \Tilde{\bm{y}}_b^* - \hat{f}_b(\Tilde{\bm{X}}_b^*).
\end{equation}
From each collection $\bm{r}_b^*$, a random sample of size one is collected---these values are denoted here as $e_b^*$, $b=1,...,B$. Then, for each $\hat{f}_b(\bm{x}_{test})$, the random-error-adjusted value, $\hat{g}_b(\bm{x}_{test})$, is calculated:
\begin{equation} \label{calculate_g_hat}
\hat{g}_b(\bm{x}_{test}) = \hat{f}_b(\bm{x}_{test}) +e_b^*.
\end{equation}
Using the set of values $\hat{g}_b(\bm{x}_{test})$, $b=1,…,B$, an ECDF, $\hat{G}$, can be constructed. Then, for any desired significance level $\alpha$,
\\
\begin{equation} \label{G_percentile_property}
\text{P} \left(\hat{G}_{(\alpha⁄2)} < y_{test} < \hat{G}_{(1-\alpha⁄2)} \right) \approx 1-\alpha,
\end{equation}
resulting in a $1-\alpha$ PI for $y_{test}$ of the form
\\
\begin{equation} \label{percentile_PI}
\Gamma_{percentile} = \left[\hat{G}_{(\alpha⁄2)},\text{  } \hat{G}_{(1-\alpha⁄2)} \right].
\end{equation}
Algorithm \ref{alg:percentile_bootstrap} provides a pseudocode implementation of the percentile bootstrap PI method \citep{davidson:1997}.
\\

\begin{algorithm}[H]
\caption{Percentile Bootstrap Method}\label{alg:percentile_bootstrap}
\KwInput{training data \bm{$X$} and \bm{$y$}, test observation $\bm{x}_{test}$, learning algorithm $L$, desired number of bootstrap resamples $B$, and desired coverage probability $1-\alpha$}
\For{$b=1$ to $B$}
    {Generate bootstrap resamples of \bm{$X$} and \bm{$y$}; denote them as $\bm{X}^*_b$ and $\bm{y}^*_b$ \\
    Find the out-of-bag sets and denote them as $\Tilde{\bm{X}}^*_b$ and $\Tilde{\bm{y}}^*_b$ \\
    Train learning algorithm $L$ on $\bm{X}^*_b$ and $\bm{y}^*_b$; denote the trained regressor as $\hat{f}_b$ \\
    Calculate $\hat{f}_b(\bm{x}_{test})$ and $\hat{f}_b(\tilde{\bm{X}}^*_b)$ \\
    Calculate the prediction errors $\bm{r}^*_b$ as in (\ref{random_resid_sample}) \\
    From $\bm{r}^*_b$, randomly sample a value $e^*_b$ \\
    Calculate $\hat{g}_b(\bm{x}_{test})$ as in (\ref{calculate_g_hat})}
{Construct the ECDF $\hat{G}$ from the values $\hat{g}_b(\bm{x}_{test})$, $b=1,...,B$}\\
\KwOutput{a $1-\alpha$ prediction interval, as constructed in (\ref{percentile_PI})}
\end{algorithm}

\subsection{Conformal Inference}

Conformal inference (CI) methods provide a potentially attractive alternative to the bootstrap method for constructing PIs from neural network outputs, with many of the same advantages as bootstrapping. With CI we can avoid some of the computational cost associated with training hundreds or thousands of networks. In particular, CI methods can be easily implemented with no changes to the underlying prediction algorithm \citep{papadopoulos:2008}. They are also agnostic to the quantitative and distributional properties of the target variable, applicable to both continuous and discrete outcomes, and thus, can be applied to any machine learning task \citep{schafer-vovk:2008}. Like bootstrapping, CI does not require weighty assumptions regarding the distribution of the target variable. The only assumption guaranteeing the validity of the PIs constructed is that the data be exchangeable \citep{schafer-vovk:2008}. A sequence of $n$ random variables is said to be exchangeable if any of $n!$ possible permutations of observing them are equally likely. 

The central idea of CI is to use a measure of ``nonconformity" to evaluate how ``strange” a potential test observation is compared to previously observed data \citep{gammerman-et-al:1998}. Typically, a context-appropriate distance function is chosen as the nonconformity measure; the absolute residual is generally used for regression tasks. 

To understand how conformal PIs are constructed, suppose a value $q$ is being considered for potential inclusion into the set $\Gamma$, a $1-\alpha$ PI for $y_{test}$. The choice of whether to include $q$ in $\Gamma$ is based upon a statistical test of the null hypothesis $y_{test}=q$ \citep{lei-et-al:2018}.

The first step of this test is to calculate ``conformity score” $R(\bm{x}_{test},q)$, such that 
\\
\begin{equation} \label{Conform_Score}
R(\bm{x}_{test},q) = d\left(\hat{f}(\bm{x}_{test}),q\right),
\end{equation}
where $\hat{f}(\bm{x}_{test})$ is the prediction for $y_{test}$ and $d$ is a measurable function mapping to $\mathbb{R}$. The strangeness of $q$ is determined by comparing the conformity score associated with $(\bm{x}_{test},q)$ to a set of other conformity scores, denoted here as $C$. With what data these other scores are calculated varies by the different conformal inference methods and is discussed further in the succeeding subsections. However, for all methods, a test statistic for $R(\bm{x}_{test},q)$ can be calculated: 
\\
\begin{equation} \label{calc_p_values}
\pi(q) = \frac{\sum_{i=1}^{l} \mathbb{I}\left[R(\bm{x}_{i},y_i)> R(\bm{x}_{test},q)\right]}{l+1},
\end{equation}
where $\mathbb{I}$ is the indicator function and $l=|C|$. By exchangeability, $\pi(q)$ is sub-uniform over the set $\left\{\frac{0}{l+1},\frac{1}{l+1},...,\frac{l}{l+1}\right\}$, implying: 
\begin{align*}
\mathbb{P}\left(\pi(q) \leq \alpha \right) \leq \alpha.
\end{align*} \\
Since $\pi(q)$ is sub-uniform, it is a valid $p$-value. Thus, if $\pi(q) < \alpha$ then there is sufficient evidence that $y_{test}$ will not equal $q$, and thus, $q$ should not be included in $\Gamma$. 

When $\bm{y}$ is continuous, the number of candidates that can be considered for this process is infinite. To practically implement CI for regression tasks, a very fine grid of target values $Q=\{q_1,q_2,…,q_M\}$ is examined, where $Q$ represents a size $M$ partition of the candidate range. A conservative $1-\alpha$ PI is then constructed as:
\\
\begin{equation} \label{conformal_PI}
\Gamma = \left\{q_j \in Q \text{ }| \text{ } \pi(q_j) \geq \alpha \right\}
\end{equation}. 

This research evaluates the performance and computational feasibility of four conformal inference algorithms: full conformal inference, split conformal inference, cross-conformal inference and bootstrap conformal inference. Operating in the trade-space of performance and computational efficiency, each has their own advantages and drawbacks. These methods, in addition to a modification in which a fitted kernel density function is used as the nonconformity measure, are discussed in the following sections.

\subsubsection{Full Conformal Inference Method}

Given training data $\bm{X}$ and $\bm{y}$ of size $n$, the full CI method provides confidence estimation for “particular to particular” new observations \citep{saunders-et-al:1999}. That is, if a new instance $\bm{x}_{test}$ is observed, the algorithm learns a PI particular to that new instance, as opposed to a general rule to use for all future instances. 

To construct a $1-\alpha$ PI for the target $y_{test}$, each $q_j$ in a set of candidate values $Q$ is evaluated against the originally observed data $\bm{X}$ and $\bm{y}$. In particular, $\bm{X}$ and $\bm{y}$ are augmented with the data pair $(\bm{x}_{test}, q_j)$. A learning algorithm, $L$, is trained on this augmented data set; denote the trained regressor as $\hat{f}_{aug}$. For every data pair in the augmented data set, conformity scores are computed as in (\ref{Conform_Score}) using $\hat{f}_{aug}$. The set of conformity scores against which every candidate value $q$ is compared is then:
\\
\begin{equation} \label{full_conformal}
C_{full}=\{R(\bm{x}_i,y_i)\text{ }|\text{ } (\bm{x}_i,y_i) \in (\bm{X}_{aug},\bm{y}_{aug})\},
\end{equation}
where $(\bm{X}_{aug},\bm{y}_{aug})$ is the augmented training set. Next, $\pi(q_j)$ is calculated using (\ref{calc_p_values}); $q_j$'s inclusion into $\Gamma$ is regulated by whether or not $\pi(q_j) \geq \alpha$. Repeating over each value in $Q$, the $1-\alpha$ PI is constructed as in (\ref{conformal_PI}). Algorithm \ref{alg:full_conformal} provides a pseudocode implementation of the full CI method. 

\begin{algorithm}[h]
\caption{Full Conformal Inference Method}\label{alg:full_conformal}
\KwInput{training data \bm{$X$} and \bm{$y$}, test observation $\bm{x}_{test}$, learning algorithm $L$, a set of candidate values $Q$, nonconformity measure $d$, and desired coverage probability $1-\alpha$}
\For{$q_j \in Q$}
    {Augment \bm{$X$} and \bm{$y$} with the data pair $(\bm{x}_{test},q_j)$ and denote them as $\bm{X}_{aug}$ and $\bm{y}_{aug}$ \\
    Fit learning algorithm $L$ to the augmented training set and denote the trained regressor as $\hat{f}_{aug}$ \\
    Compute $R(\bm{x}_i,y_i)$ for $(\bm{x}_i,y_i) \in (\bm{X}_{aug}, \bm{y}_{aug})$ as in (\ref{Conform_Score}) and $\hat{f}_{aug}$ \\
    Build $C_{full}$ as in (\ref{full_conformal}) \\
    Compute the $p$-value of $R(\bm{x}_{test},q_j)$ as in (\ref{calc_p_values})}
\KwOutput{a $1-\alpha$ prediction interval, as defined in (\ref{conformal_PI})}
\end{algorithm}

While experimental results suggest this approach produces efficient and valid PIs (\citealt{linusson-et-al:2014}; \citealt{lei-et-al:2018}), its computational burden becomes undesirable when the training process for $L$ is time-consuming. Since, in this approach, an augmented regression function is learned for each $q_j$ in $Q$, $L$ must be trained $|Q|$ times. While under certain regularity conditions we can avoid such repeated computations (\citealt{nouretdinov:2001}; \citealt{ndiaye:2022}), we do not explore these approaches in our work.   

\subsubsection{Split Conformal Inference Method}

The split conformal inference method alters the CI approach to be inductive. That is, given training data $\bm{X}$ and $\bm{y}$, split CI creates a general rule to construct a prediction interval for any future observation.

To do so, $\bm{X}$ and $\bm{y}$ are randomly partitioned into two subsets: ($\bm{X}_{proper}$, $\bm{y}_{proper}$) and ($\bm{X}_{cal}$, $\bm{y}_{cal}$), the ``proper training" and the ``calibration" sets, respectively. A learning algorithm $L$ is trained on ($\bm{X}_{proper}$, $\bm{y}_{proper}$); denote the estimated regression function as $\hat{f}_{split}$. The set of conformity scores $C_{split}$, used to evaluate candidate values, is built as: 
\\
\begin{equation} \label{split_conformal}
C_{split}=\{R(\bm{x}_i,y_i)\text{ }|\text{ } (\bm{x}_i,y_i) \in (\bm{X}_{cal},\bm{y}_{cal})\},
\end{equation}
where each conformity score $R(\bm{x}_i,y_i)$ is calculated as in (\ref{Conform_Score}) using $\hat{f}_{split}$. Now, for any future instance $\bm{x}_{test}$ the scores in $C_{split}$ are used to evaluate candidate values for potential inclusion in the interval, specifically by calculating a $p$-value as in (\ref{calc_p_values}). A pseudocode implementation of the split conformal inference method is provided in Algorithm \ref{alg:split_conformal} \citep{lei-et-al:2018}.

\begin{algorithm}[h]
\caption{Split Conformal Inference Method}\label{alg:split_conformal}
\KwInput{training data \bm{$X$} and \bm{$y$}, test observation $\bm{x}_{test}$, learning algorithm $L$, a set of candidate values $Q$, nonconformity measure $d$, and desired coverage probability $1-\alpha$}
{Evenly split \bm{$X$} and \bm{$y$} into two sub-sets: the proper training set ($\bm{X}_{proper}$, $\bm{y}_{proper}$), and the calibration set ($\bm{X}_{cal}$, $\bm{y}_{cal}$)}\\
{Fit learning algorithm $L$ to the proper training set and denote the trained regressor as $\hat{f}$}\\
{Build $C_{split}$ as in (\ref{split_conformal})}\\
\For{$q_j \in Q$}
    {Compute $R(\bm{x}_{test},q_j)$ as in (\ref{Conform_Score}) using $\hat{f}$ \\
    Compute the $p$-value of $R(\bm{x}_{test},q_j)$ as in (\ref{calc_p_values}) using $\hat{f}$}
\KwOutput{a $1-\alpha$ prediction interval, as defined in (\ref{conformal_PI})}
\end{algorithm}

Split CI presents a clear computational advantage over the full conformal inference and bootstrap methods. For learning algorithms relying on time-consuming gradient descent optimization (e.g., neural networks) the reduction in computation time is non-trivial. On the other hand, the PIs constructed with the split CI method are ``less informationally efficient” by virtue of using only a subset of the original sample for training the learning algorithm \citep{vovk:2015}. Moreover, the randomness introduced by the single splitting of the data---along with, in this research, the use of a single, trained neural network for predictions---increases the variability of PI performance across test sets. 

\subsubsection{Aggregated Conformal Predictors}

A compromise between the computationally-intensive full CI method and the cheap but high-variance split CI method is to leverage an ``aggregated conformal predictor" (ACP) \citep{carlsson-et-al:2014}. An ACP uses a resampling procedure, such as $k$-fold partitioning or bootstrapping, to generate multiple resamples of the training data set. Given a test observation $\bm{x}_{test}$, candidate value $q$, and conformity score $R(\bm{x}_{test}, q)$, these resamples are leveraged to produce several estimates of the $p$-value for $q$. The idea of using an ACP is to calculate more accurate and stable $p$-values for constructing PIs. Two common aggregated CI approaches, which are discussed here, are the cross-conformal inference (cross-CI) and bootstrap CI methods. 

The implementation of the cross-CI is similar to split CI, as both methods are inductive. However the process of fitting a learning algorithm $L$ and calculating $\pi(q_j)$ (for a candidate value $q_j \in Q$), is repeated across $K$ splits of the data. The separate $K$ estimates for $\pi(q_j )$, $\pi_k(q_j)$ for $k=1,…,K$, are then aggregated to produce a single value, $\bar{\pi}(q_j)$ \citep{vovk:2015}. Assuming that the splits are approximately of equal size, and that $K$ is significantly smaller than the number of observations in the original sample, the aggregation function can be the average of each $\pi_k(q_j)$:
\\
\begin{equation} \label{aggregating_p_values}
\bar{\pi}(q_j)=\frac{1}{K}\sum_{k=1}^{K} \pi_k(q_j).
\end{equation}
The $1-\alpha$ PI can then be constructed for a test observation $\bm{x}_{test}$, with unknown target value $y_{test}$, as 
\\
\begin{equation} \label{aggregate_conformal_PI}
\Gamma_{aggregated} = \left\{q_j \in Q \text{ } | \text{ } \bar{\pi}(q_j) \geq \alpha \right\}, 
\end{equation}
where $Q$ is the set of candidates considered for $y_{test}$ \citep{vovk:2015}. Algorithm \ref{alg:cross_conformal} provides a pseudocode implementation of the cross-CI algorithm.

\begin{algorithm}[h]
\caption{Cross-Conformal Inference Method}\label{alg:cross_conformal}
\KwInput{training data \bm{$X$} and \bm{$y$}, test observation $\bm{x}_{test}$, desired number of folds $K$, learning algorithm $L$, a set of candidate values $Q$, nonconformity measure $d$, and desired coverage probability $1-\alpha$}
{Evenly split \bm{$X$} and \bm{$y$} into $K$ sub-sets, denoting each as $(\bm{X}_k,\bm{y}_k)$} \\
\For{$k=1$ to $K$}
{The calibration set is $(\bm{X}_k,\bm{y}_k)$ and the proper training set is $(\bm{X}_{-k},\bm{y}_{-k}) \equiv \left \{(\bm{x}_i,y_i) \in (\bm{X},\bm{y}) \text{ } | \text{ } (\bm{X},\bm{y}) \notin (\bm{X}_k,\bm{y}_k) \right \}$ \\
Fit learning algorithm $L$ to the proper training set and denote the trained regressor as $\hat{f}_k$ \\
\For{$q_j \in Q$}
{$C_k = \left \{R(\bm{x}_i,y_i); \text{ } (\bm{x}_i,y_i) \in (\bm{X},\bm{y}) \right \}$ where each $R(\bm{x}_i,y_i)$ is calculated as in (\ref{Conform_Score}) using $\hat{f}_k$ \\
Compute $R(\bm{x}_{test},q_j)$ as in \ref{Conform_Score} using $\hat{f}_k$ \\
Compute the p-value of $R(\bm{x}_{test},q_j)$ as in (\ref{calc_p_values})}}
{Compute $\bar{\pi}(q_j)$ for $q_j \in Q$ as in (\ref{aggregating_p_values})}\\
\KwOutput{a $1-\alpha$ prediction interval, as defined in (\ref{aggregate_conformal_PI})}
\end{algorithm}

An alternate approach, bootstrap CI, leverages the concepts of bootstrap resampling to generate several calculations of $p$-values which are then aggregated. The algorithmic approach of the bootstrap CI method is effectively the same as cross-CI, with the only change being the method of resampling.

Suppose the training data $\bm{X}$ and $\bm{y}$ are bootstrap resampled $B$ times; denote each resample as $\bm{X}_b^*$ and $\bm{y}_b^*$. Then, each $\bm{X}_b^*$ and $\bm{y}_b^*$ serves as the proper training set, with its corresponding set of out-of-bag samples, $\Tilde{\bm{X}}_b^*$ and $\Tilde{\bm{y}}_b^**$, serving as the calibration set. Following the same process as in cross-CI, conformity scores are calculated from each $\Tilde{\bm{X}}_b^*$ and $\Tilde{\bm{y}}_b^**$. Then, for a test observation $\bm{x}_{test}$, the $p$-value of a candidate value $q_j$ can be calculated on each of the $b$ sets of conformity scores: $\pi_b(q_j)$, $b=1,...,B$ \citep{vovk:2015}. These values are then aggregated to a single value, $\bar{\pi}(q_j)$, calculated as their arithmetic mean. A $1-\alpha$ PI is then constructed as in (\ref{aggregate_conformal_PI}). Algorithm \ref{alg:bootstrap_conformal} provides a pseudocode implementation of the bootstrap CI method.

\begin{algorithm}[h]
\caption{Bootstrap Conformal Inference Method}\label{alg:bootstrap_conformal}
\KwInput{training data \bm{$X$} and \bm{$y$}, test observation $\bm{x}_{test}$, desired number of resamples $B$, learning algorithm $L$, a set of candidate values $Q$, nonconformity measure $d$, and desired coverage probability $1-\alpha$}
{Generate $B$ resamples of \bm{$X$} and \bm{$y$} into $K$ sub-sets, denoting each as $(\bm{X}_b^*,\bm{y}_b^*)$} \\
\For{$b=1$ to $B$}
{The proper training set is $(\bm{X}_b^*,\bm{y}_b^*)$ and the calibration set is $(\Tilde{\bm{X}}_b^*,\Tilde{\bm{y}}_b^*)$ \\ 
Fit learning algorithm $L$ to the proper training set and denote the trained regressor as $\hat{f}_b$ \\
\For{$q_j \in Q$}
{$C_b = \left \{R(\bm{x}_i,y_i); \text{ } (\bm{x}_i,y_i) \in (\bm{X},\bm{y}) \right \}$ where each $R(\bm{x}_i,y_i)$ is calculated as in (\ref{Conform_Score}) using $\hat{f}_k$ \\
Compute $R(\bm{x}_{test},q_j)$ as in (\ref{Conform_Score}) using $\hat{f}_b$ \\
Compute the p-value of $R(\bm{x}_{test},q_j)$ as in (\ref{calc_p_values})}}
{Compute $\bar{\pi}(q_j)$ for $q_j \in Q$ as in (\ref{aggregating_p_values})} \\
\KwOutput{a $1-\alpha$ prediction interval, as defined in (\ref{aggregate_conformal_PI})}
\end{algorithm}

\subsubsection{Conformal Inference with Kernel Density Estimation}

While the absolute error is the typical choice for the nonconformity measure for regression tasks, the performance of these CI algorithms may suffer if the distribution of prediction errors does not behave as expected (e.g., if they are biased, skewed, or multi-modal). Neural networks compound this problem, as well. The parameter values of a neural network will generally converge to different values across different training sessions due to randomness in the optimization procedure, specifically random initialization and the existence of local minima. This effect adds an additional layer of noise to the network’s fit to the data and, by extension, the prediction errors.

One solution to this issue is kernel density estimation (KDE), a non-parametric method for estimating a probability density function (PDF) over a set of observed data \citep{rosenblatt:1956}. Estimating the PDF through KDE preserves the general shape of the data, while smoothing over spurious deviations arising from random sampling.

The two hyperparameters to be tuned for fitting the KDE to observed data are the kernel function ($K$) and bandwidth ($h$). The bandwidth parameter is effectively the strength of smoothing applied to the prediction errors. As $h\rightarrow0$, the estimated PDF more closely resembles the observed data; as $h\rightarrow\infty$, it becomes smoother and flatter over the range of observed values \citep{rosenblatt:1956}. There exists several rules of thumb and analytic methods for tuning $h$ \citep{lei-robins-was:2011}. However, for most practitioners, constructing and evaluating several density estimates from a finite grid of potential bandwidths $H=\{h_1,h_2,…,h_l\}$ is the most readily applied approach. 

A natural choice for $K$ in the context of evaluating regression estimates is the Gaussian kernel. As an example, suppose a KDE with Gaussian kernel and bandwidth $h$ is fitted to a set of observed values $\bm{z}=\{z_1,z_2,…,z_n\}$. Then for an arbitrary value $u$, the estimated density, $\hat{p}(u)$, under the fitted KDE is
\begin{align*}
\hat{p}(u) = \frac{1}{n}\sum_{i=1}^{n}\frac{1}{h}K\left(\frac{u-z_i}{h}\right),
\end{align*}
where the Gaussian kernel $K$ is calculated as
\begin{align*}
K \left(\frac{u-z}{h} \right) = \frac{1}{\sqrt{2\pi}}e^{\frac{1}{2}\left(\frac{u-z}{h}\right)^2}
\end{align*}
for any $z$ \citep{casella:2002}. Thus, for every potential value of $u$, the fitted KDE outputs the estimated density at $u$. The KDE function fitted to $\bm{z}$ does not necessarily resemble the original kernel function; so long as the smoothing parameter $h$ is not overly strong, the modalities and asymmetries in the empirical distribution of $\bm{z}$ will still be present. 

KDE gives a methods for estimating the likelihood (i.e., density) associated with a particular observation. Values in regions of high density are more likely, while values in regions of low density are less likely. Given the goal of CI to construct prediction intervals based on the "strangeness" of some candidate value, we can use KDE to construct a conformity measure where strangeness is inherently related to the estimated distribution of residuals, as opposed to the absolute value of residuals. Specifically, for a candidate value $q_j$, we can use $-ln\left(R(\bm{x}_i,y_i)\right)$ as our conformity measure, or any monotone function thereof. 

\section{Experimental Setup}

This research executes a two-step experiment on eleven data sets to evaluate how the PI methods discussed in Section 2 perform when used in conjunction with MLPs and CNNs. The first experimental step seeks to understand how changes in the settings of key neural network hyperparameters affect PI performance. In the second step, bootstrapping and CI PI methods are compared to determine which better optimizes the trade-off between performance and computational burden. In this section, we first describe the breadth of the data sets, and therefore data structures, used. Then, we provide further details about the experiments to be performed.

\subsection{Data Sets}

The data sets chosen for this research are intended to capture a variety of data structures, such that the expected performance of different PI methods can be well understood regardless of the specific task at hand. Additionally, the data sets used are familiar in the neural networks literature, used to support a variety of research goals (\citealt{hernandez-lobato-adam:2015}; \citealt{gal-ghahramani:2016}; \citealt{foong-et-al:2019}). Ten of these data sets (summarized in Table \ref{table:dataset-summary}) have one-dimensional feature vectors, for which MLPs are constructed during experimentation. 

\begin{table}[h]
\begin{center}
\begin{tabular}{|p{4.2cm}|r|r|r|}
  \hline
  \textbf{Data Set} & \textbf{Samples} (\bm{$n$}) & \textbf{Features} (\bm{$d$}) & \bm{$n/d$} \\ 
  \hline
  Boston Housing & $506$ & $13$ & $38.9$ \\ 
  \hline
  Wine Quality* & 1,599 & $11$ & $145.9$ \\ 
  \hline
  Concrete Strength* & 1,030 & 8 & 128.8 \\ 
  \hline
  Energy Efficiency* & 768 & 8 & 96.0 \\ 
  \hline
  Kinematics & 8,192 & 8 & 1024.0 \\ 
  \hline
  Naval Propulsion* & 11,934 & 16 & 745.9 \\ 
  \hline
  Power Plant* & 9,568 & 4 & 2392.0 \\ 
  \hline
  Protein Structure & 45,730 & 9 & 5081.1 \\ 
  \hline
  Yacht Hydrodynamics* & 308 & 6 & 51.3 \\
  \hline
  Year Prediction MSD* & 515,345 & 90 & 5726.1  \\ 
  \hline
\end{tabular}
\caption{Summary of benchmark data sets. An asterisk (*) denotes data sets accessed from the UCI Machine Learning Repository \citep{UCI_ML_Repo:2021}.}
\label{table:dataset-summary}
\end{center}
\end{table}

The eleventh data set is the Rotated Modified National Institute of Standards (RotNIST) data set of handwritten digits. The MNIST data set has been used regularly as a benchmark data set to evaluate the performance of novel network architectures (\citealt{jarrett-et-al:2009}; \citealt{Glorot-et-al:2011}; \citealt{lin-et-al:2014}). While the original MNIST data set is generally used for the classification of the handwritten digits, RotNIST extends its scope to regression tasks by applying random rotations to the digits. The task is then to train a neural network to predict these angles of rotation, which are applied uniformly from $-45$ and $45$ degrees.

\begin{figure}[h]
 \centering
 \resizebox{4.5in}{!}{
    \includegraphics{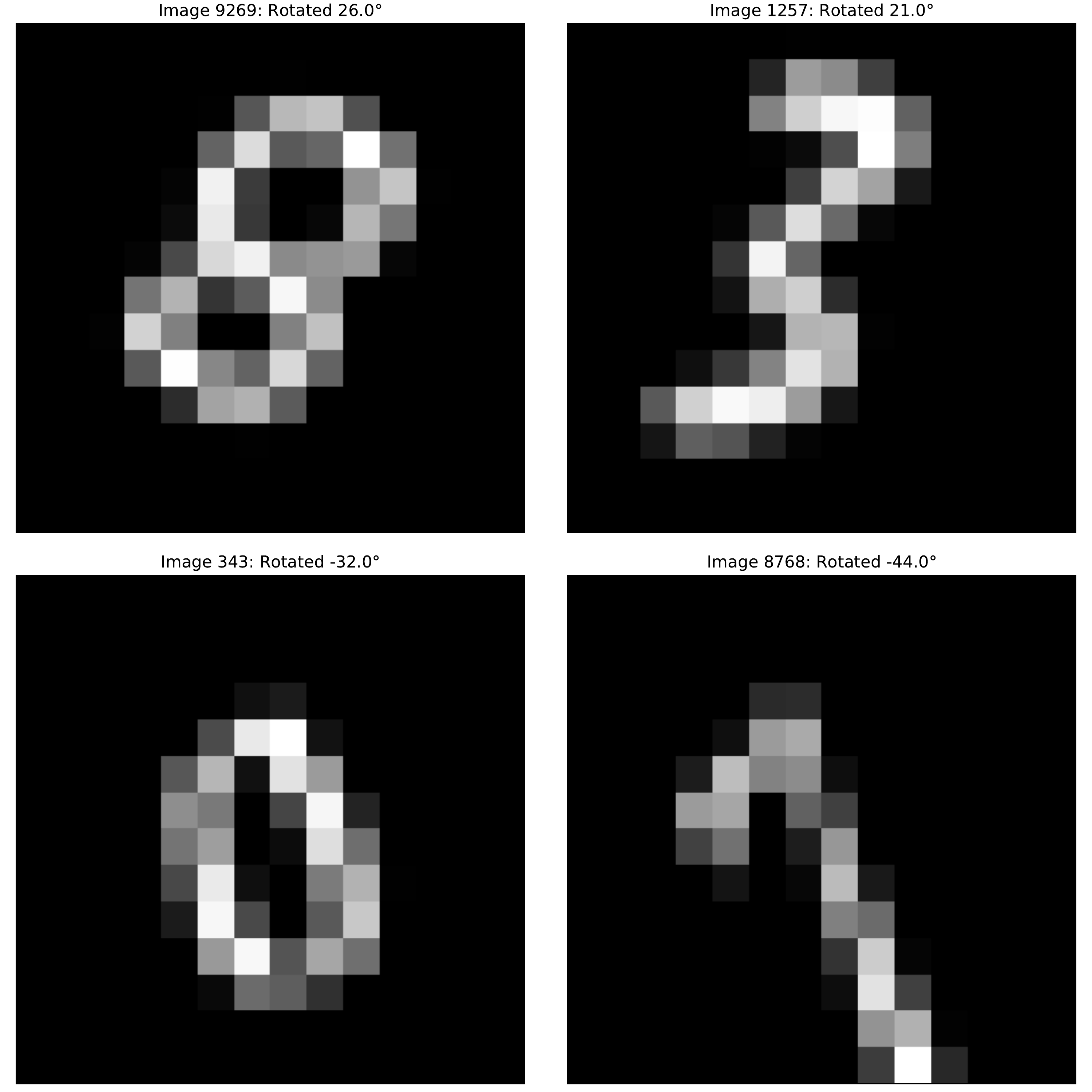}}
 \caption{Example images from the RotNIST dataset.}
 \label{fig:RotNIST_examples}
\end{figure}

The particular iteration of this data set used in this paper is downloaded from the MATLAB code platform (link: \href{https://www.mathworks.com/help/deeplearning/ug/data-sets-for-deep-learning.html}{instructions to download data set}). It contains 10,000 observations of $28\times28\times1$ images. Pixel values are scaled such that every entry is between zero and one. For this analysis, images are resized to $14\times14\times1$ using bilinear interpolation. Figure \ref{fig:RotNIST_examples} displays four sample images from the processed data set.

\subsection{Design of Experiment}

The first experimental step seeks to determine the effects of neural network hyperparameter selection on PI performance using an ANOVA approach. To do so, experimental designs of various MLP and CNN hyperparameters are constructed. These hyperparameter designs are then used to fit corresponding neural networks. The networks are trained on each data set in conjunction with the pivot bootstrap (1,000 resamples) and split conformal inference methods to construct 95\% PIs (i.e., $\alpha=0.05$) for observations in a pre-sequestered test set (20\% of original data set). The two PI methods are chosen as exemplars of the bootstrap and conformal inference approaches, respectively. The validity and efficiency of the constructed PIs are evaluated across the modeled network architectures. For robustness, a five-fold cross validation approach is implemented, with each instance in a data set used as a test observation exactly once. 

The experimental design for evaluating MLPs is summarized in Table \ref{table:DOE-benchmark}. Three factors are considered: activation function, number of hidden layers, and number of nodes in each hidden layers. Every combination of levels is explored, resulting in a total of $6\times3\times3=54$ network architectures. MLPs typically used to model the benchmark data sets utilize one or two hidden layers, each with 50 to 100 nodes with ReLU or Tanh activation functions (\citealt{hernandez-lobato-adam:2015}; \citealt{gal-ghahramani:2016}; \citealt{foong-et-al:2019}). By examining other popular activation functions (i.e., sigmoid) and constructing both small and large networks, the experimental design covers the  decision space typical of these key hyperparameters. Other network hyperparameters not included in this design are either assigned to constant values for each data set, such as the settings of the Adam optimization algorithm for training \citep{kingma-ba:2014}, or are outside of the scope of the analysis. In particular, regularization schemes are not considered for these MLPs. Algorithm \ref{alg:step1_exp} provides a pseudocode execution for implementing this first experimental design. 

\begin{table}[h]
\begin{center}
\begin{tabular}{|l|l|}
  \hline \textbf{Factor}& \textbf{Levels} \\
  \hline Activation & ReLU, Sigmoid, Tanh \\
  \hline Layers & 1, 2, 3 \\
  \hline Nodes & 5, 10, 25, 50, 75, 100 \\
  \hline
\end{tabular}
\caption{Design of experiment factor and levels for the benchmark data sets.}
\label{table:DOE-benchmark}
\end{center}
\end{table}

\begin{algorithm}[h]
\caption{First Experimental Step (Benchmark Data Sets)}\label{alg:step1_exp}
\For{each benchmark data set}
{\For{each design point in Table \ref{table:DOE-benchmark}}
{Build network using combination of hyperparameters; designate the untrained network as $L$ \\ 
Use 5-fold cross validation to create 5 pairs of training and test sets \\
\For{each training/test set pair}
{Run Algorithm \ref{alg:pivot_bootstrap} to implement Pivot Bootstrap method ($B=1,000$); calculate network ensemble's RMSE using out-of-bag sets \\ 
Run Algorithm \ref{alg:split_conformal} to implement Split Conformal Inference method; calculate network's RMSE using validation set}}
Perform ANOVA modeling to assess PI performance across network design choices}
\end{algorithm}

A similar experimental methodology to that portrayed in Algorithm \ref{alg:step1_exp} is pursued for the RotNIST dataset. However, this experimental design uses the relevant hyperparameters for a CNN; namely, the number of convolutional layers, the number of filters for each layer, and the kernel size. These experimental factors and their corresponding levels are shown in Table \ref{table:DOE-RotNIST}. All the CNNs fit in the experiment also employ regularization to improve test set performance. In particular, batch normalization is executed after each convolutional layer, as well as dropout before the output layer (probability$=0.05$) \citep{srivastava-et-al:2014}. 

\begin{table}[h]
\begin{center}
\begin{tabular}{|l|l|}
  \hline \textbf{Factor}& \textbf{Levels} \\
  \hline Convolutional Layers & 2, 3, 4 \\
  \hline Kernel Size & $1\times1$, $3\times3$, $5\times5$\\
  \hline Pooling & Average, Maximum \\
  \hline
\end{tabular}
\caption{Design of experiment factors and levels for the RotNIST data set.}
\label{table:DOE-RotNIST}
\end{center}
\end{table}

At the conclusion of this first experimental step, further analysis is performed to determine the sensitivity of network hyperparameters to PI performance and optimal network structure for each data set. Architectures are assessed according to the validity and efficiency of the PIs constructed with them. The optimal network architecture results in the most efficient PIs (smallest average widths) that maintain validity. The validity of a set of PIs is tested using the \citeauthor{agresti-coull:1998} method for binomial proportions \citep{agresti-coull:1998}. Valid PIs are those whose average coverage is not statistically different from 0.95. The average width of the PIs for the optimal network architecture and PI method is recorded as $W_j$ for the $j$\textsuperscript{th} data set. 

After determining the optimal neural network architecture for each data set, the second experimental step begins. In this step, the optimal network structure is used to implement each of the PI methods described in Section 2, such that the validity, efficiency, and computational burden of each can be evaluated and compared. For a given data set, 100 observations are randomly selected and sequestered as the test set. 95\% PIs are constructed for the test set using each PI method, with coverage and width recorded for each test observation. This portion of the experiment is repeated 10 times, after which summary statistics for PI coverage and average width are calculated using each repetition. The examined methods are summarized in Table \ref{table:step2-methods}. All CI methods are implemented using both the absolute residual and a fitted KDE as the conformity measure, with the results of both presented in Section 4. The pseudocode to execute this second experimental step is provided in Algorithm 8. Here, computational burden is assessed as the number of times the architecture has to be trained in order to implement the PI method, as the training of neural networks is generally the most time and resource intensive task for inference within deep learning tools. 

\begin{table}[h]
\begin{center}
\begin{tabular}{|l|r|r|}
  \hline \textbf{Method}& \textbf{Execution} & 
    \begin{tabular}{@{}r@{}}\textbf{Computational Burden} \\ (number of networks trained)\end{tabular} \\
  \hline Pivot Bootstrap & 100 resamples & 100 \\
  \hline Pivot Bootstrap & 500 resamples & 500 \\
  \hline Pivot Bootstrap & 1,000 resamples & 1,000 \\
  \hline Percentile Bootstrap & 1,000 resamples & 1,000 \\
  \hline \begin{tabular}{@{}l@{}}Bootstrap Conformal Inference \end{tabular} & 5 resamples & 5 \\
  \hline \begin{tabular}{@{}l@{}}Bootstrap Conformal Inference \end{tabular} & 10 resamples & 10 \\
  \hline \begin{tabular}{@{}l@{}}Bootstrap Conformal Inference \end{tabular} & 20 resamples & 20 \\
  \hline \begin{tabular}{@{}l@{}}Cross-Conformal Inference \end{tabular} & 5 folds & 5 \\
  \hline \begin{tabular}{@{}l@{}}Cross-Conformal Inference \end{tabular} & 10 folds & 10 \\
  \hline \begin{tabular}{@{}l@{}}Cross-Conformal Inference \end{tabular} & 20 folds & 20 \\
  \hline \begin{tabular}{@{}l@{}}Full Conformal Inference \end{tabular} & & 10,001 \\
  \hline \begin{tabular}{@{}l@{}}Split Conformal Inference \end{tabular} & & 1 \\
  \hline
\end{tabular}
\caption{Execution and computation burden of methods examined during experimentation.}
\label{table:step2-methods}
\end{center}
\end{table}

\begin{algorithm}[h]
\caption{Second Experimental Step}\label{alg:step1_exp}
\For{each data set}
{\For{trial 1 to 10}
{Randomly sample and remove 100 test observations from the original data set; designate the remaining observations as the training set \\
\For{each method \textit{A} in Table \ref{table:step2-methods}}
{Implement $A$ using the data set's optimal neural network structure \\ 
Record the performance metrics of the constructed PIs}}
Average replicates of PI performance metrics \\
Analyze PIs}
\end{algorithm}

Recall from Section 2 that the conformal inference methods require the examination of a fine grid of candidate target values. With the scale and variability of observations varying by data set, and considering the computational burden of full conformal inference, a systematic approach is needed to ensure that the location, size, and fineness of $Q$ considered for each observation is appropriate and experimentally feasible. To that end, the fineness of $Q$ is determined by the conformal inference method used. PIs constructed with full conformal inference examine a search grid of 100 values, while other conformal inference methods examined grids of 1,000 values. The width of $Q$ is set as twice the value of $W_j$, the average width of the optimal set of PIs for data set $j$ found in the first experimental step. Lastly, for each test observation $\bm{x}_{test}$, $Q$ is constructed such that it is centered around the point prediction for its target value. This prediction, $\hat{f}(\bm{x}_{test})$, is calculated from the underlying learning algorithm fit to the training data. For the full conformal inference method, point predictions for each test observation are found by training a neural network on the original training set, before the PI algorithm is implemented. Thus, the search grid for $\bm{x}_{test}$ from dataset $j$ is the set of 100 (or 1,000) evenly-spaced values in the interval $[\hat{f}(\bm{x}_{test})-W_j,\text{  }\hat{f}(\bm{x}_{test})+W_j]$.

\subsection{Code Implementation and Other Details}

Neural networks are trained in the Python coding language using the Keras library, with TensorFlow as the backend platform. Notebooks are executed on Google Colaboratory (“Google Colab”). The code is publicly available at \url{https://github.com/alexcontarino/Constructing-Prediction-Intervals-for-Neural-Networks}. 

For the 10 benchmark data sets shown in Table \ref{table:dataset-summary}, the Adam optimizer is used for training neural networks. Optimization settings, such as the learning rate, batch size, and number of training epochs, are tuned for each data set to ensure networks fit well to the training data and in a timely fashion. Other settings for the Adam optimizer function are left to their default value assigned in Keras. The optimization settings used for each data set are summarized in Table \ref{table:optimizer-settings}.

\begin{table}[h]
\begin{center}
\begin{tabular}{|l|r|r|l|}
  \hline \textbf{Data Set} & \textbf{Epochs} & \textbf{Batch Size} & \textbf{Learning Rate}\\
  \hline Boston Housing & 200 & 32 & 0.001 \\
  \hline Wine Quality & 75 & 32 & 0.001 \\
  \hline Concrete Strength & 200 & 32 & 0.001 \\
  \hline Energy Efficiency & 250 & 32 & 0.001 \\
  \hline Kinematics & 150 & 32 & 0.001 \\
  \hline Naval Propulsion & 80 & 256 & 0.001 \\
  \hline Power Plant & 350 & 600 & 0.01 \\
  \hline Protein Structure & 75 & 1024 & 0.025 \\
  \hline Yacht Hydrodynamics & 500 & 64 & 0.001 \\
  \hline Year Predictions MSD & 15 & 4096 & 0.1 \\
  \hline
\end{tabular}
\caption{Optimization settings for benchmark data sets.}
\label{table:optimizer-settings}
\end{center}
\end{table}

Parameters of the training algorithm for the CNNs tasked for learning the RotNIST data set are similarly tuned. For that data set, the stochastic gradient descent (SGD) algorithm with a learning rate of 0.001 is used to train networks for 20 epochs. The batch size for training is 64 samples. Other settings for the SGD optimizer are left to their Keras-assigned default values.

\section {Results}

This section provides results and discussion related to our two-stage experiment. We first present the effects of neural network hyperparameters on PI performance. The statistical significance of effects are measured and evaluated using an Analysis of Variance (ANOVA) approach. Next, we compare the performance of the PI methods discussed in Section 2, examining the differences in the coverage and average widths on similar data sets. Lastly, we perform a case study assessing the conditional coverage of select PI methods on various data sets. 

\subsection{Effects of Neural Network Hyperparameters}

The results of the first experimental step reveals how changes in neural network hyperparameter settings affect PI performance. Performance in terms of coverage and efficiency is discussed in the succeeding subsections. 

\subsubsection{Coverage}

We analyze PI coverage of both the pivot bootstrap and split conformal inference methods is analyzed using an ANOVA approach. With the factors identified in Table \ref{table:DOE-benchmark} and Table \ref{table:DOE-RotNIST} for MLPs and CNNs, respectfully, full factorial models are built to estimate coverage for each data set. These models include main effects, as well as all second- and third-order effects, in order to determine which hyperparameters are significant and, in particular, understand how neural network architecture affects PI coverage. 

Table \ref{table:coverage_sig} summarizes the significance of effects for the benchmark regression data sets. A significance level of 0.05 was chosen for this analysis. The most striking result is in the significance in effects when comparing the bootstrapping and split CI methods. For the former, in which it is expected that coverage remain around the nominal level, none of the effects are significant in any of the experimental data sets. However, a majority of effects are significant for modeling pivot bootstrap PI coverage in a given data set. For the latter PIs, coverage generally improves with the fit of the network. This result is intuitive given the pivot bootstrap method relies on the distribution of the target outcome, conditioned on the observed feature vector, being approximately Normal. 

\begin{table}[h]
 \centering
 \resizebox{6in}{!}{%
 \includegraphics{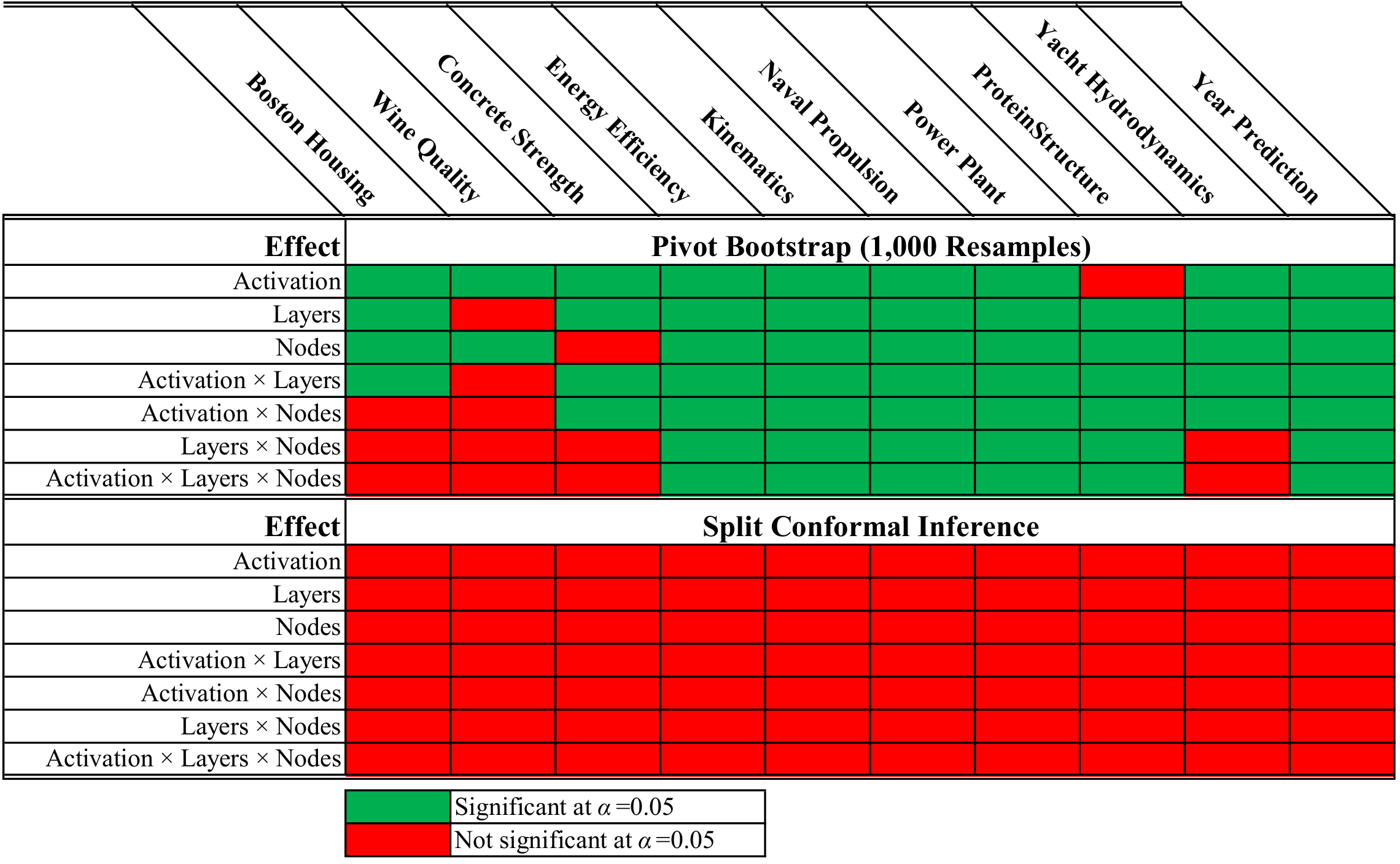}}
 \caption{Significance of effects modeling PI coverage.}
 \label{table:coverage_sig}
\end{table}

\begin{figure}[h]
\begin{center}
\begin{tabular}{|c|c|}
  \hline \textbf{Pivot Bootstrap ($\bm{1,000}$ Resamples) }& \textbf{Split Conformal Inference} \\
  \hline \resizebox{2.8in}{!}{\includegraphics{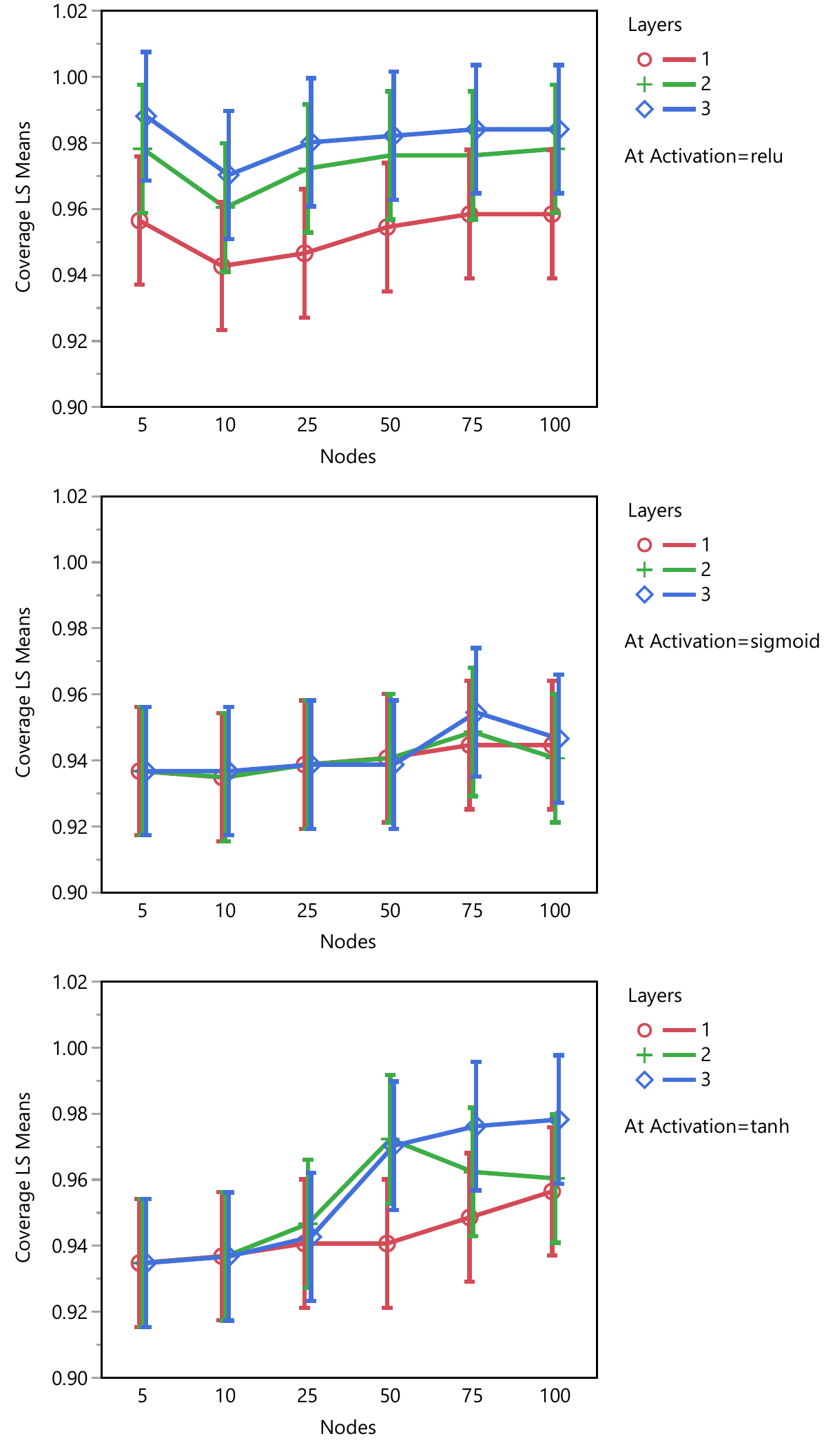}} & %
  \resizebox{2.8in}{!}{\includegraphics{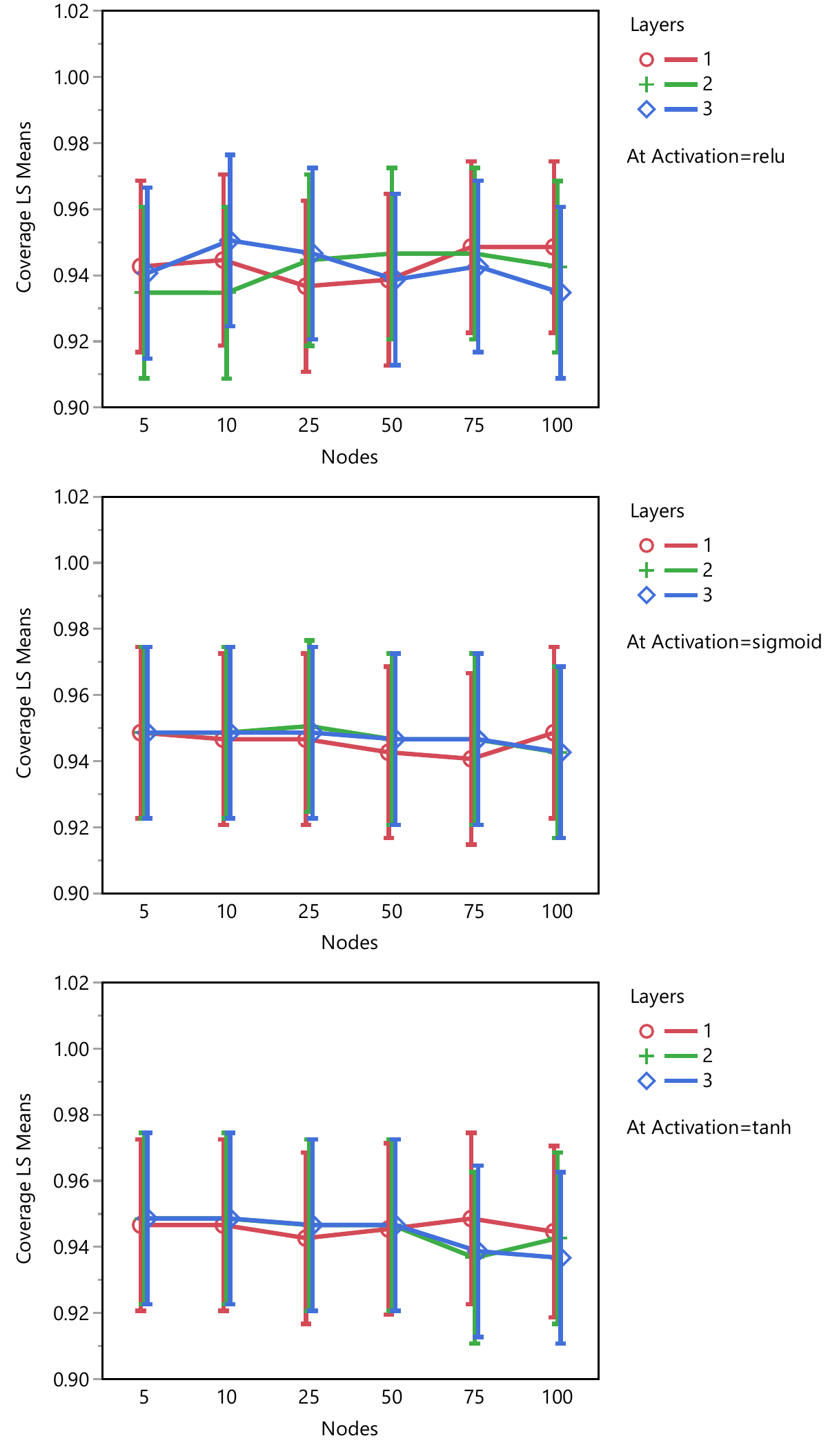}} \\
  \hline
\end{tabular}
\caption{PI Coverage in the Boston Housing Data Set.}
\label{figure:boston_PI_cov}
\end{center}
\end{figure}

\begin{figure}[h]
\begin{center}
\begin{tabular}{|c|c|}
  \hline \textbf{Pivot Bootstrap ($\bm{1,000}$ Resamples) }& \textbf{Split Conformal Inference} \\
  \hline \resizebox{2.8in}{!}{\includegraphics{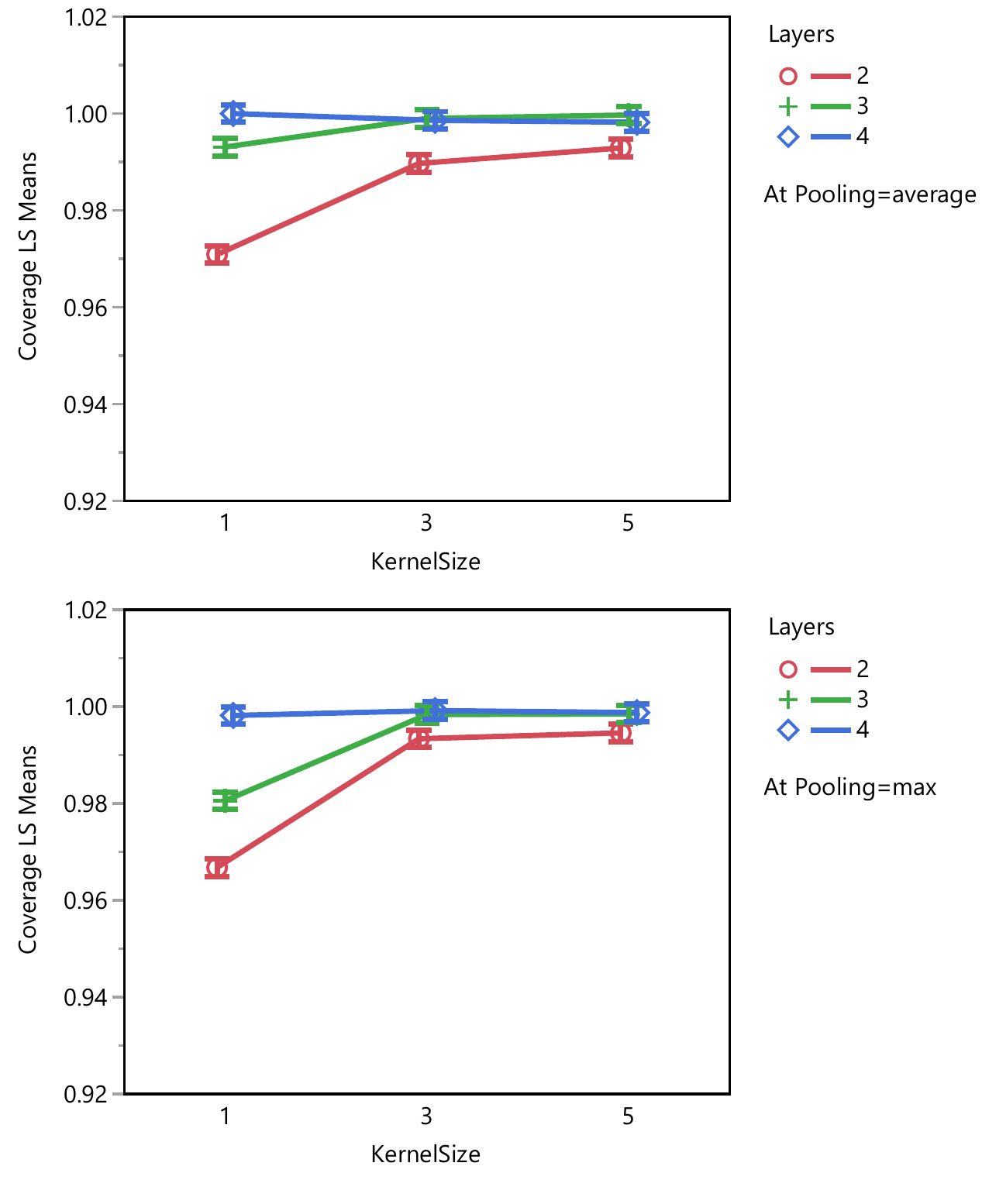}} & %
  \resizebox{2.8in}{!}{\includegraphics{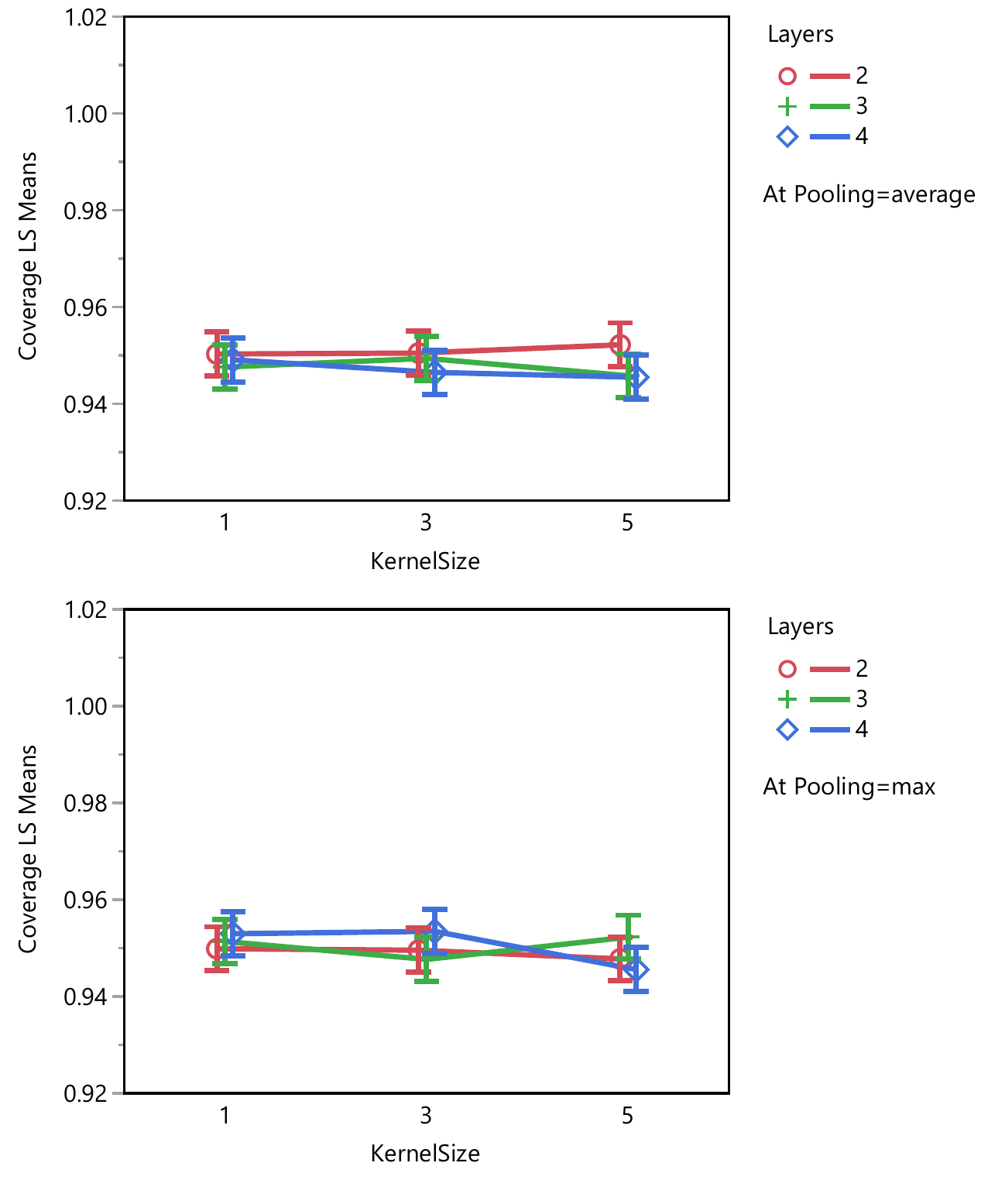}} \\
  \hline
\end{tabular}
\caption{PI Coverage in the RotNIST Data Set.}
\label{figure:RotNIST_PI_cov}
\end{center}
\end{figure}

Figure \ref{figure:boston_PI_cov} provides a comparison between the two PI methods for the Boston Housing data set. The significant differences in coverage among the activation functions can be observed for pivot bootstrap PIs, indicated by the non-overlapping error bars in the trend lines. Note too that these significant differences result in relatively conservative PIs; the empirical coverages are above the desired confidence level of 0.95. The analogous plot of split conformal inference PI coverage in Figure \ref{figure:boston_PI_cov} reveals no design choice in the network architecture which yields a statistically significant change in coverage. This pattern extends to the image-based RotNIST data set (Figure \ref{figure:RotNIST_PI_cov}), with the modeled PI coverage plotted by the relevant CNN hyperparameters. Graphical results of PI coverage for the remaining data sets are provided in \ref{appendix_a}. 

\begin{figure}[h]
 \centering
 \resizebox{6in}{!}{\includegraphics{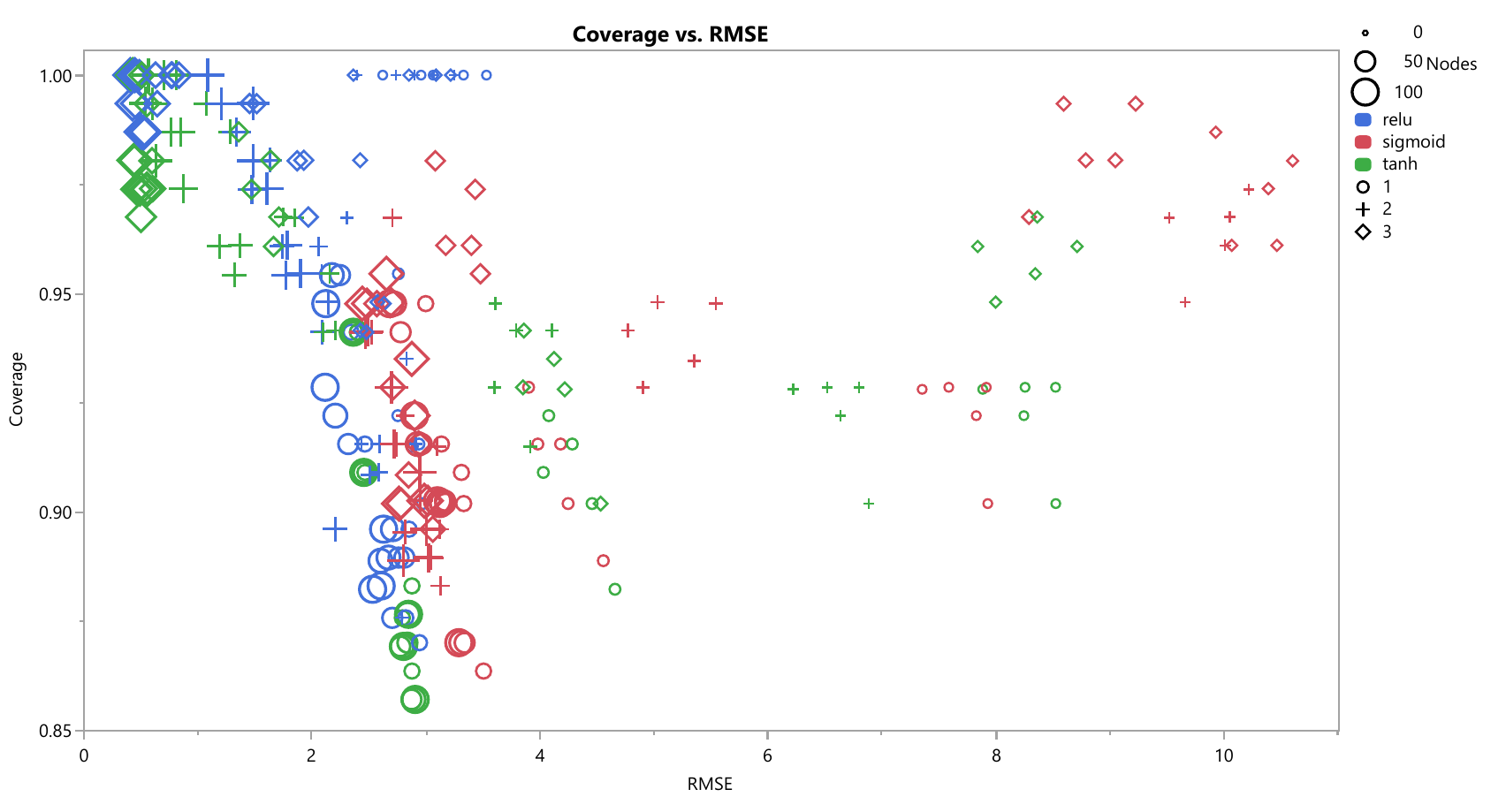}}
 \caption{PI Coverage versus RMSE for the Energy Efficiency Data Set.}
 \label{figure:EnergyEfficiency_Cov_vs_RMSE}
\end{figure}

The results presented here reveal that the coverage of the pivot bootstrap method is sensitive to the underlying neural network structure. The apparent mechanisms for this relationship are the fit of the network (measured by root mean squared error on the test set) and the response variable being modeled. Figure \ref{figure:EnergyEfficiency_Cov_vs_RMSE} provides an example plot of coverages of the bootstrap PIs against the test RMSEs from the networks used to construct the PIs. In this example from the Energy Efficiency data set, PI coverage is highest for networks at the extremes of observed RMSE. However, the relationship between coverage of the bootstrapped PIs and RMSE is not consistent across the response variables of the modeled data sets. Plots of bootstrapped PI coverage versus RMSE for other data sets are provided in \ref{appendix_c}.

Meanwhile, the split conformal inference method is generally insensitive to both structure and fit of the underlying networks. The PIs constructed with this method have an average coverage near the nominal level of 0.95, independent of any factors relating to the learning algorithm used. 

\subsubsection{Average Width}

A similar analysis approach is taken to examine PI average width as a function of neural network hyperparameter effects. In particular, full factorial models are built from the factors shown in Table \ref{table:DOE-benchmark} (for benchmark data sets) and Table \ref{table:DOE-RotNIST} (for RotNIST data set) to estimate pivot bootstrap and split conformal inference PI average widths. As before, these models include all main effects, as well as second- and third-order interactions. In the models for all eleven data sets, as well as for both pivot bootstrap and split conformal inference PIs, all modeled effects are statistically significant at $\alpha=0.05$. Moreover, for every data set, the constructed models had a $R^2$ of at least $0.98$, indicating they accounted for approximately 98\% of the observed variance in average width. 

\begin{figure}[h]
\begin{center}
\begin{tabular}{|c|c|}
  \hline \textbf{Pivot Bootstrap ($\bm{1,000}$ Resamples) }& \textbf{Split Conformal Inference} \\
  \hline \resizebox{2.8in}{!}{\includegraphics{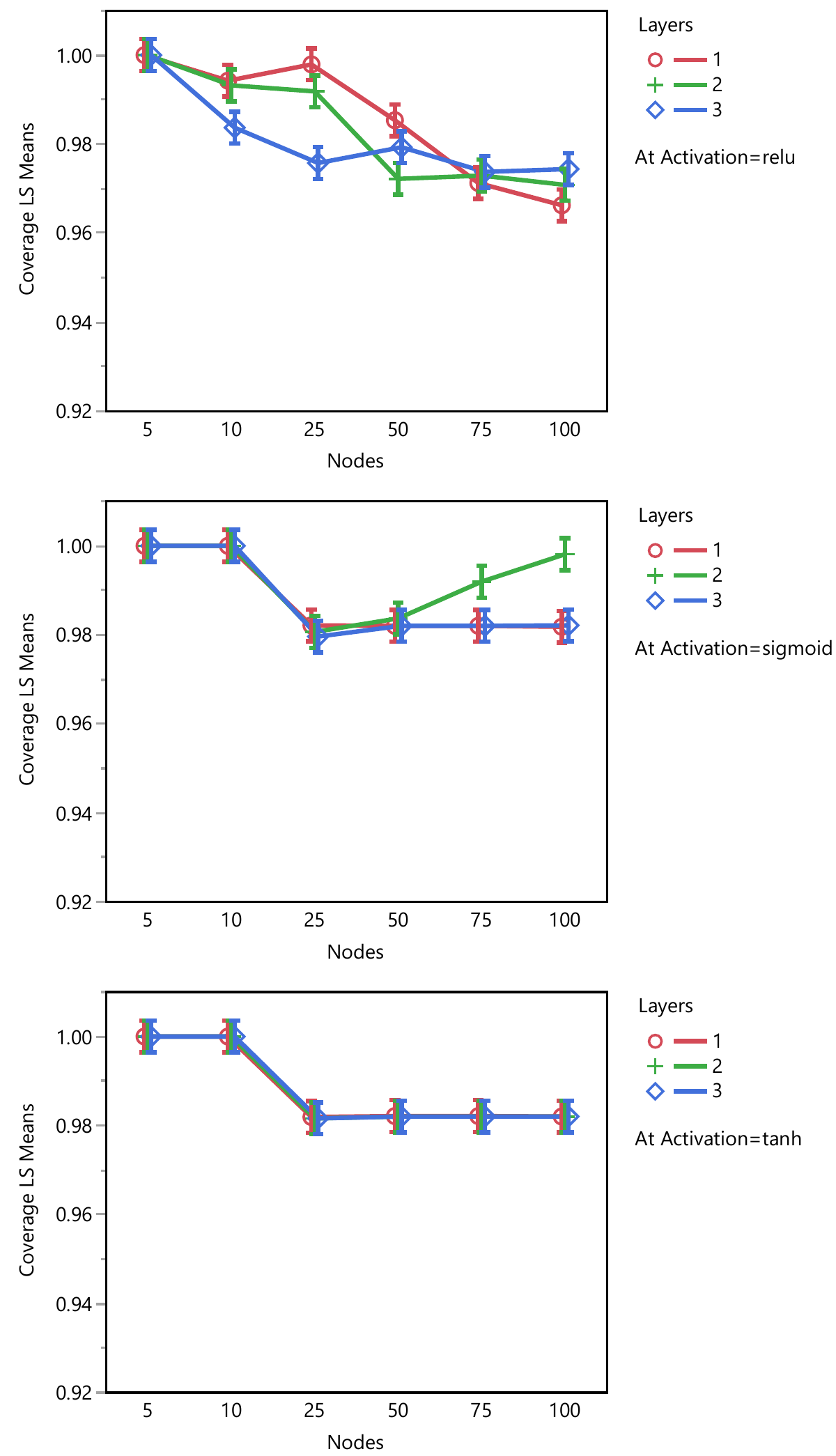}} & %
  \resizebox{2.8in}{!}{\includegraphics{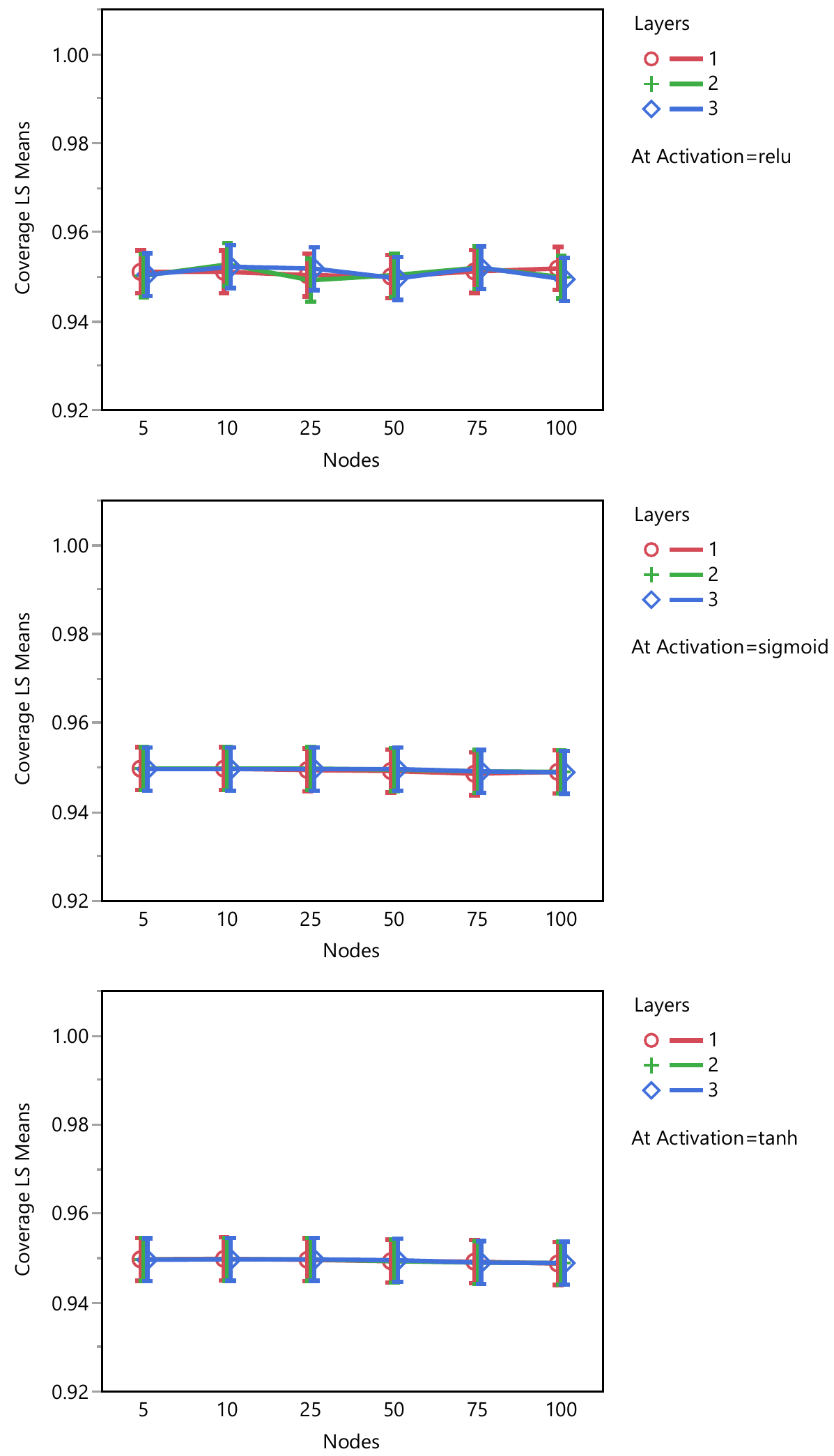}} \\
  \hline
\end{tabular}
\caption{PI Average Width in the Power Plant Data Set.}
\label{figure:PowerPlant_Width}
\end{center}
\end{figure}

In general, design choices which increase network capacity (i.e., ability to encode relationships between the features and target variable) result in better fitting networks (as measured by test RMSE) and smaller prediction intervals. Thus, networks having more layers or nodes will typically provide narrower PIs than networks having fewer. Similarly, networks utilizing the ReLU activation also provide an advantage in average width over those using Tanh or Logistic. However, these advantages often diminish as the size of the network grows. For example, Figure \ref{figure:PowerPlant_Width} for the Power Plant data set, shows the difference in average width between networks using five and ten nodes is generally larger than the difference between networks using 75 and 100. 

Similarly, the advantage of using three layers instead of two is insignificant regardless of the chosen activation for many data sets. In general, for the benchmark regression data sets, increasing network capacity beyond two layers and fifty nodes provides statistically insignificant changes in PI average width. Exceptions to this rule include the Energy Efficiency and Protein Structure data sets, which have target variables with more complex behaviors.

\begin{figure}[h]
\begin{center}
\begin{tabular}{|c|c|}
  \hline \textbf{Pivot Bootstrap ($\bm{1,000}$ Resamples) }& \textbf{Split Conformal Inference} \\
  \hline \resizebox{2.8in}{!}{\includegraphics{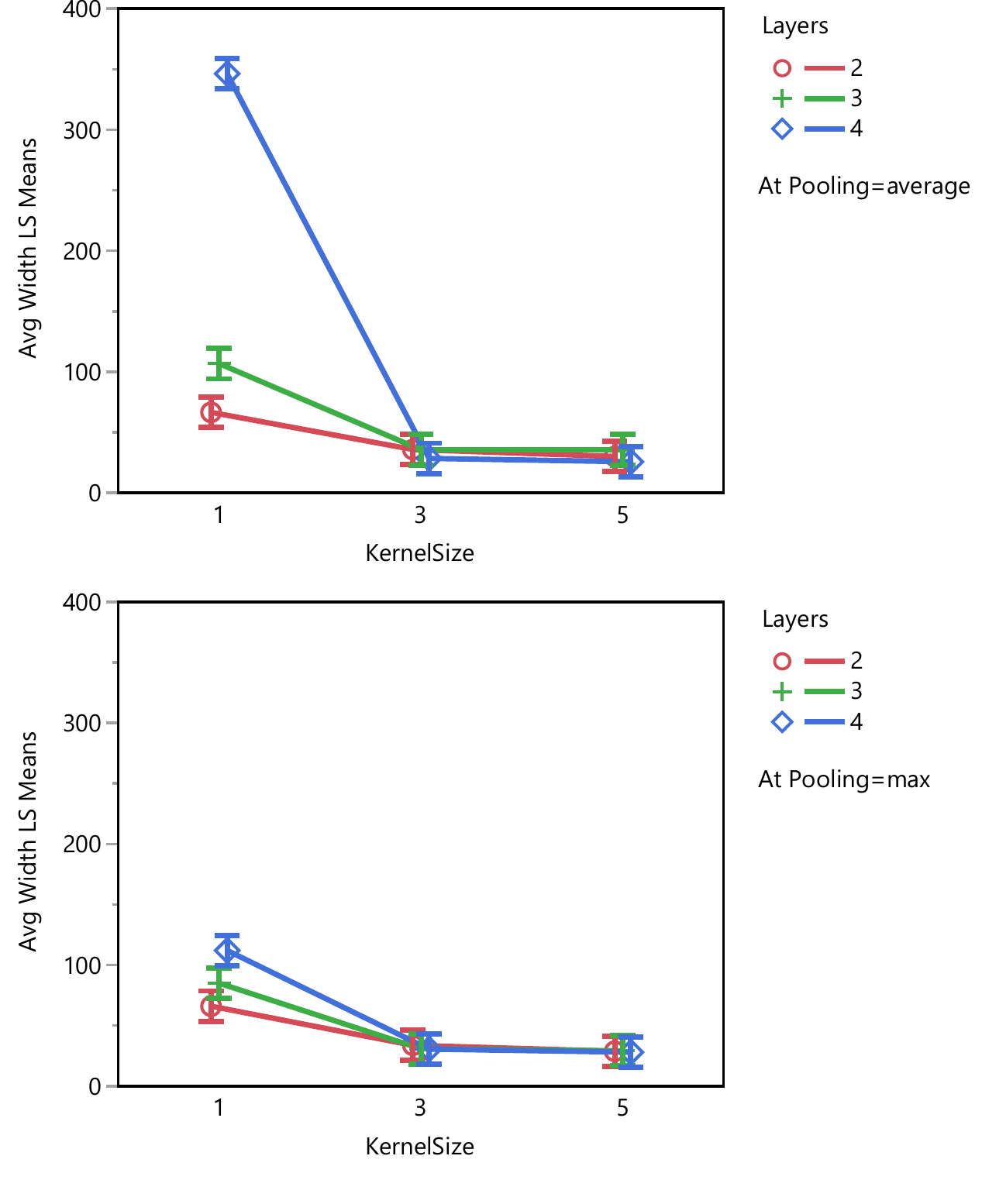}} & %
  \resizebox{2.8in}{!}{\includegraphics{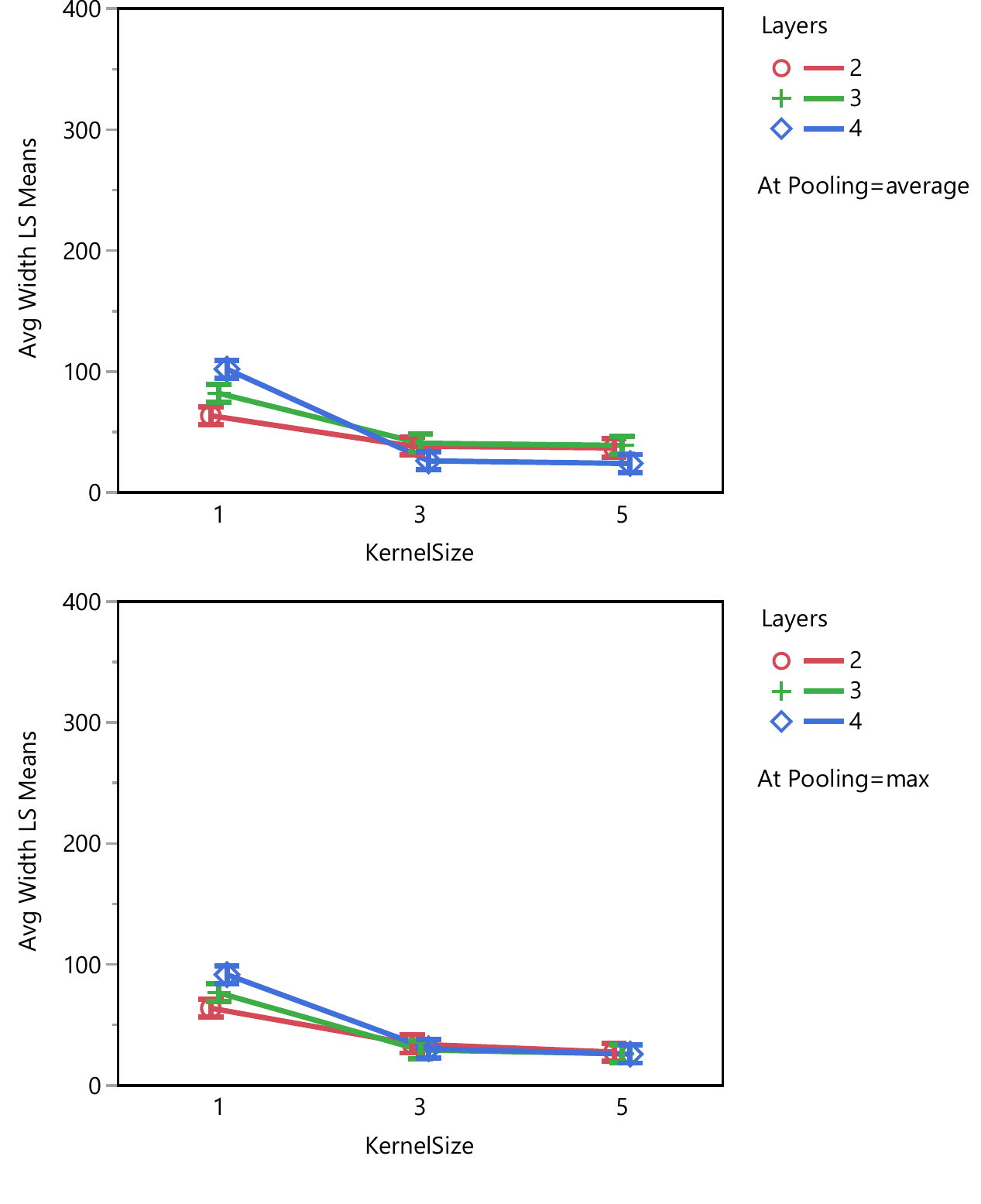}} \\
  \hline
\end{tabular}
\caption{PI Average Width in the RotNIST Data Set.}
\label{figure:RotNIST_PIAvgW}
\end{center}
\end{figure}

Focusing solely on the CNNs constructed for RotNIST data set (Figure \ref{figure:RotNIST_PIAvgW}), a similar pattern in the effects of their hyperparameters emerges. The network's prediction performance measured by RMSE and the average widths of the corresponding PIs improve significantly as layers increases from two to three, and as kernel size increases from $1\times1$ to $3\times3$. Other changes in hyperparameters yield less noticeable and generally insignificant differences in width. Graphical results of PI average width for the remaining data sets are provided in \ref{appendix_b}.

However, the results presented thus far should not be interpreted to say that arbitrarily increasing network size will always result in better PI performance. Rather, in select data sets, increasing network capacity yields poor performing PIs. The experimental results reveal that in none of the eleven data sets modeled did the network (split conformal inference method) or network ensemble (pivot bootstrap method) minimizing RMSE also produce the most efficient set of PIs (that also maintained coverage). Thus, optimal neural network performance does not necessarily translate to optimal average widths for its corresponding PIs. These results suggests that further heuristics in the optimization of neural networks--beyond, say, monitoring RMSE on a validation set--are needed if the end goal of the network is to provide efficient PIs. 

\subsubsection{Other Results and Discussion}

The empirical results presented thus far illustrate how the choice of neural network hyperparameters affect PI performance. Most notably, the coverage of the pivot bootstrap PIs are sensitive to changes in network architecture. Larger networks, or more generally, those that are able to better learn the training data, lead to bootstrapped PIs with higher coverage than smaller, poorer fitting networks. Conversely, the coverage of split conformal inference PI sets remain close to the nominal level regardless of how the size or fit of the neural network used for regression learning, which is not surprising.

Moreover, PI average width is closely related to the fit of the neural network. This is evidenced in summary plots for each data set: PI average width and test set RMSE move in tandem with one another. Despite this persistent relationship between PI average width and RMSE, it is not necessarily the case that the neural network (or ensemble of networks) which provide the minimum test set RMSE also provide the PIs with the smallest average widths, particularly for the networks constructed in conjunction with the bootstrap PI method. 

The exact mechanism for this somewhat contradictory result is unknown. Further exploration is needed to understand the relationship between the fit of the network and the distribution of its resulting residuals, especially as the former improves from highly biased to minimally bias.

Table \ref{table:step1-results} provides network architecture for each data set producing the optimal PIs, along with their average width and the PI algorithm used. Split conformal inference produces the optimal set of PIs for all but one of the data set. This result is driven largely by the conservative nature of the bootstrap method; the result is PIs whose coverage is much higher than nominal. The difference in overall performance between the split conformal inference method and pivot bootstrap method is large, with the former providing roughly nominal coverage while being, on average, 3.4\% narrower. A more detailed comparison of PI performance is provided in Section \ref{PI_comparison}. The lone exception to this rule is seen in the Year Prediction data set. The PIs of this data set generally large (usually more than half of the original target range from the training set); the lack of precision in these estimates may mean that the relative advantage of using one PI method over another is negligible.

\begin{table}[h]
\begin{center}
\begin{tabular}{|p{3cm}| c | c | c | c | c|}
  \hline \multirow{2}{3cm}{\textbf{Benchmark Data Sets}} & \multicolumn{3}{c|}{\textbf{MLP Settings}} & \multirow{2}{1.5cm}{\textbf{Method}} & \multirow{2}{1.3cm}{\textbf{Width}} \\
  \cline{2-4}
  & \textbf{Layers} & \textbf{Nodes} & \textbf{Activation} & &  \\
  \hline Boston Housing & 3 & 100 & ReLU & Split Conformal & 13.135 \\
  \hline Wine Quality & 2 & 50 & ReLU & Split Conformal & 2.560 \\
  \hline Concrete Strength & 3 & 100 & ReLU & Split Conformal & 24.836 \\
  \hline Energy Efficiency & 3 & 100 & Tanh & Split Conformal & 4.243 \\
  \hline Kinematics & 3 & 75 & Tanh & Split Conformal & 0.303 \\
  \hline Naval Propulsion & 3 & 100 & ReLU & Split Conformal & 0.126 \\
  \hline Power Plant & 3 & 50 & ReLU & Split Conformal & 16.005 \\
  \hline Protein Structure & 3 & 50 & Tanh & Split Conformal & 18.218 \\
  \hline Yacht Hydrodynamics & 3 & 100 & ReLU & Split Conformal & 4.560 \\
  \hline Year Prediction MSD & 1 & 50 & ReLU & Pivot Bootstrap & 59.917 \\
  \hline \multirow{2}{3.3cm}{\textbf{Image-Based Data Set}} & \multicolumn{3}{c|}{\textbf{CNN Settings}} & \multirow{2}{1.5cm}{\textbf{Method}} & \multirow{2}{1.3cm}{\textbf{Width}} \\
  \cline{2-4}
  & \textbf{Layers} & \textbf{Kernel Size} & \textbf{Pooling} & & \\
  \hline RotNIST & 3 & $5\times5$ & Maximum & Split Conformal & 26.170 \\
  \hline
\end{tabular}
\caption{Optimal neural network hyperparameter settings found for each data set. Also shown is the PI method producing the optimally performing PIs with their corresponding average width.}
\label{table:step1-results}
\end{center}
\end{table}

Considering the observed PI performance in coverage, width, as well as the potential effects of over-fitting, the recommended benchmark neural network structure to begin analysis of data set is an architecture consisting of two hidden layers, with 50 nodes in each, and employing the ReLU activation. Additionally, single-layer networks generally do not provide enough opportunity for the network to map the feature space to the response effectively. Specifically, that lack of depth must be overcome by prolonging the training time or by adding nodes to the structure. A similar modification must also be implemented if using Tanh or Sigmoid activation function, as these functions require more parameters and longer training times to effectively learn the patterns in the data when compared to the ReLU activation. For data sets containing more complicated patterns (e.g., the Energy Efficiency data set) increasing network capacity—through the addition of layers or nodes—will improve both metrics. 

An analogous approach can be taken with CNNs. Based on the results from the RotNIST data set, deeper networks (three or four hidden layers) provide a marked improvement in PI and network point prediction performance as compared to CNNs having just two hidden layers. In the relatively limited design of architectures examined here, use of $3\times3$ and $5\times5$ convolution kernels provided a similar advantage over $1\times1$ kernels. However, design space of CNN depth and kernel size remains an area of active research (\citealt{simonyan-zisserman:2014}, \citealt{howard-et-al:2017}).

\subsection{Comparison of PI Methods}\label{PI_comparison}

Table \ref{table:step2_width} provides the computed average widths for each method implemented across each data set (standard errors in parentheses) from the second experiment. Table \ref{table:step2_coverage} similarly displays coverage values, which are evaluated according to their closeness to the nominal coverage level of 0.95. To enable easier interpretation of the average width results, the values are normalized by data set. In particular, every average width is expressed as a ratio to the narrowest PIs produced for that data sets with the most efficient PIs for each data set having a value of 1.00. These values are shown in Table \ref{table:step2_rel_eff}.

\begin{table}[p]
 \centering
 \resizebox*{!}{8.25in}{%
    \rotatebox{270}{%
        \includegraphics{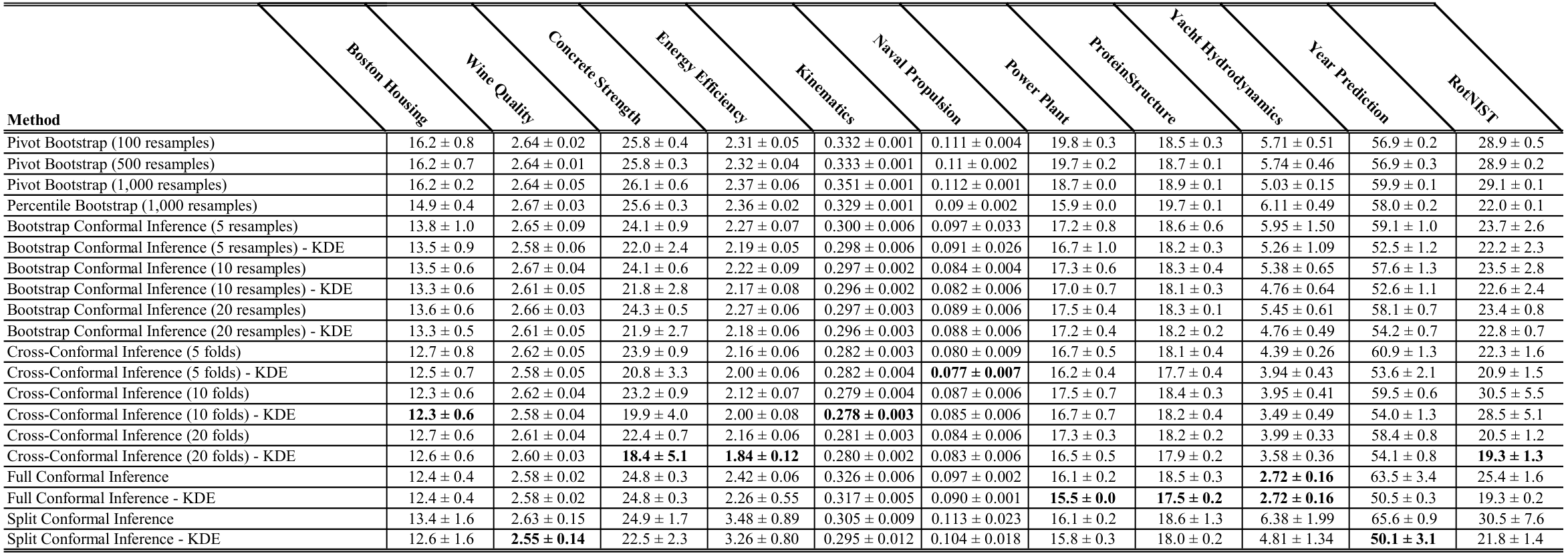}}}
 \caption{Average widths of PI methods with calculated standard errors.}
 \label{table:step2_width}
\end{table}

\begin{table}[p]
 \centering
 \resizebox*{!}{8.25in}{%
    \rotatebox{270}{%
        \includegraphics{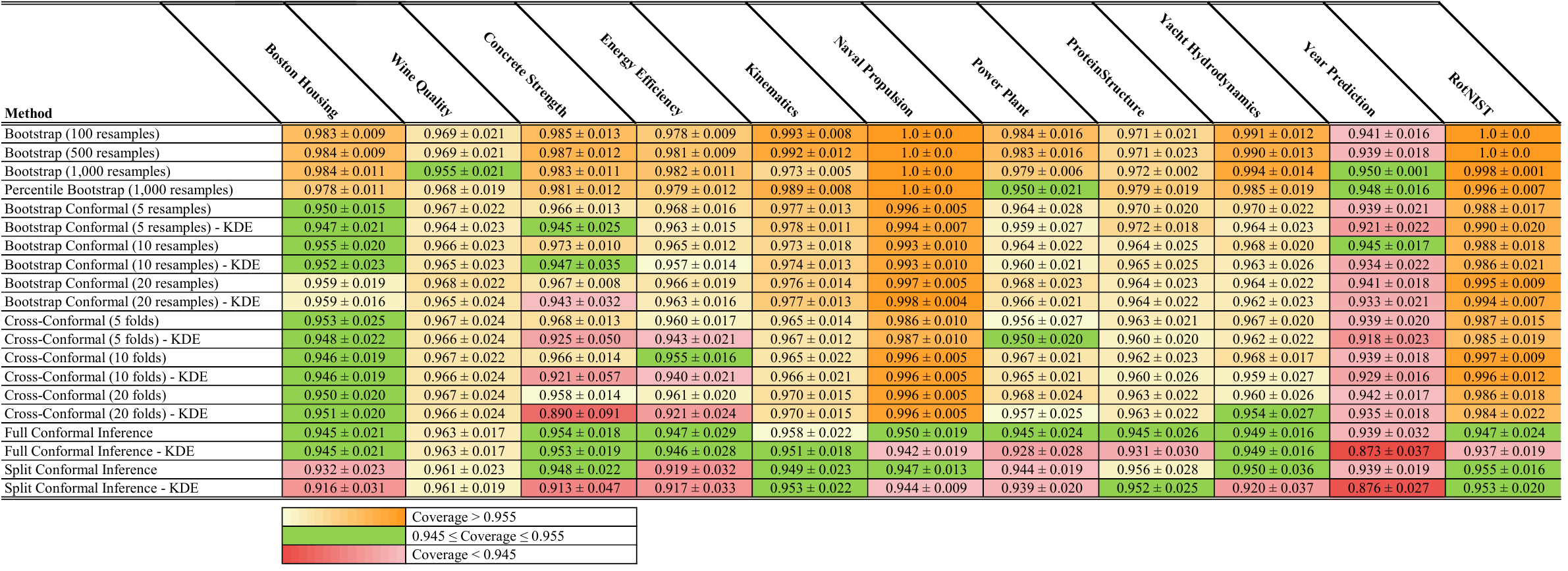}}}
 \caption{Average coverage of PI methods with calculated standard errors.}
 \label{table:step2_coverage}
\end{table}

\begin{table}[p]
 \centering
 \resizebox*{!}{8.25in}{%
 \rotatebox{270}{%
 \includegraphics{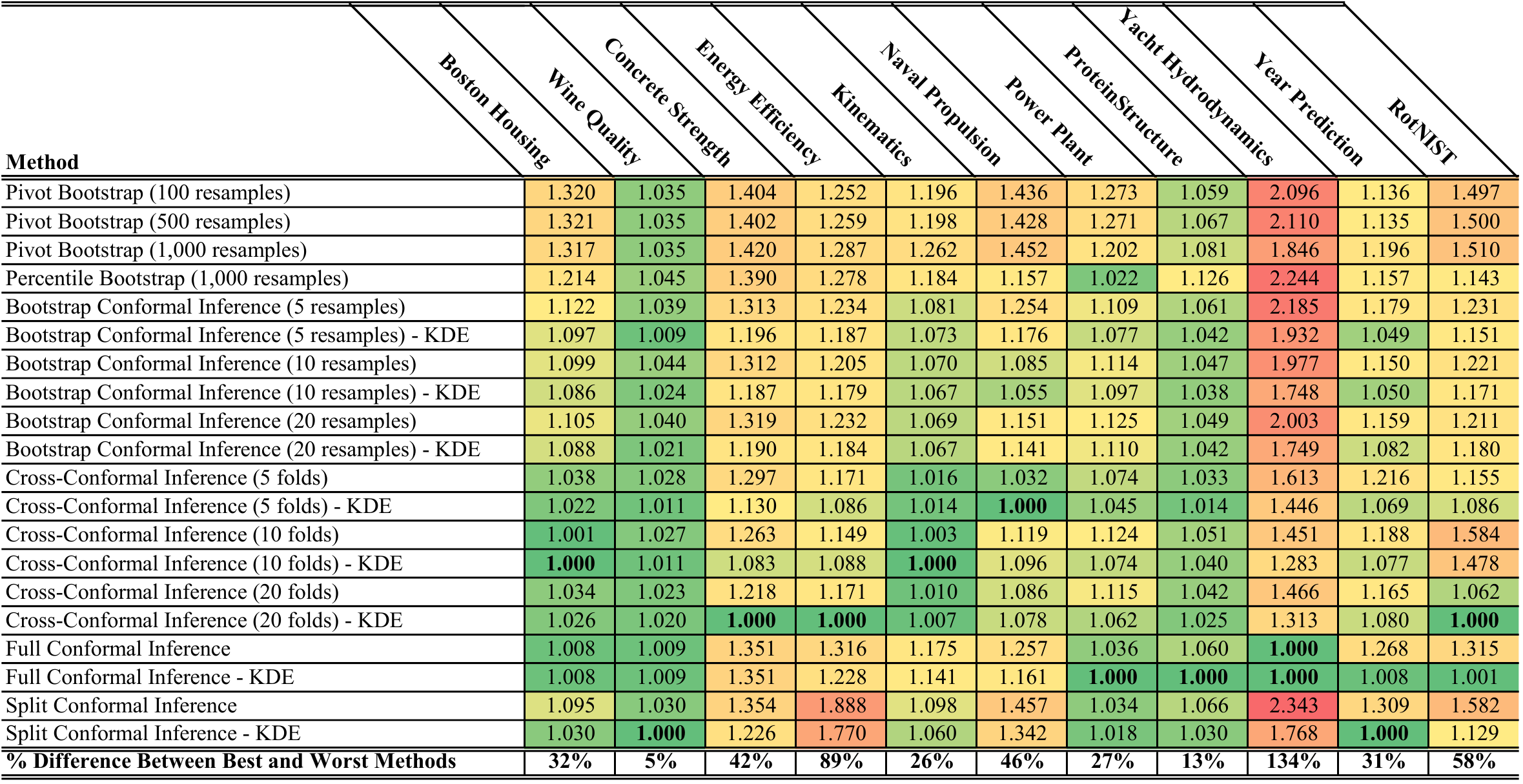}}}
 \caption{Relative efficiencies of PI methods.}
 \label{table:step2_rel_eff}
\end{table}

For all eleven data sets examined, using a CI method with KDE yields the most efficient PIs. Across all CI methods and data sets, the use of KDE improves PI efficiency by an average of 5.8\%. When examining Table \ref{table:step2_rel_eff}, cross-CI with $K=20$, appears to be the most efficient method overall. While the full CI method does produce the most efficient set of PIs for three data sets, its performance varies substantially in other data sets. Note that the bootstrap methods produce generally the least efficient PIs compared to the CI methods.

Supplementing CI method with KDE further improves their efficiency, however in certain cases this comes at the cost of unacceptable variance or coverage. In particular, the narrowest PIs for the Concrete Strength and Energy Efficiency data sets, each produced using cross-CI ($K=20$) with KDE, also have coverages well below the nominal level of 95\%. Moreover, these PIs also the highest variance in coverage and average width across test sets when compared to the other methods. The extra computation time required for KDE varies by data set, increasing as the number of training samples grows. For the larger data sets--in particular, the Year Prediction data set--this extra time is substantial. The average 5.8\% improvement in efficiency suggests that the additional computation time remains worthwhile in most cases so long as coverage is acceptable. If time or computational resources are limited, work-arounds such as allowing non-zero error in the KDEs \citep{VanderPlas:2013}, or reducing the number of folds implemented in cross-CI, can be pursued with minimal degradation in PI performance. 

In terms of validity, full and split CI provide the PIs whose coverage most closely match the nominal level of 95\%. The PIs of the aggregated CI methods are generally conservative, although the implementing with KDE reduces coverage (in some cases below 95\%, as discussed). Bootstrapped PIs, regardless of the number of resamples, produce coverages exceeding 98\% for many data sets, as seen previously in the first experimental step. 

Across test sets, the bootstrap method provides the most stable PI performance, both in average width and coverage, while the set of CI methods exhibit higher levels of variation. Among the CI methods, the full conformal inference method has the lowest variance--an intuitive result following from its transductive nature. Using aggregated conformal predictors greatly reduces the variance in PI performance across test sets as compared to the split conformal inference method. CI methods implemented with KDE have higher variance in coverage and average width than those implemented using the absolute residual as the nonconformity measure. 

When considering aggregate and variable PI performance, both across data sets and replicates, cross-CI seems to provide the highest quality PIs. While implementing any number of folds appears adequate, using a relatively large number such as $K=20$ appears to provide more stable results. Even so, cross-CI has a significantly smaller computational burden when compared to bootstrapping approaches and the full conformal inference method. Split CI has a potentially unacceptable degree of variance in its performance--induced by generating one random split in the data--while also having sub-optimal efficiency. 

\subsubsection{Case Study: Conditional Coverage}

In addition to the PI performance metrics discussed in previous sections, an additional area in which to evaluate PI methods is their conditional coverage. In the methods and results so far, coverage has been calculated and discussed in marginal terms (i.e., across the entire test set). Conditional coverage is calculated separately across subsets of the feature space. Neither the bootstrap nor the set of conformal inference methods (in the simple implementations utilized here) guarantee conditional validity \citep{lei-wasserman:2014}. Regardless, a practical consideration for constructing PIs is how well a $1-\alpha$ PI method maintains nominal coverage when examining a particular subset of the feature space.
 	
As a case study, the conditional coverage of three separate PI methods were calculated for combinations of features in the Boston Housing and Energy Efficiency data sets. The selected methods were the pivot bootstrap (500 resamples), the split CI method and the cross-CI method ($K=20$). Table \ref{table:cond-cov-boston} displays the conditional coverage of these methods for each value of the binary variable “Charles River” in the Boston Housing data set. Values of one or zero correspond to whether the neighborhood borders the river or does not border the river, respectively. The means of the median home price corresponding to both cases are included in the table for reference. The conditional coverage of the bootstrap method exceeds 95\% regardless of the value of the Charles River variable. However, the split CI and cross-CI predictors struggle to correctly predict median home price for neighborhood bordering the Charles River. The split CI method has a coverage rate of 79.0\% for such homes, while cross-CI has a rate of 86.4\%. Since homes in these neighbors tend to command a higher price, incorrect valuations of the home prices carry more cost. 
\begin{table}[h]
\begin{center}
\begin{tabular}{|r| c | c|}
  \hline \textbf{Charles River}& \textbf{0} & \textbf{1} \\
  \hline Count & 919 & 81 \\
  Mean Home Price (\$1000's) & 22.236 & 30.868 \\
  \hline Pivot Bootstrap (500 resamples) & 0.9869 & 0.9506 \\
  Cross-Conformal (20 folds, no KDE) & 0.9675 & 0.7901 \\
  Split Conformal (no KDE) & 0.9445 & 0.8642 \\
  \hline
\end{tabular}
\caption{Conditional coverage of select PI methods according to value of Charles River variable in Boston Housing data set.}
\label{table:cond-cov-boston}
\end{center}
\end{table}

A similar exercise is conducted for the Energy Efficiency data set. However, instead of examining conditional coverage given the value of a single feature, conditional coverage is computed across bins of the predicted target value: 0-10, 10-20, 20-30, 30-40, 40-50 (the chosen target variable for this data set is heating load, measured in British Thermal Units).

The results are summarized in Table \ref{table:cond-cov-energy}. Again, the pivot bootstrap method maintains coverage of at least 95\% in every bin of values for the ensemble’s predicted value. The coverages of the CI methods fall below 95\% for the predicted heating loads between 20 and 40 BTUs. While deficiency is small for cross-CI, the split conformal inference method in particular struggles to predict this middle range of the target variable. When the neural network trained in conjunction with the split CI method predicts a heating load between 20 and 30 BTUs, the method correctly predicts the range of the true heating load in 80.6\% of cases. 

\begin{table}[h]
\begin{center}
\begin{tabular}{|l| c  c c c c|}
  \hline \textbf{Predicted Heating Load (BTUs)}& \bm{$<10$} & \bm{$10-20$} & \bm{$20-30$} & \bm{$30-40$} & \bm{$>40$} \\
  \hline Pivot Bootstrap (500 resamples) & 1.000 & 0.9892 & 0.9914 & 0.9563 & 0.9500 \\
  Cross-Conformal (20 folds, no KDE) & 1.000 & 0.9871 & 0.9397 & 0.9301 & 0.9250 \\
  Split Conformal (no KDE) & 1.000 & 0.9806 & 0.8060 & 0.8901 & 0.9500 \\
  \hline
\end{tabular}
\caption{Conditional coverage of select PI methods according to predicted value of the target variable in the Energy Efficiency data set.}
\label{table:cond-cov-energy}
\end{center}
\end{table}

\begin{figure}[h]
 \centering
 \resizebox{6in}{!}{\includegraphics{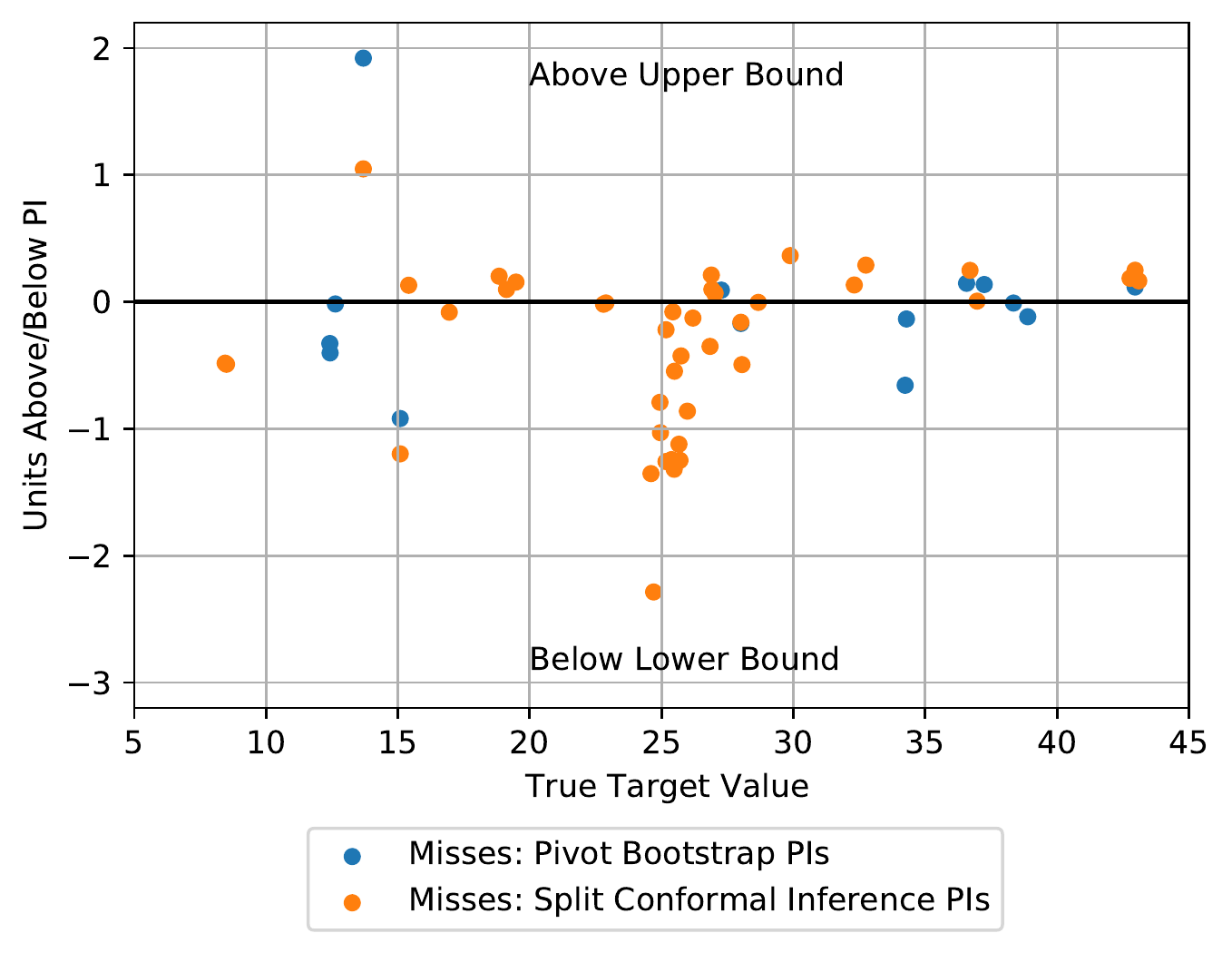}}
 \caption{Incorrectly predicted test observations in the Energy Efficiency data set.}
 \label{figure:EnergyEfficiency_misses}
\end{figure}

A further investigation into the performance of these intervals reveals a clear pattern into the incorrect predictions of the split CI PI method. As seen in Figure \ref{figure:EnergyEfficiency_misses}, the split CI PIs routinely over-predict the target value when the prediction is between 20 and 30 BTUs (incorrect predictions from pivot bootstrap method provided for reference). The lack of balance in the coverage of these intervals—that is, the relatively low left coverage rate compared to right coverage—suggests either some skewness in the distribution of the target \bm{$y$}, or the underlying neural network is not fully learning the relationship between the features \bm{$X$} and \bm{$y$}. Coverage balance is sometimes a consideration when constructing PIs, as providing an accurate lower or upper bound for an unknown outcome may be important in certain settings. 

In general, the conservativeness of the pivot bootstrap method, as measured by its marginal coverage, appears to be a benefit when examining these conditional cases. The additional marginal coverage—resulting in relatively inefficient PIs—effectively provides the slack to maintain conditional coverage at a rate around $1-\alpha$. The use of an aggregated conformal predictor, already noted for their slightly conservative marginal coverage, appears to be the best among the conformal predictors for optimizing conditional coverage. Alternate methods, such as Mondrian conformal predictors \citep{vovk_et_al:2005}, or ``locally-weighted" CI \citep{lei-et-al:2018}, can be implemented to account for distributional differences in the covariate and response spaces, enabling more consistent coverage.

\section{Conclusion}

We investigated the relationship between various neural network architectures and PI performance. Additionally, we compared two common approaches for constructing PIs: bootstrapping and conformal inference. Overall, PI performance, especially in terms of efficiency, improves as the fit of the underlying neural network improves. Thus, using the ReLU activation function in the neural network—which generally results in a better fitting network as opposed to Sigmoid or Tanh when given equal amount of training time—and a sufficient number of layers (at least two) and nodes (at least 50) will result in better performing PIs. Analogously, ensuring network capacity is sufficient (three or more hidden layers and $3\times3$ convolutional kernels) when constructing CNNs, specifically in the number of layers and kernel size, will result in satisfactorily performing intervals. 

However, as with point prediction, care should be taken to avoid extensive levels of over-fitting. Over-fitting can result from networks having more than sufficient capacity for the data set, or from overly-long training times. The efficiency of the pivot bootstrap PI method, in particular, appears to be prone to such issues since the effect of over-fitting is masked by the ensemble’s goodness-of-fit metrics (such as RMSE). This leads to an otherwise contradictory result; where the best fitting architectures can yield sub-optimal sets of PIs. 

For practical implementations, cross-conformal inference is flexible enough to conform to the resources at hand. If resources are limited, or if the training of more than a few networks is not feasible, using the cross-conformal inference method with a small number of folds is likely sufficient. If the resources are not constrained, then using a larger number of folds (e.g., 20) appears to provide the best results, both in average metrics such as width and coverage, and in their variance across test sets. 

Opportunities for future work along these lines of research include an extension into predicting categorical responses. The research herein focused solely on data sets having a continuous-valued target variable; in this way, accurate comparisons were made across multiple methods for metrics such as average width. Also interesting is further exploration of the cross-conformal inference method. This research was broad in its scope, seeking to evaluate and compare a suite of PI methods. As such, there was limited time to experiment with the different parameters of each PI method, such as the number of folds for cross-conformal inference. The relationship between PI performance and the number of folds implemented is not clear from the results here.


\acks{
We would 
like to thank the Air Force Research Laboratory (RHWAI) for supporting this research.}

\vskip 0.2in
\bibliography{bibliography.bib}

\appendix 
\gdef\thesection{Appendix \Alph{section}}

\section{PI Coverage Plots} \label{appendix_a}

\begin{figure}[H]
\begin{center}
\begin{tabular}{|c|c|}
  \hline \textbf{Pivot Bootstrap ($\bm{1,000}$ Resamples) }& \textbf{Split Conformal Inference} \\
  \hline \resizebox{2.8in}{!}{\includegraphics{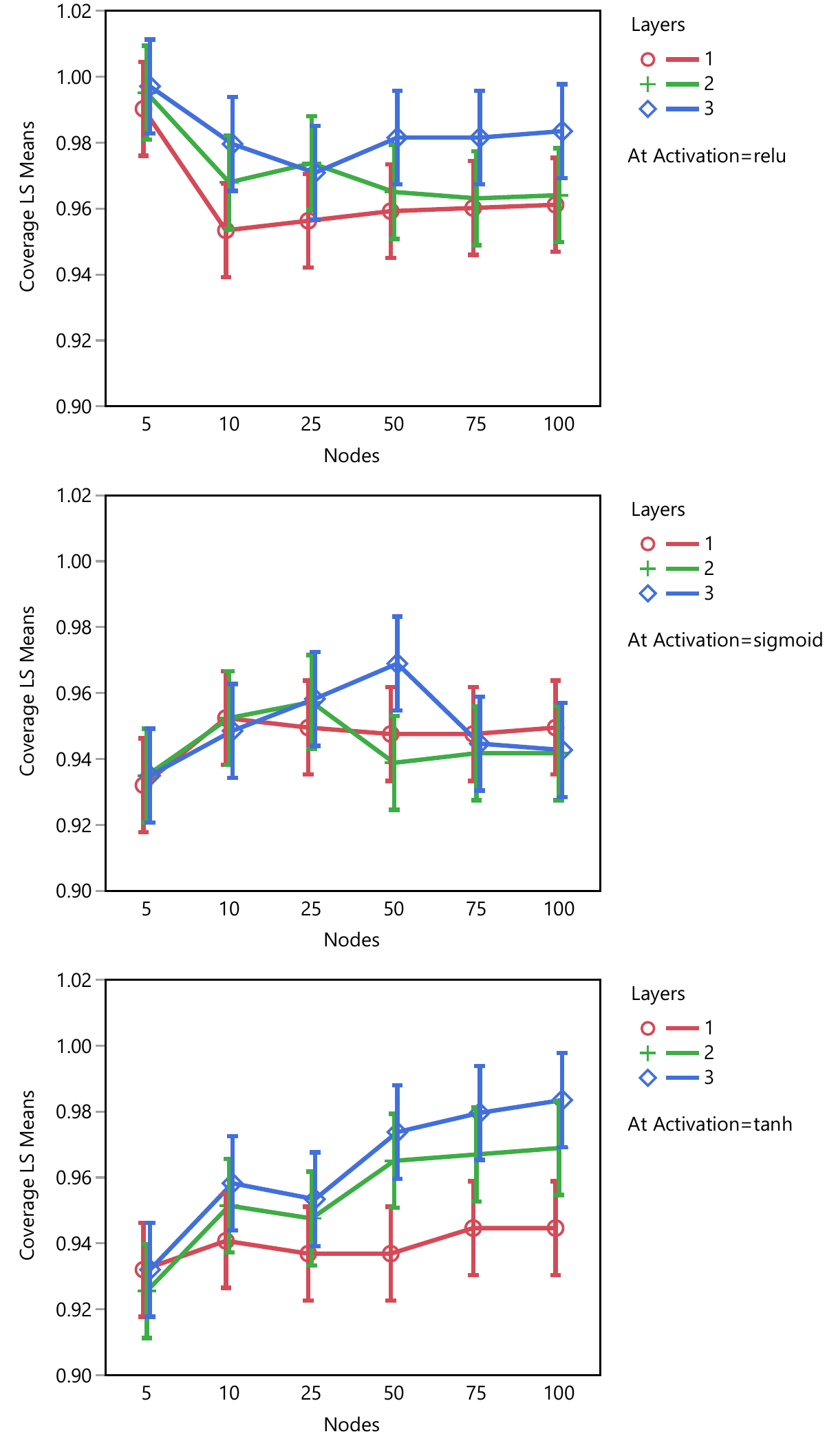}} & %
  \resizebox{2.8in}{!}{\includegraphics{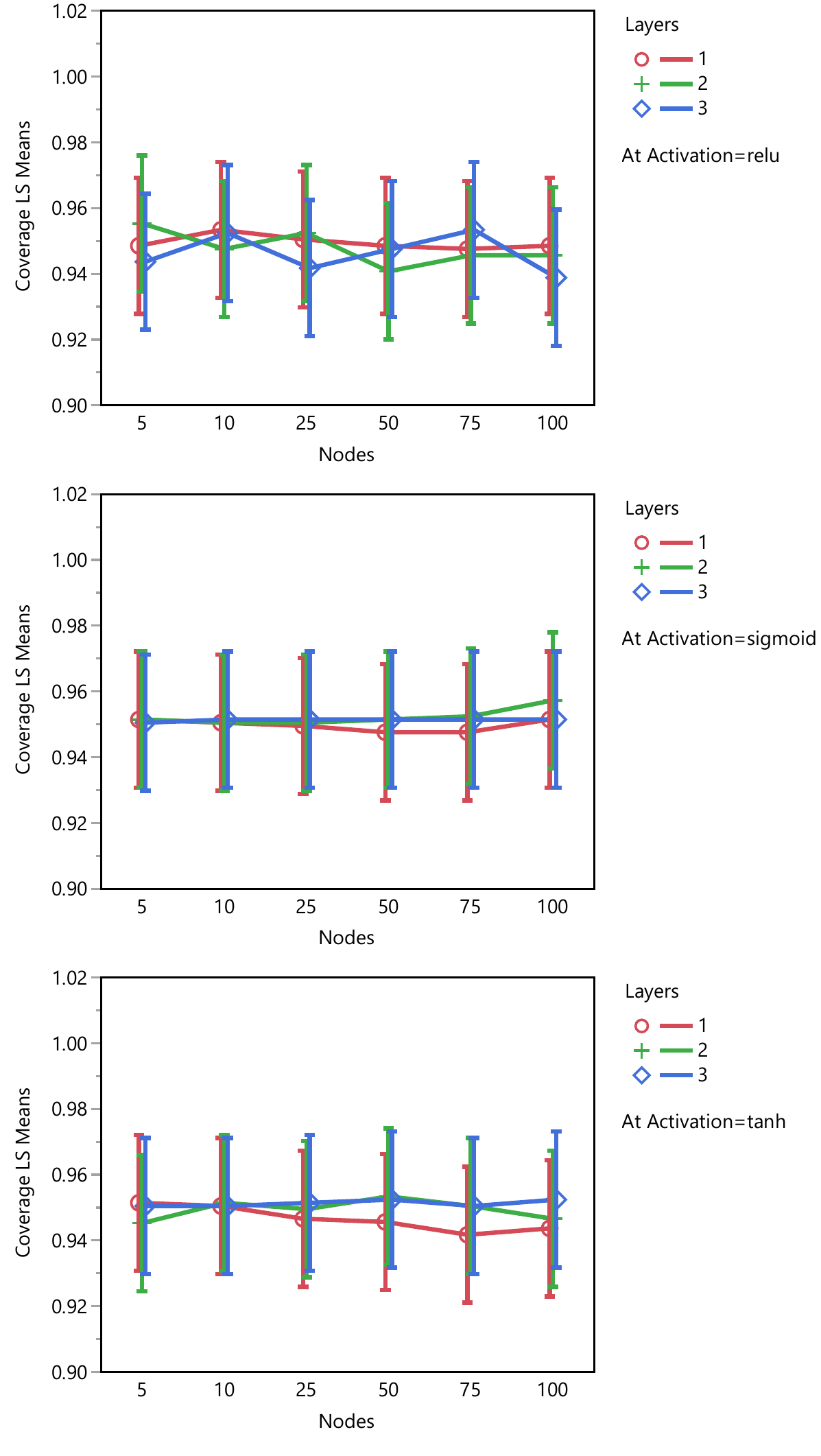}} \\
  \hline
\end{tabular}
\caption{PI Coverage in the Concrete Strength Data Set.}
\label{figure:ConcreteStrength_PICov}
\end{center}
\end{figure}

\begin{figure}[H]
\begin{center}
\begin{tabular}{|c|c|}
  \hline \textbf{Pivot Bootstrap ($\bm{1,000}$ Resamples) }& \textbf{Split Conformal Inference} \\
  \hline \resizebox{2.8in}{!}{\includegraphics{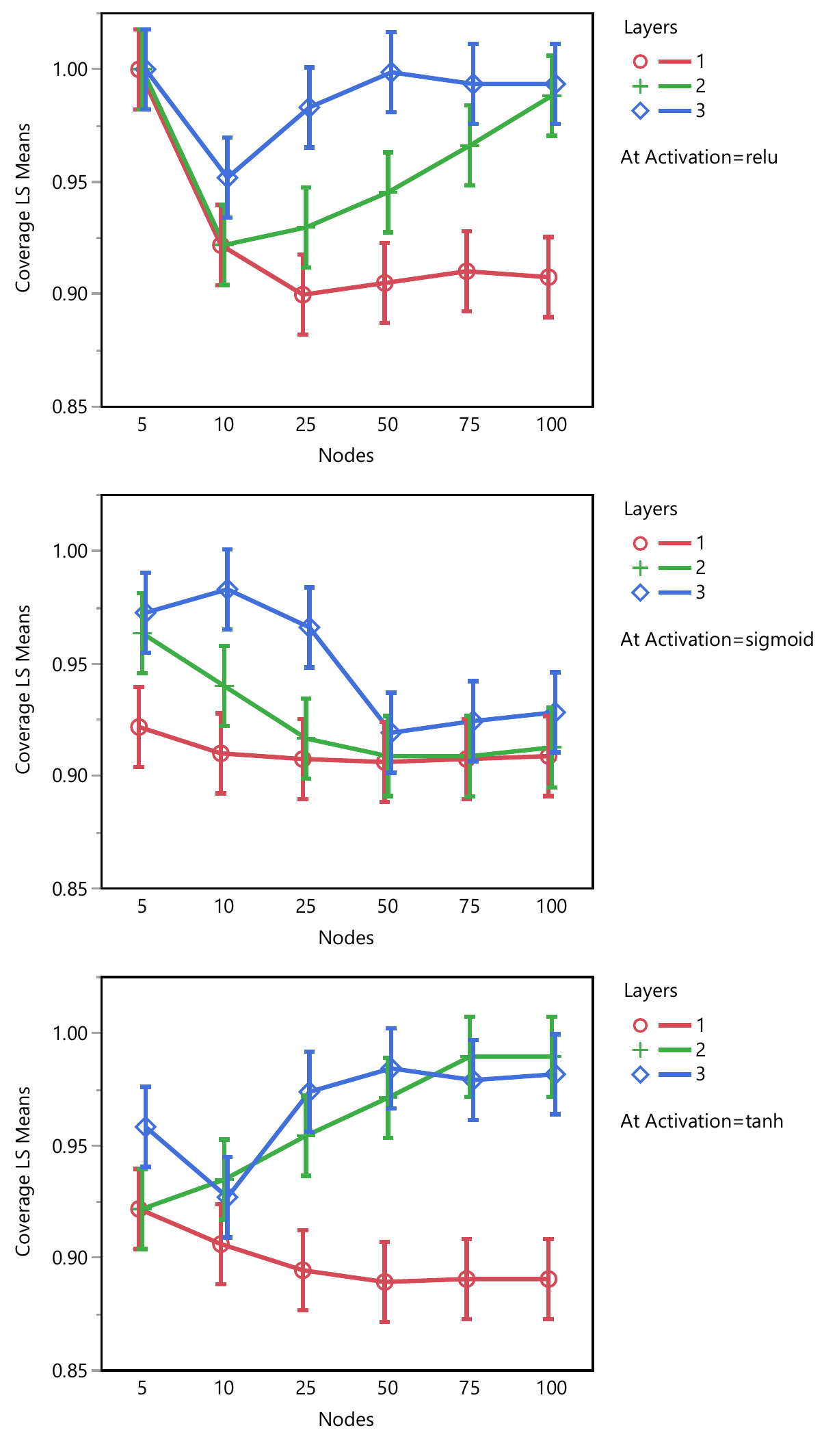}} & %
  \resizebox{2.8in}{!}{\includegraphics{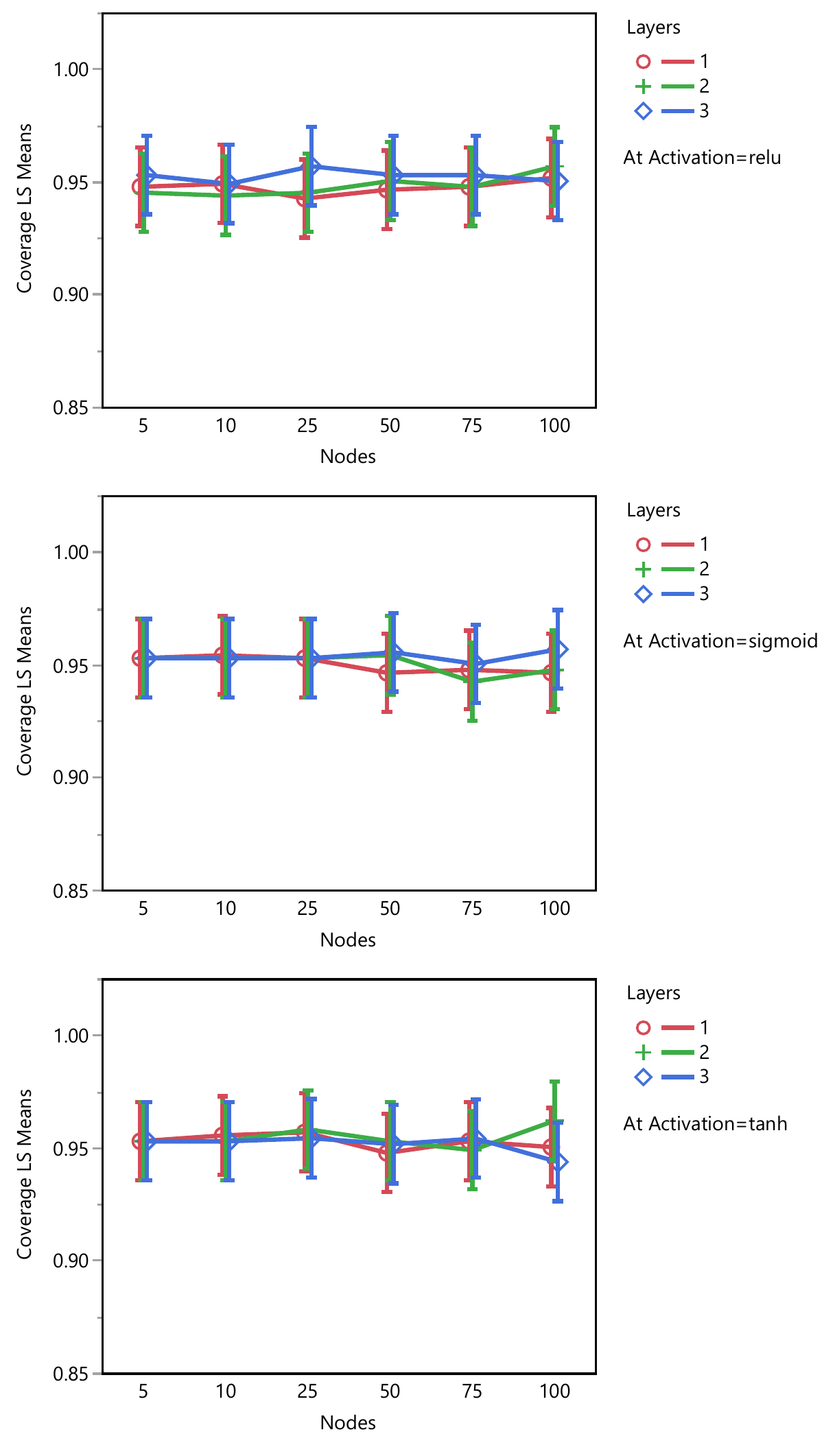}} \\
  \hline
\end{tabular}
\caption{PI Coverage in the Energy Efficiency Data Set.}
\label{figure:EnergyEfficiency_PICov}
\end{center}
\end{figure}

\begin{figure}[H]
\begin{center}
\begin{tabular}{|c|c|}
  \hline \textbf{Pivot Bootstrap ($\bm{1,000}$ Resamples) }& \textbf{Split Conformal Inference} \\
  \hline \resizebox{2.8in}{!}{\includegraphics{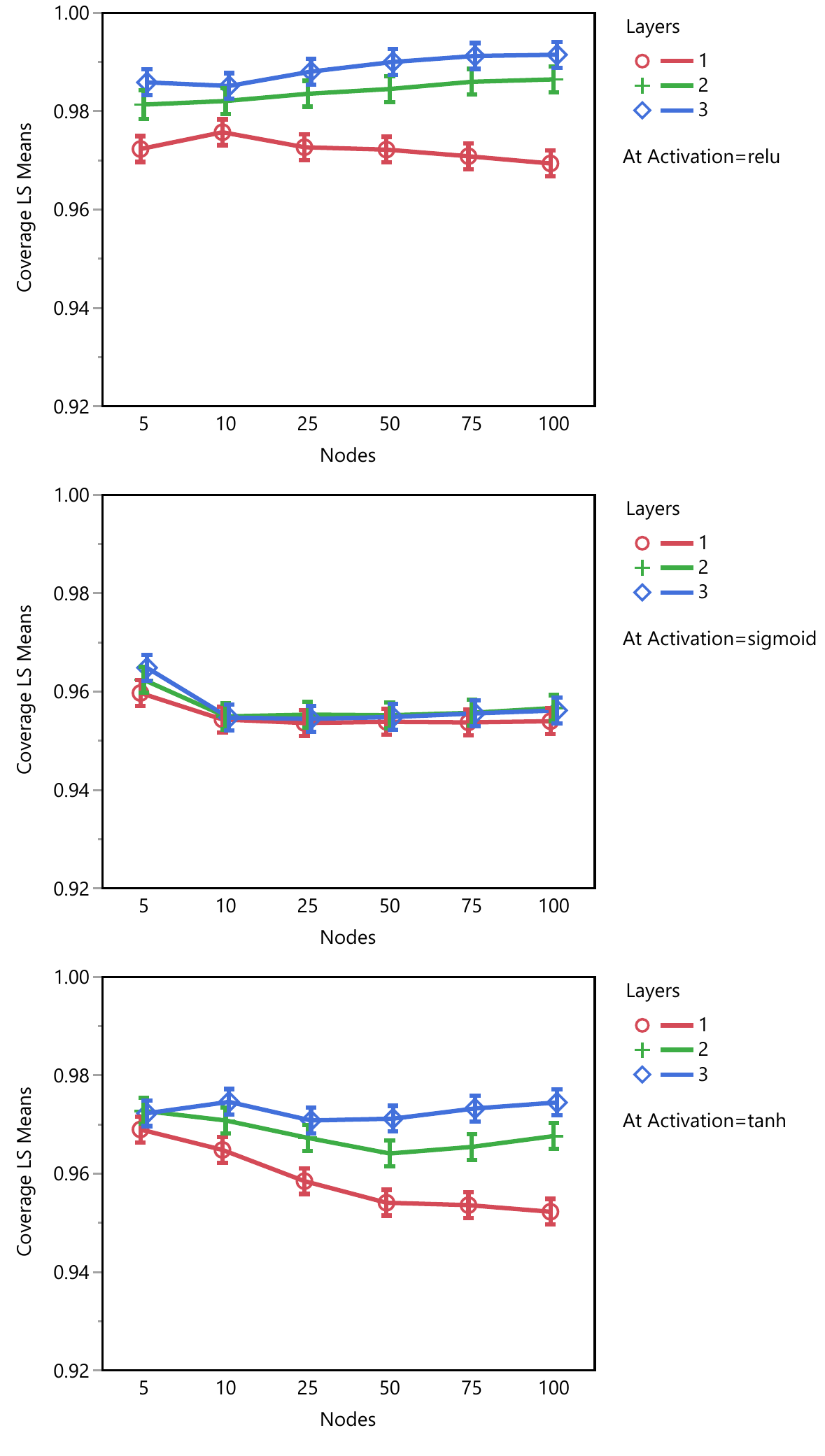}} & %
  \resizebox{2.8in}{!}{\includegraphics{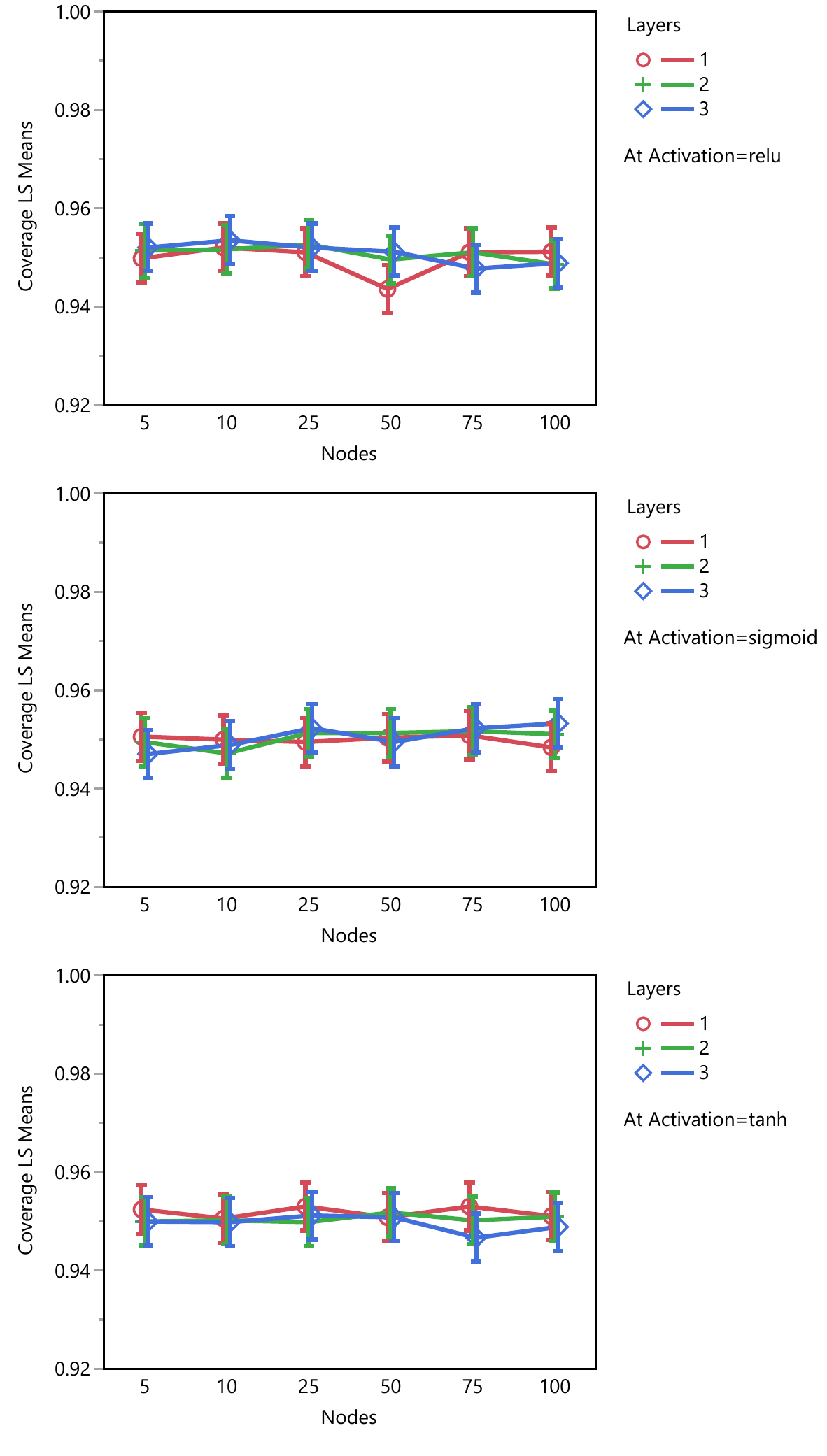}} \\
  \hline
\end{tabular}
\caption{PI Coverage in the Kinematics Data Set.}
\label{figure:Kinematics_PICov}
\end{center}
\end{figure}

\begin{figure}[H]
\begin{center}
\begin{tabular}{|c|c|}
  \hline \textbf{Pivot Bootstrap ($\bm{1,000}$ Resamples) }& \textbf{Split Conformal Inference} \\
  \hline \resizebox{2.8in}{!}{\includegraphics{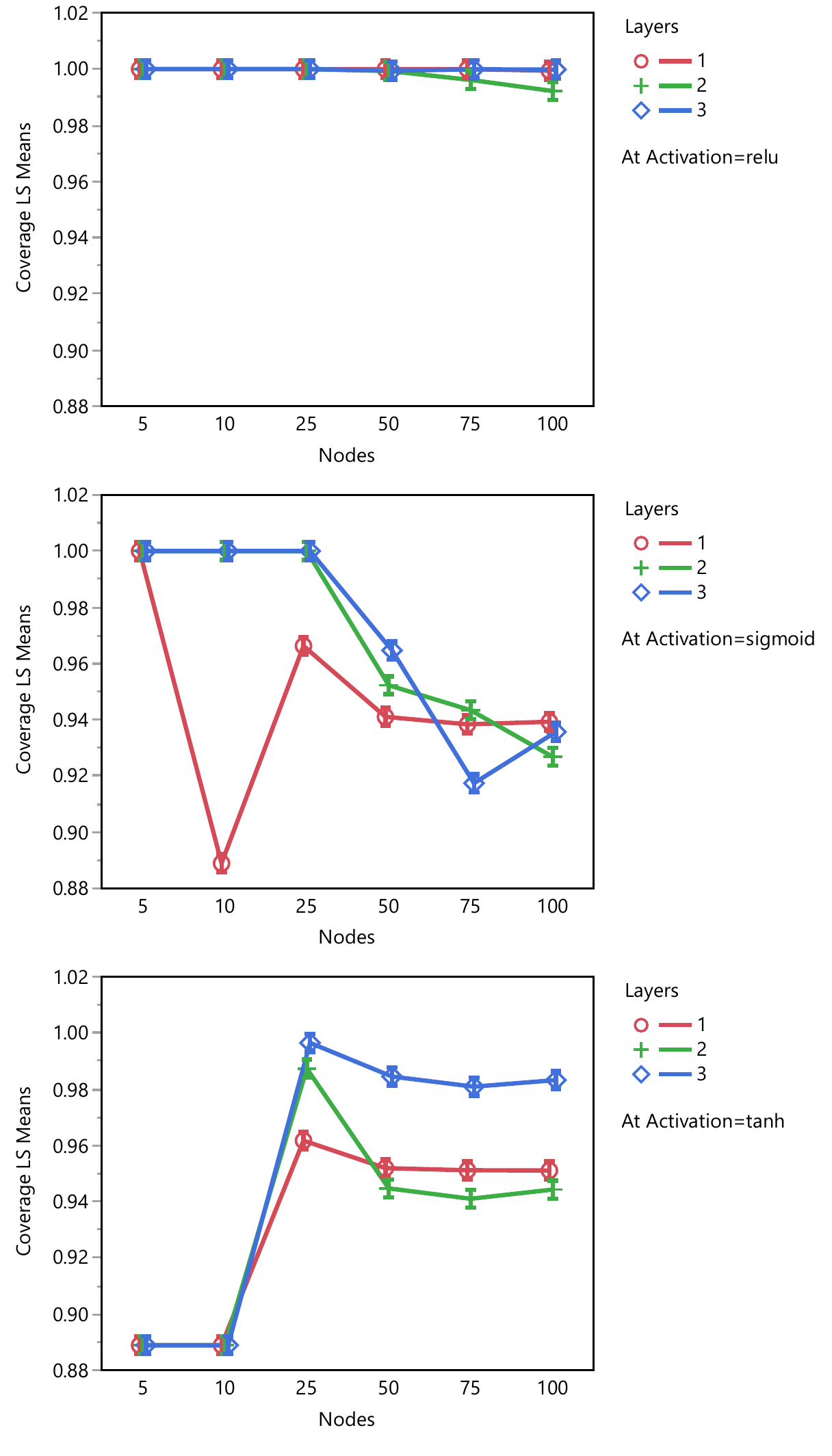}} & %
  \resizebox{2.8in}{!}{\includegraphics{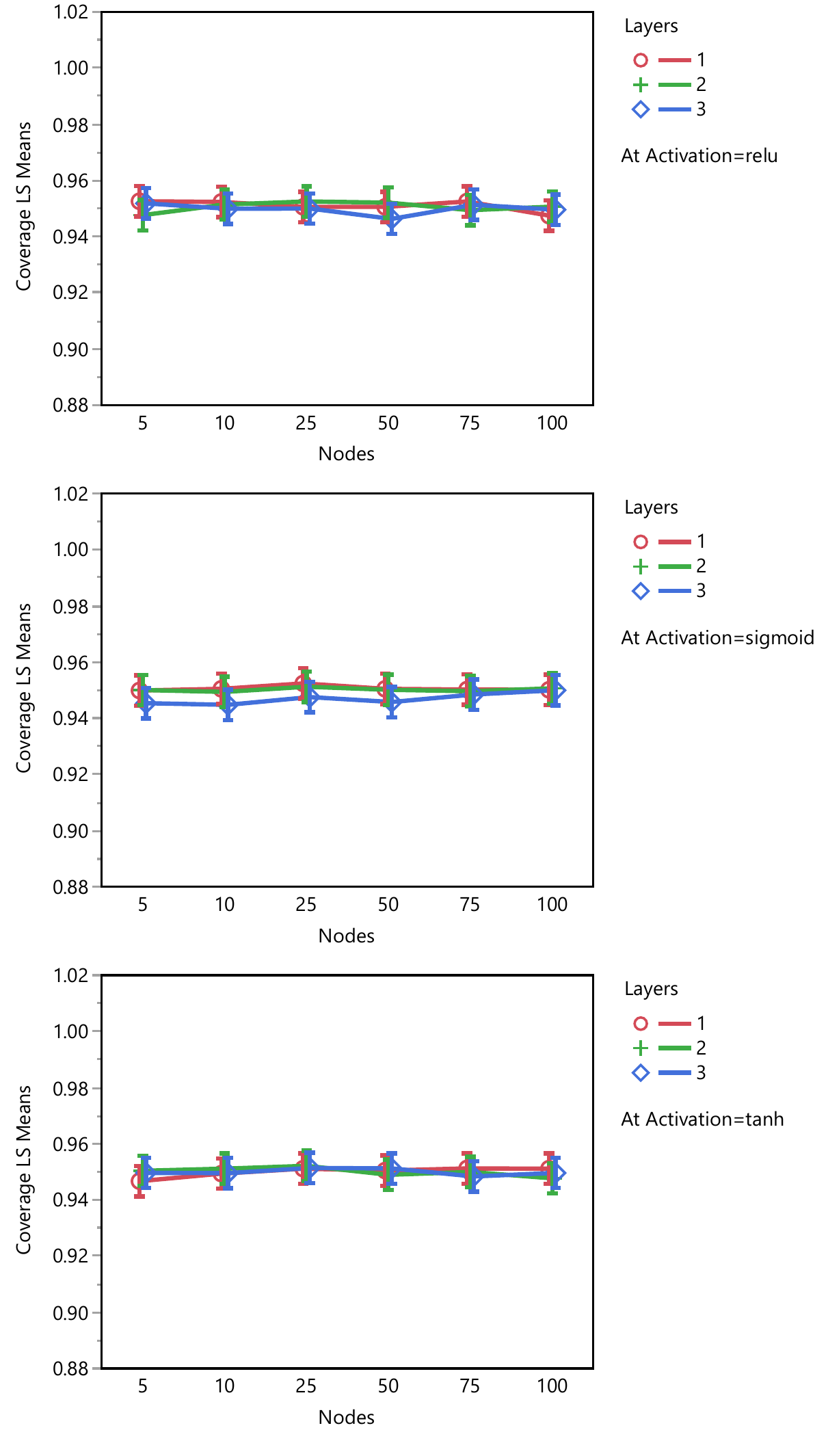}} \\
  \hline
\end{tabular}
\caption{PI Coverage in the Naval Propulsion Data Set.}
\label{figure:NavalPropulsion_PICov}
\end{center}
\end{figure}

\begin{figure}[H]
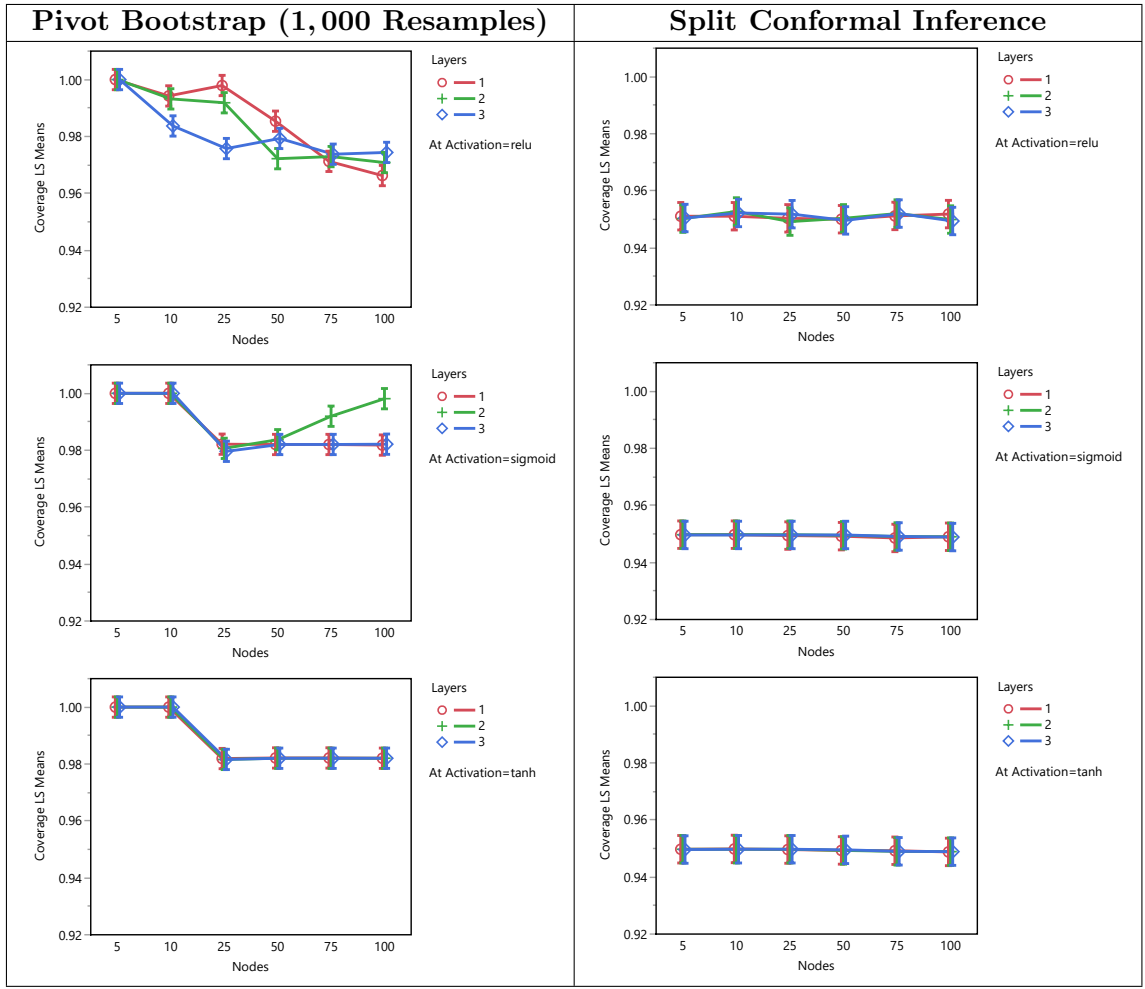

\begin{center}
\begin{tabular}{|c|c|}
  \hline \textbf{Pivot Bootstrap ($\bm{1,000}$ Resamples) }& \textbf{Split Conformal Inference} \\
  \hline \resizebox{2.8in}{!}{\includegraphics{images/PowerPlant_BootstrapResults_ANOVA_Coverage.pdf}} & %
  \resizebox{2.8in}{!}{\includegraphics{images/PowerPlant_SplitConfResults_ANOVA_Coverage.pdf}} \\
  \hline
\end{tabular}
\caption{PI Coverage in the Power Plant Data Set.}
\label{figure:PowerPlant_PICov}
\end{center}
\end{figure}

\begin{figure}[H]
\begin{center}
\begin{tabular}{|c|c|}
  \hline \textbf{Pivot Bootstrap ($\bm{1,000}$ Resamples) }& \textbf{Split Conformal Inference} \\
  \hline \resizebox{2.8in}{!}{\includegraphics{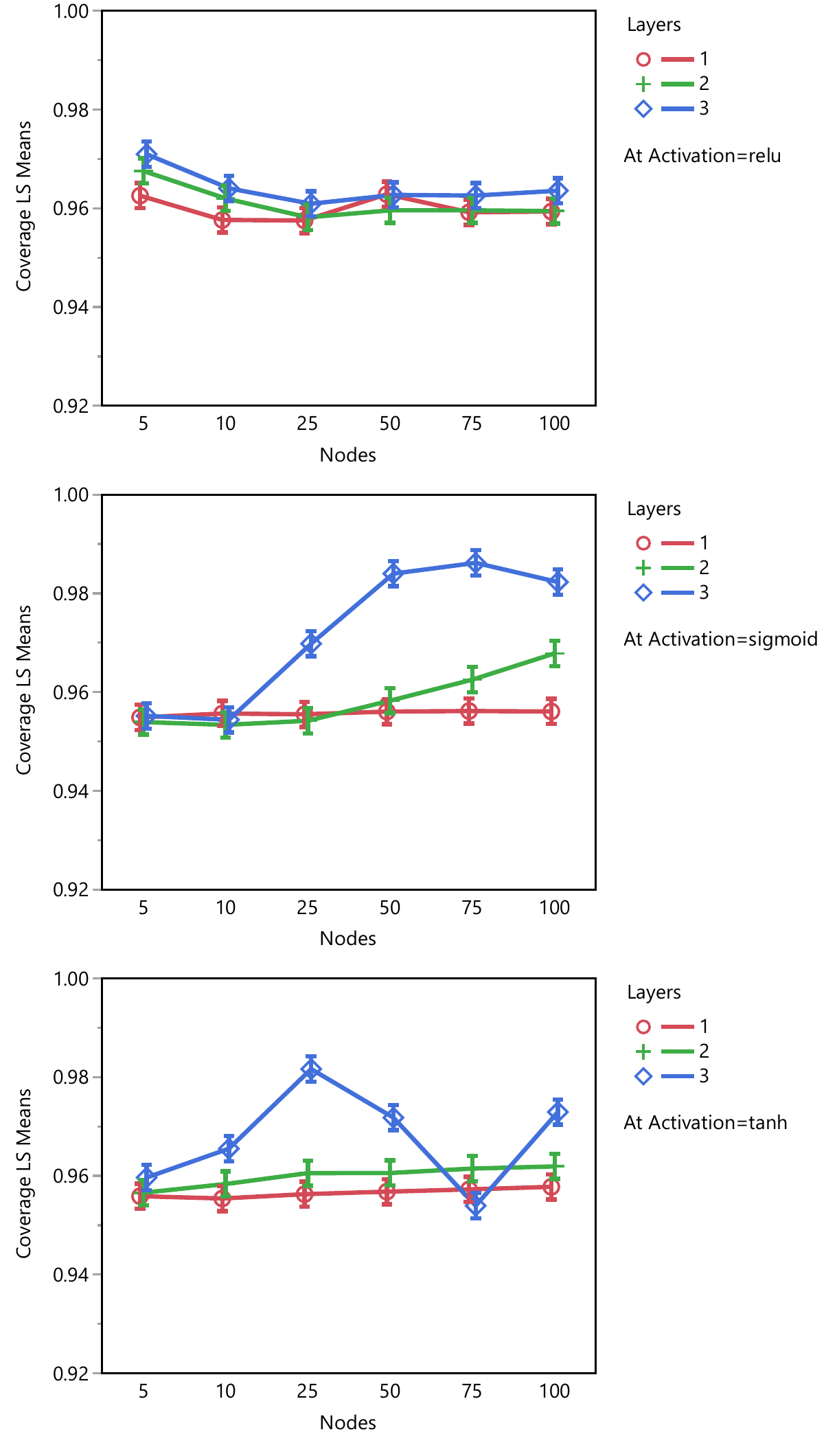}} & %
  \resizebox{2.8in}{!}{\includegraphics{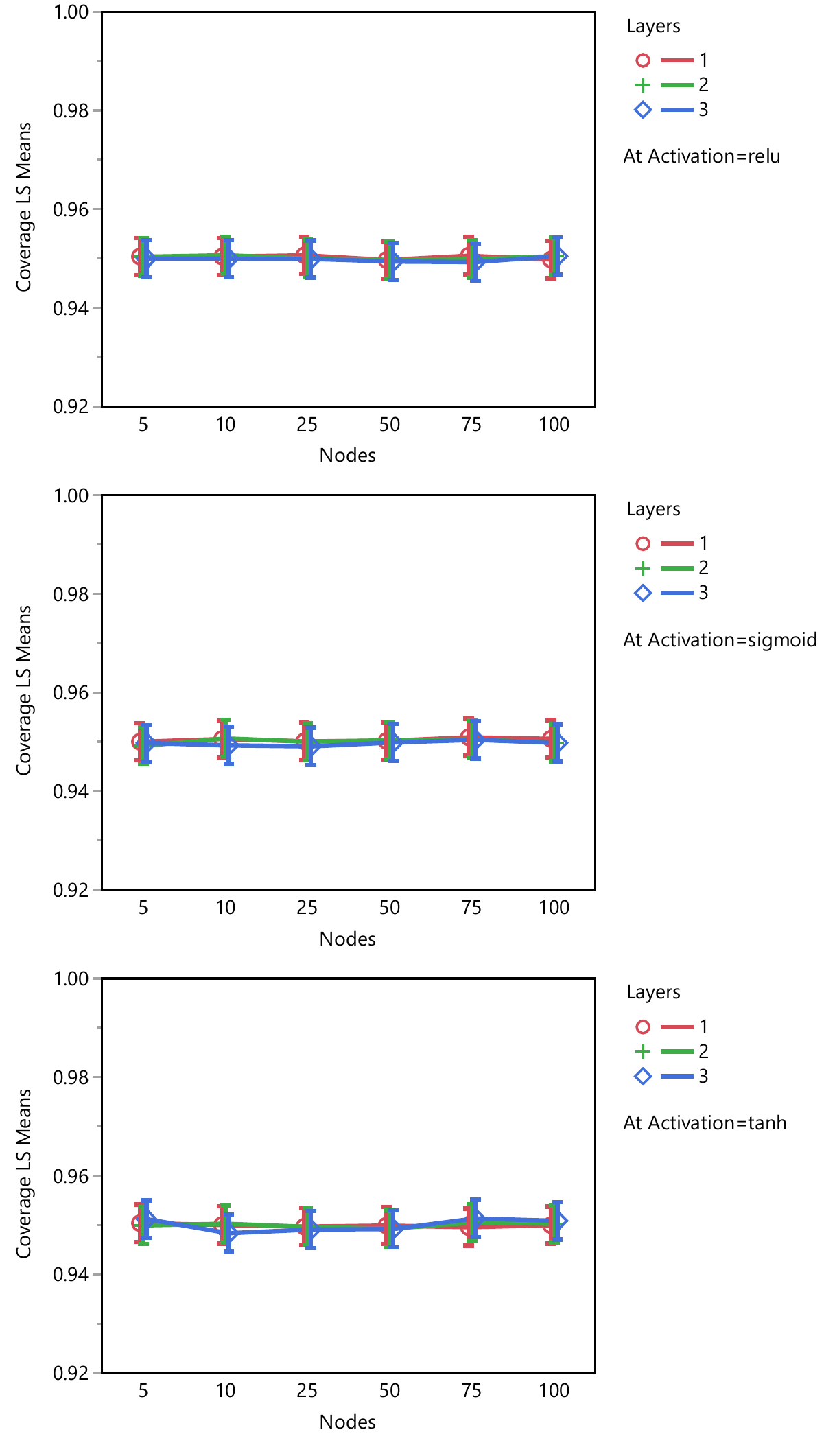}} \\
  \hline
\end{tabular}
\caption{PI Coverage in the Protein Structure Data Set.}
\label{figure:ProteinStructure_PICov}
\end{center}
\end{figure}

\begin{figure}[H]
\begin{center}
\begin{tabular}{|c|c|}
  \hline \textbf{Pivot Bootstrap ($\bm{1,000}$ Resamples) }& \textbf{Split Conformal Inference} \\
  \hline \resizebox{2.8in}{!}{\includegraphics{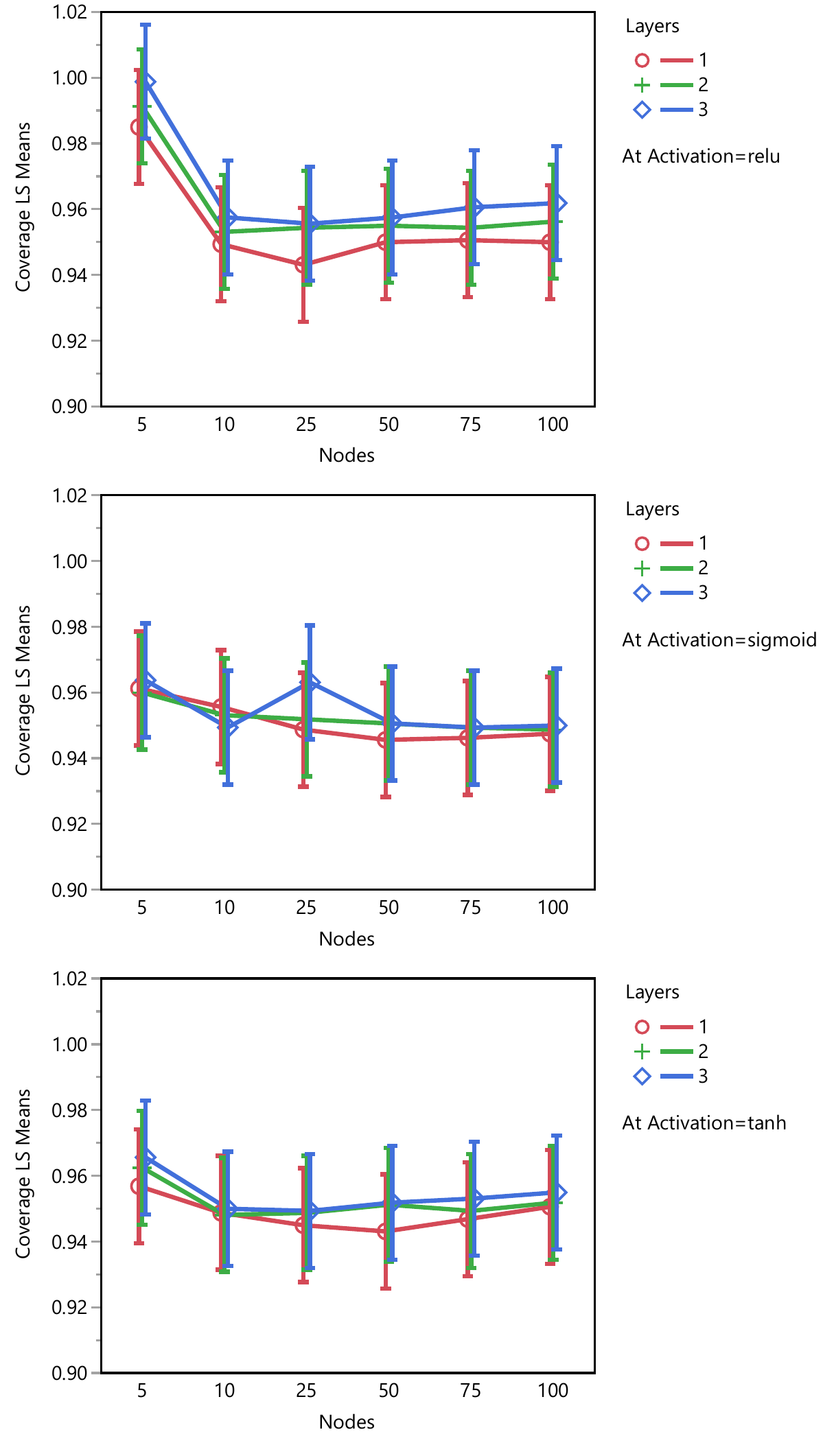}} & %
  \resizebox{2.8in}{!}{\includegraphics{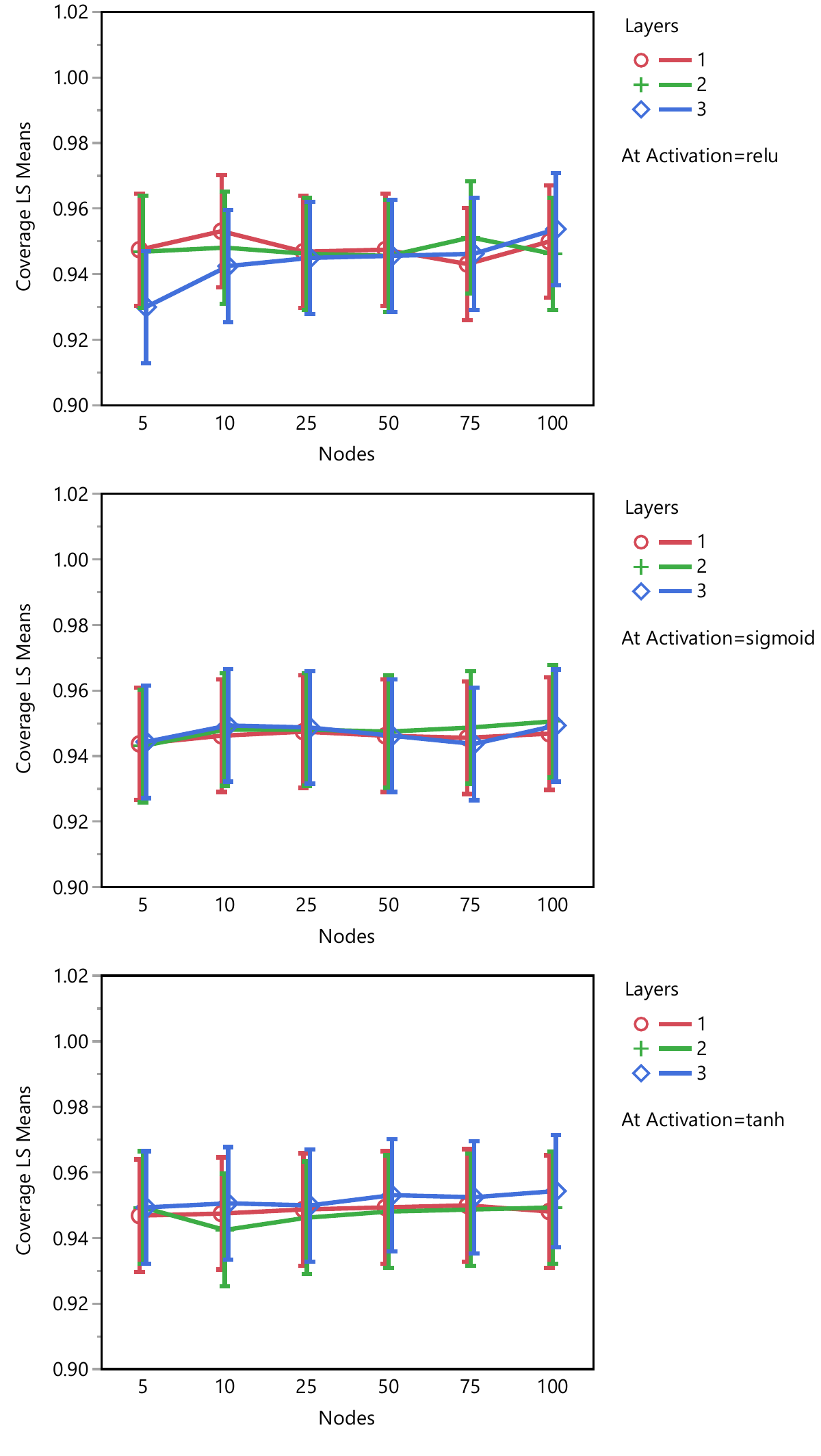}} \\
  \hline
\end{tabular}
\caption{PI Coverage in the Wine Quality Data Set.}
\label{figure:WineQuality_PICov}
\end{center}
\end{figure}

\begin{figure}[H]
\begin{center}
\begin{tabular}{|c|c|}
  \hline \textbf{Pivot Bootstrap ($\bm{1,000}$ Resamples) }& \textbf{Split Conformal Inference} \\
  \hline \resizebox{2.8in}{!}{\includegraphics{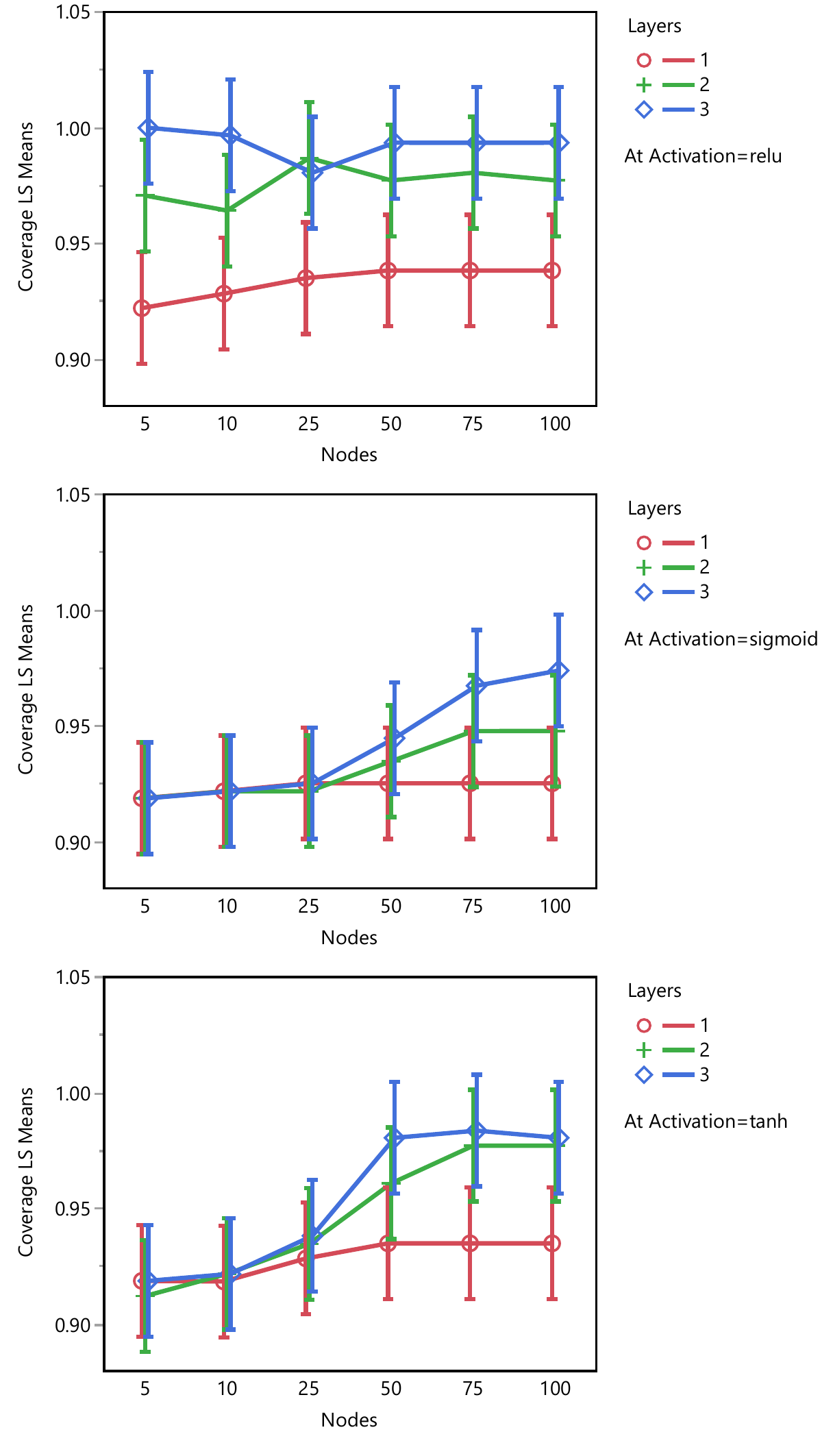}} & %
  \resizebox{2.8in}{!}{\includegraphics{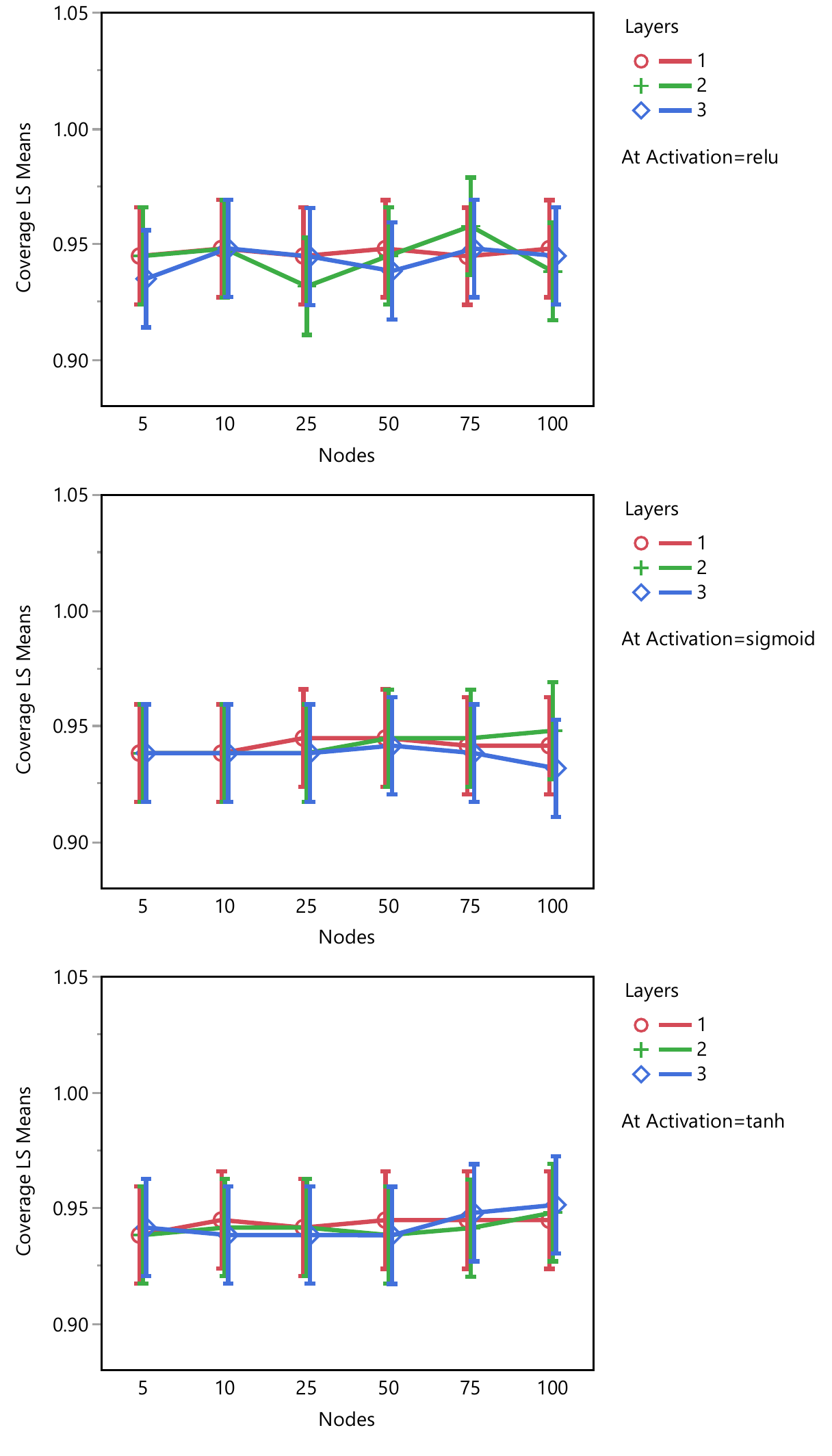}} \\
  \hline
\end{tabular}
\caption{PI Coverage in the Yacht Hydrodynamics Data Set.}
\label{figure:YachtHydro_PICov}
\end{center}
\end{figure}

\begin{figure}[H]
\begin{center}
\begin{tabular}{|c|c|}
  \hline \textbf{Pivot Bootstrap ($\bm{1,000}$ Resamples) }& \textbf{Split Conformal Inference} \\
  \hline \resizebox{2.8in}{!}{\includegraphics{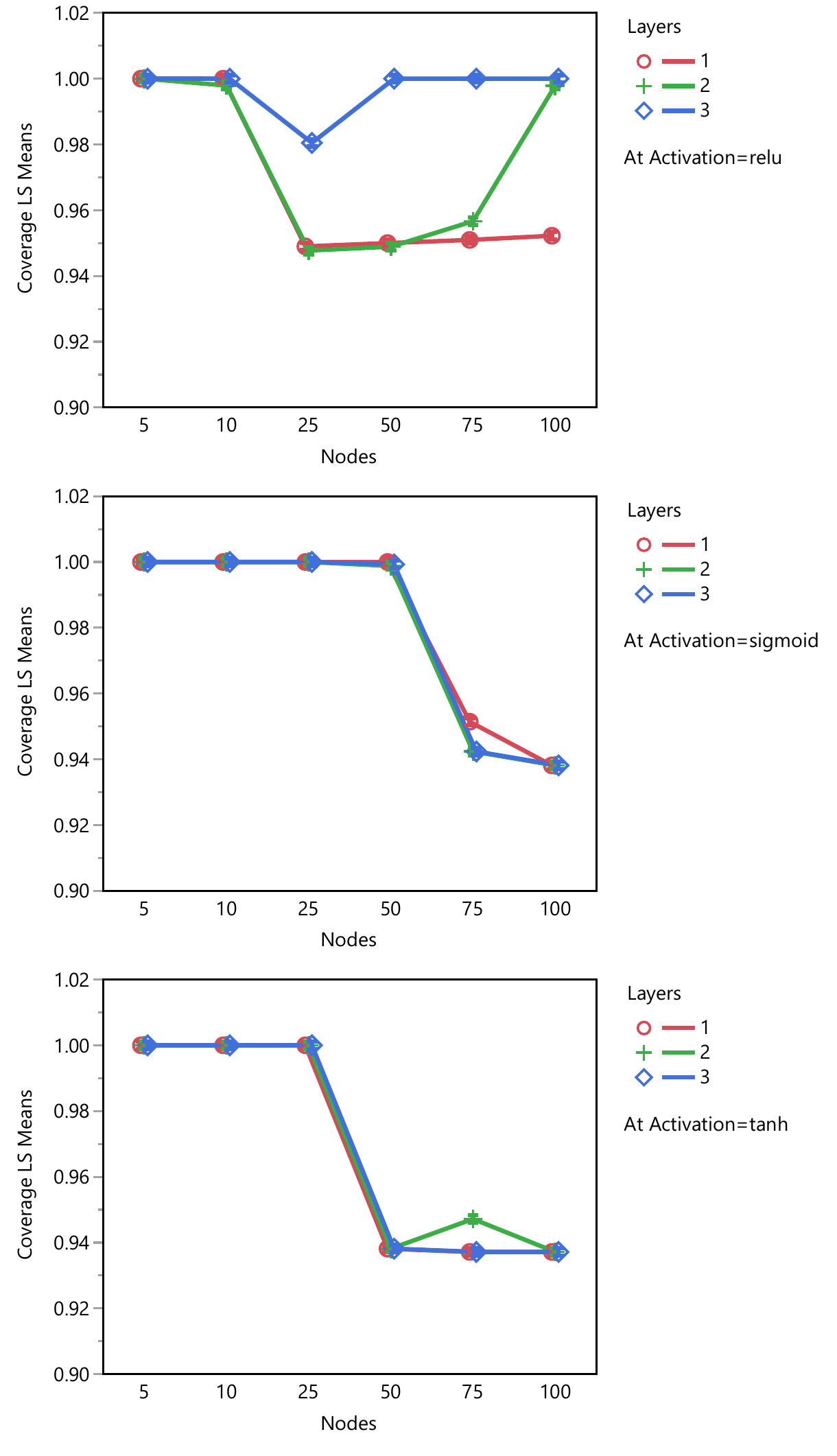}} & %
  \resizebox{2.8in}{!}{\includegraphics{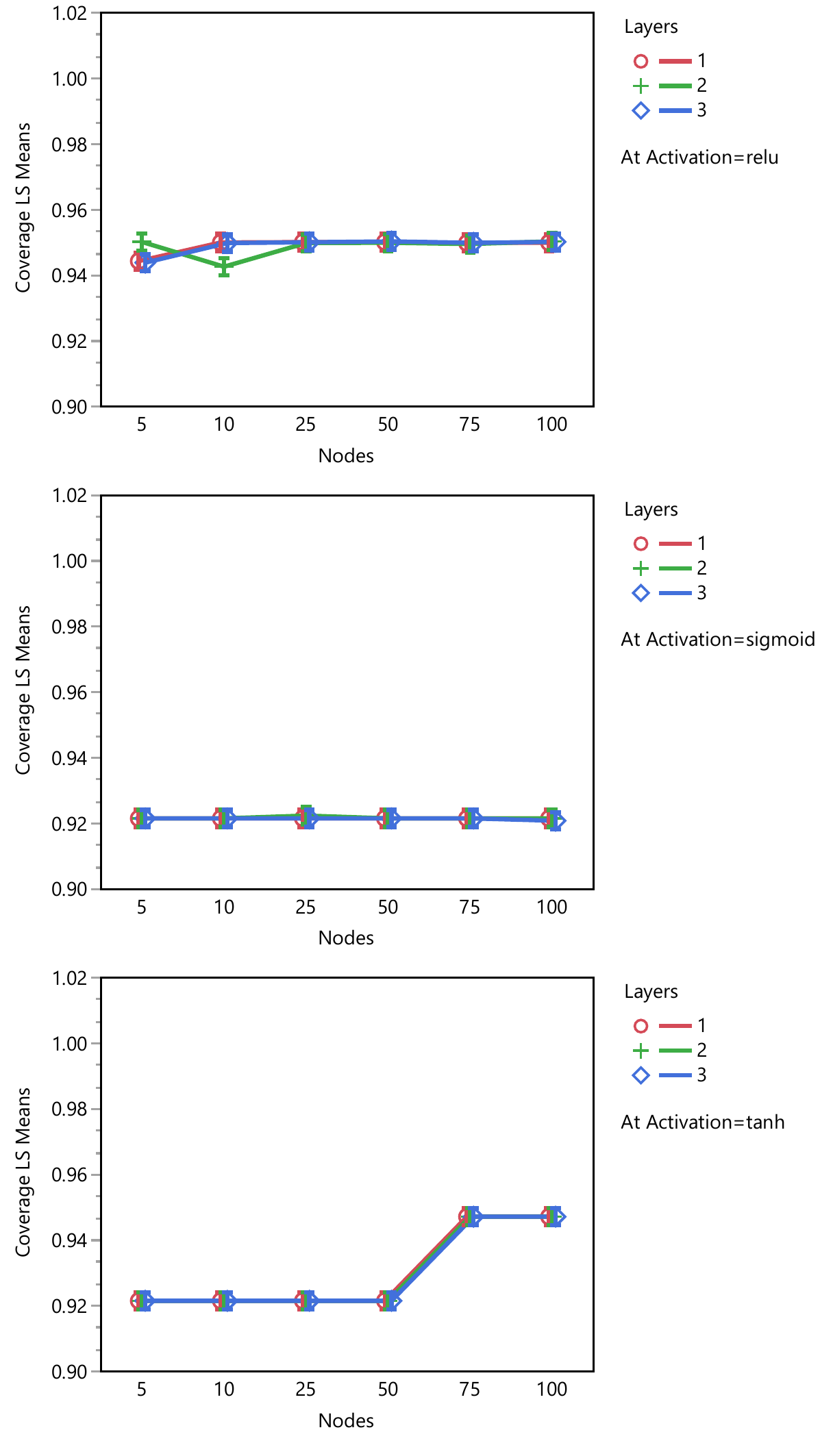}} \\
  \hline
\end{tabular}
\caption{PI Coverage in the Year Prediction Data Set.}
\label{figure:YearPrediction_PICov}
\end{center}
\end{figure}

\section{PI Average Width Plots} \label{appendix_b}

\begin{figure}[H]
\begin{center}
\begin{tabular}{|c|c|}
  \hline \textbf{Pivot Bootstrap ($\bm{1,000}$ Resamples) }& \textbf{Split Conformal Inference} \\
  \hline \resizebox{2.8in}{!}{\includegraphics{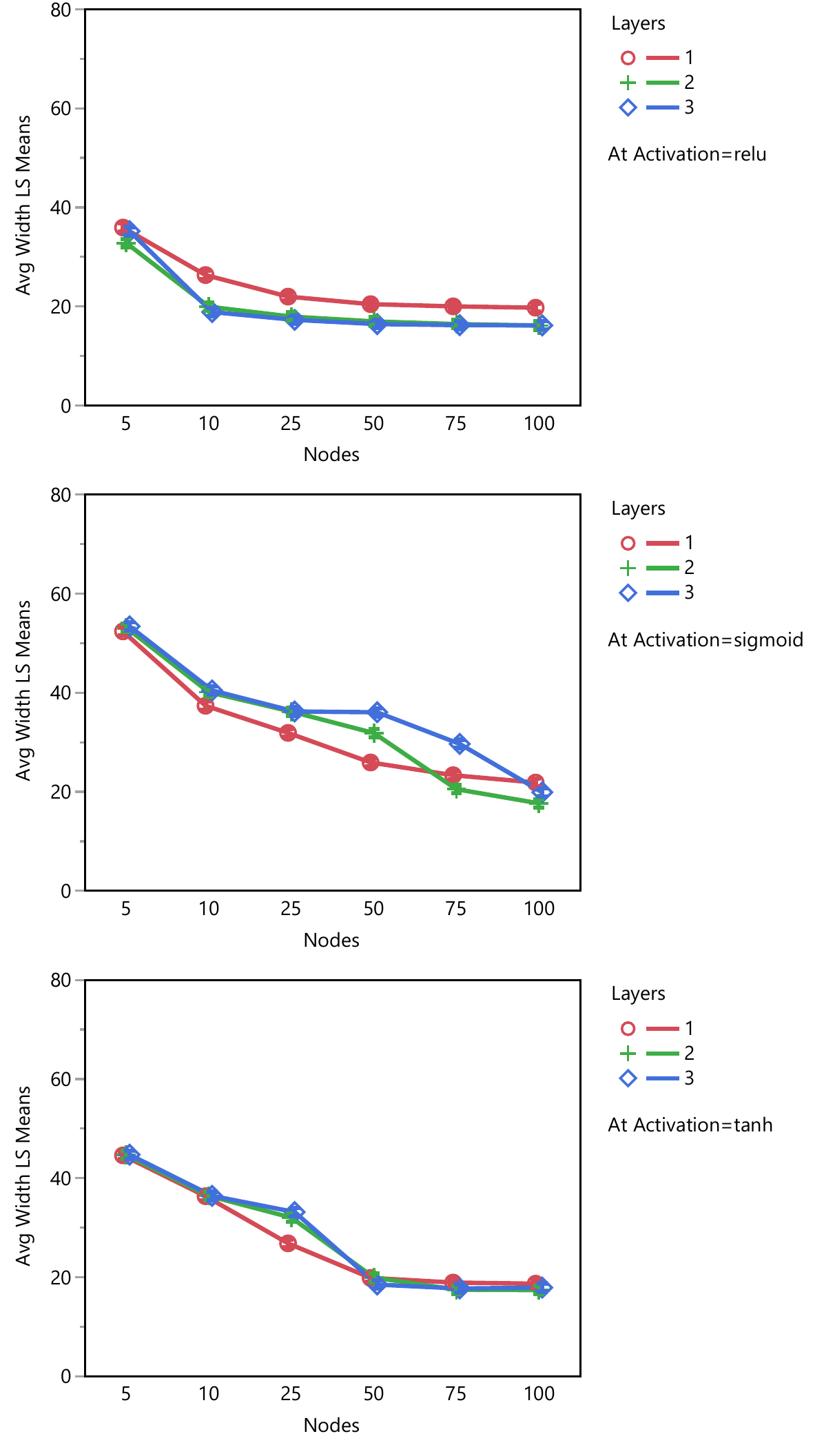}} & %
  \resizebox{2.8in}{!}{\includegraphics{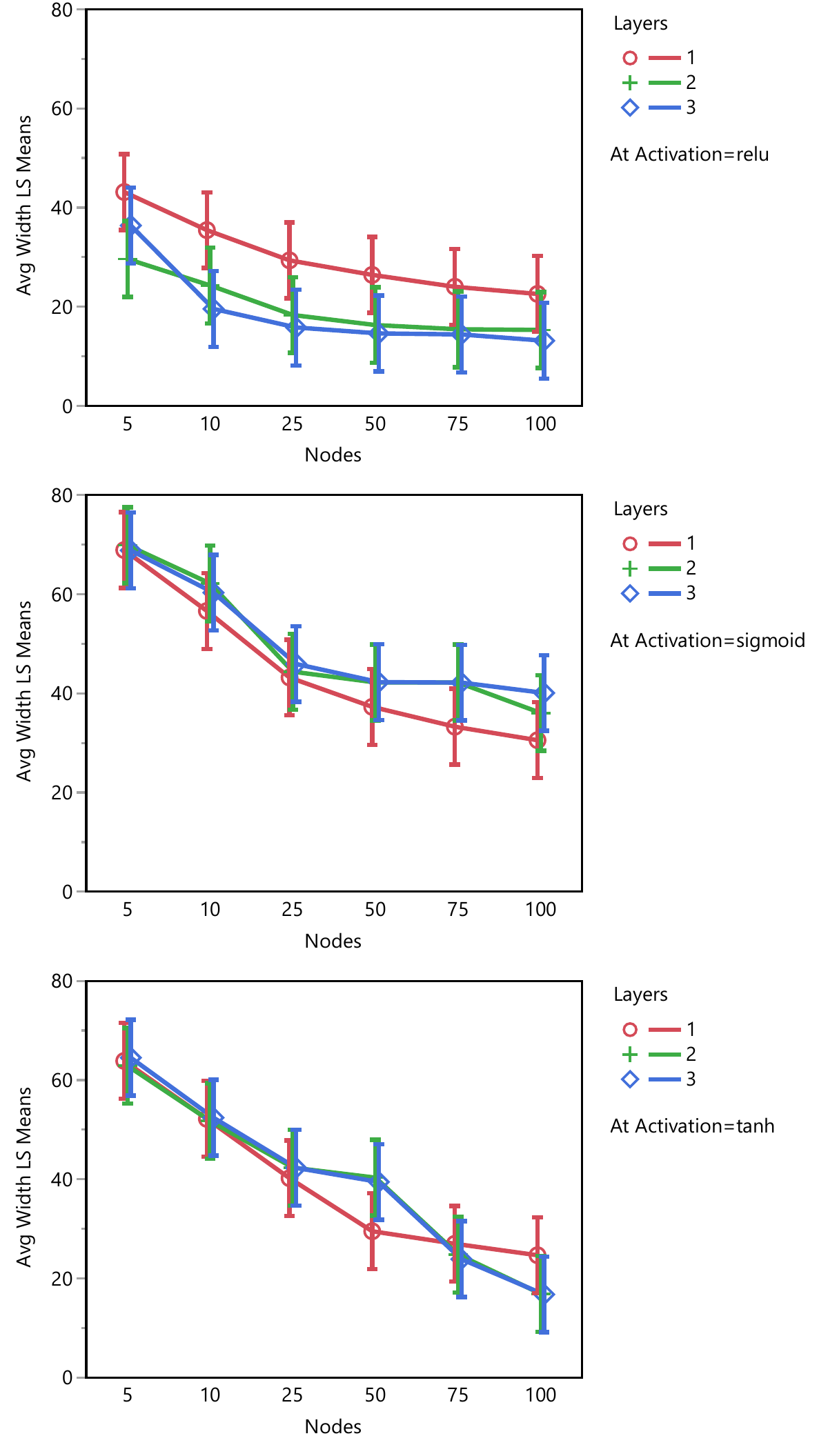}} \\
  \hline
\end{tabular}
\caption{PI Coverage in the Boston Housing Data Set.}
\label{figure:Boston_PIAvgW}
\end{center}
\end{figure}

\begin{figure}[H]
\begin{center}
\begin{tabular}{|c|c|}
  \hline \textbf{Pivot Bootstrap ($\bm{1,000}$ Resamples) }& \textbf{Split Conformal Inference} \\
  \hline \resizebox{2.8in}{!}{\includegraphics{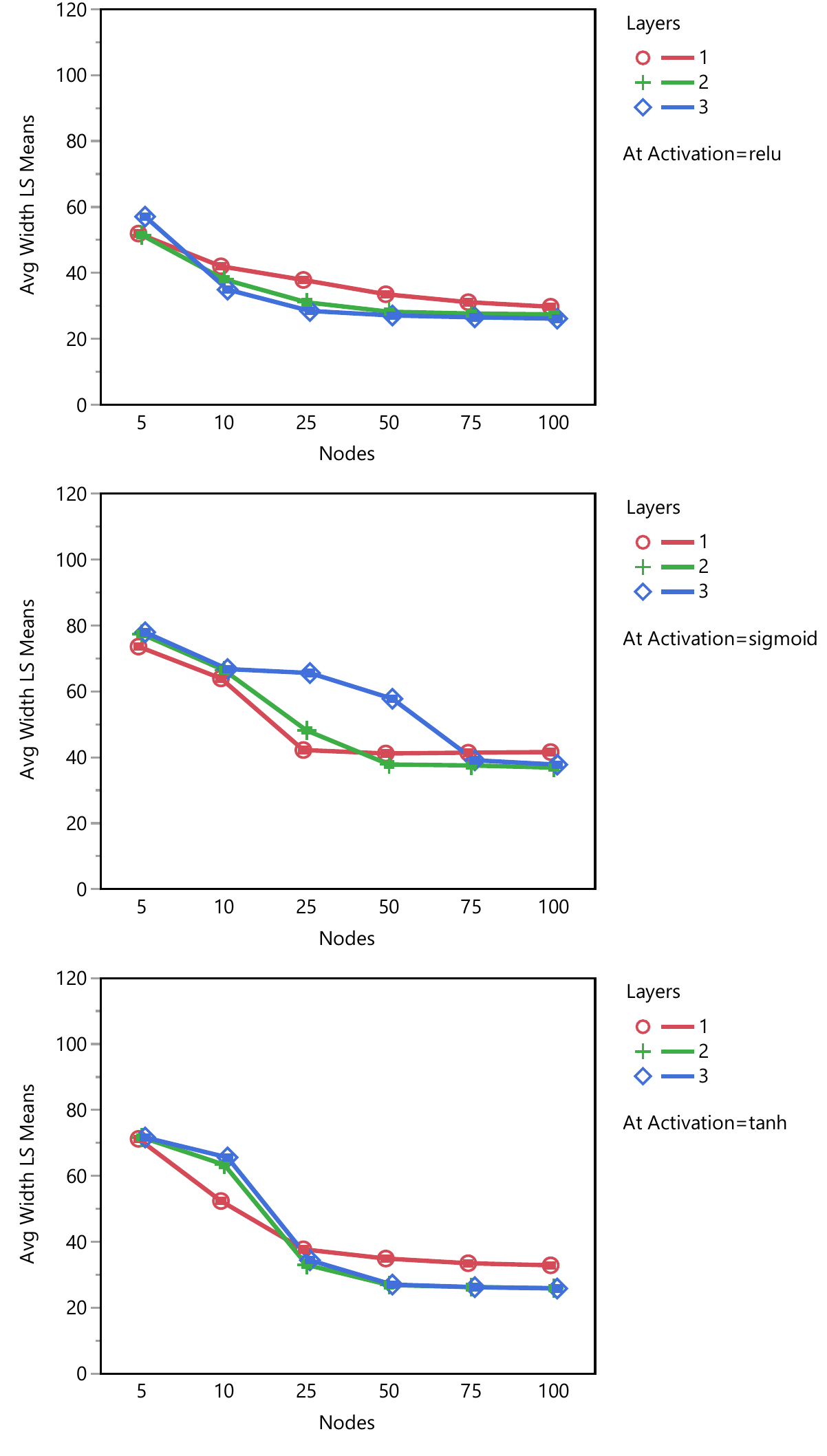}} & %
  \resizebox{2.8in}{!}{\includegraphics{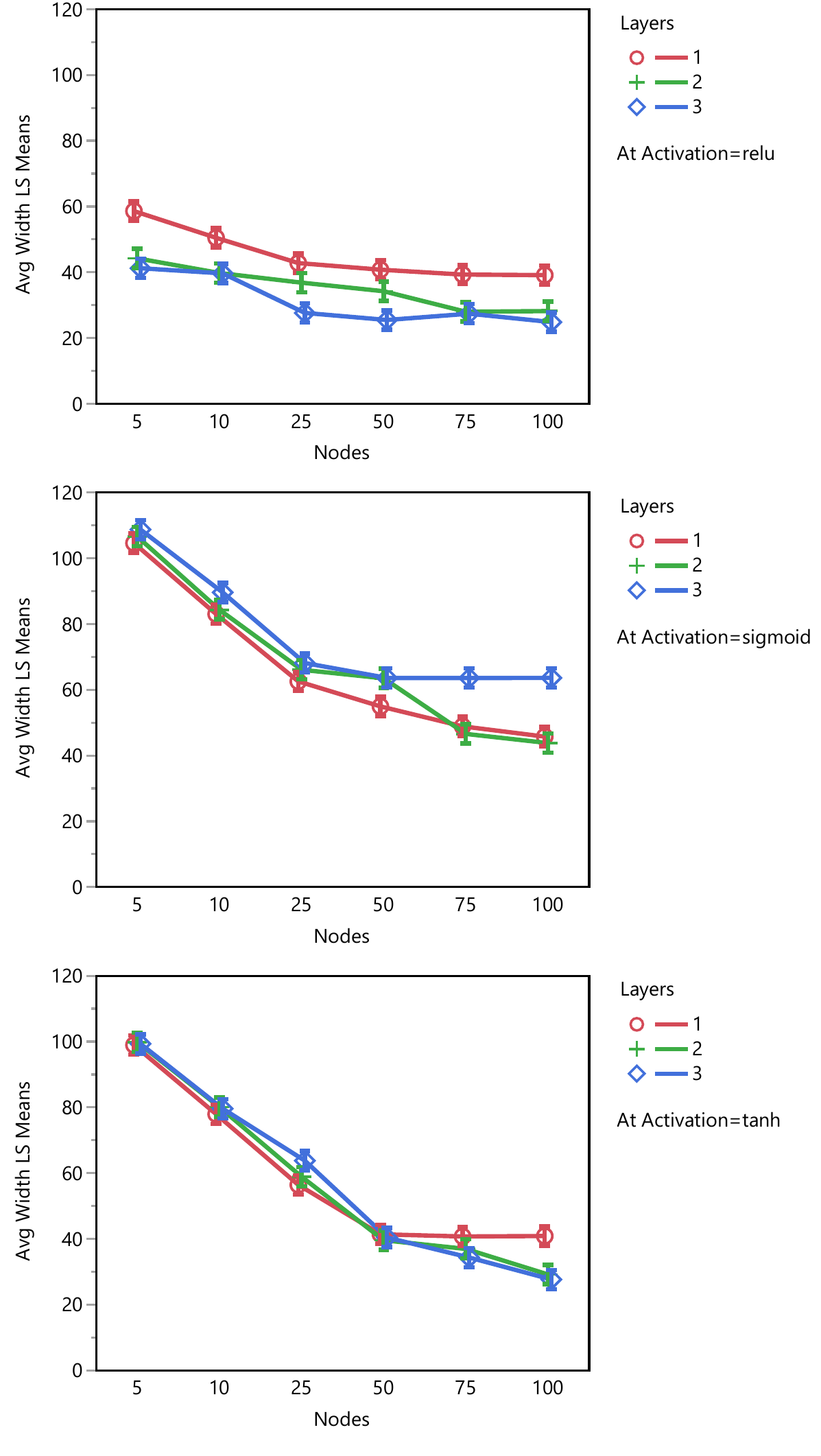}} \\
  \hline
\end{tabular}
\caption{PI Average Width in the Concrete Strength Data Set.}
\label{figure:ConcreteStrength_PIAvgW}
\end{center}
\end{figure}

\begin{figure}[H]
\begin{center}
\begin{tabular}{|c|c|}
  \hline \textbf{Pivot Bootstrap ($\bm{1,000}$ Resamples) }& \textbf{Split Conformal Inference} \\
  \hline \resizebox{2.8in}{!}{\includegraphics{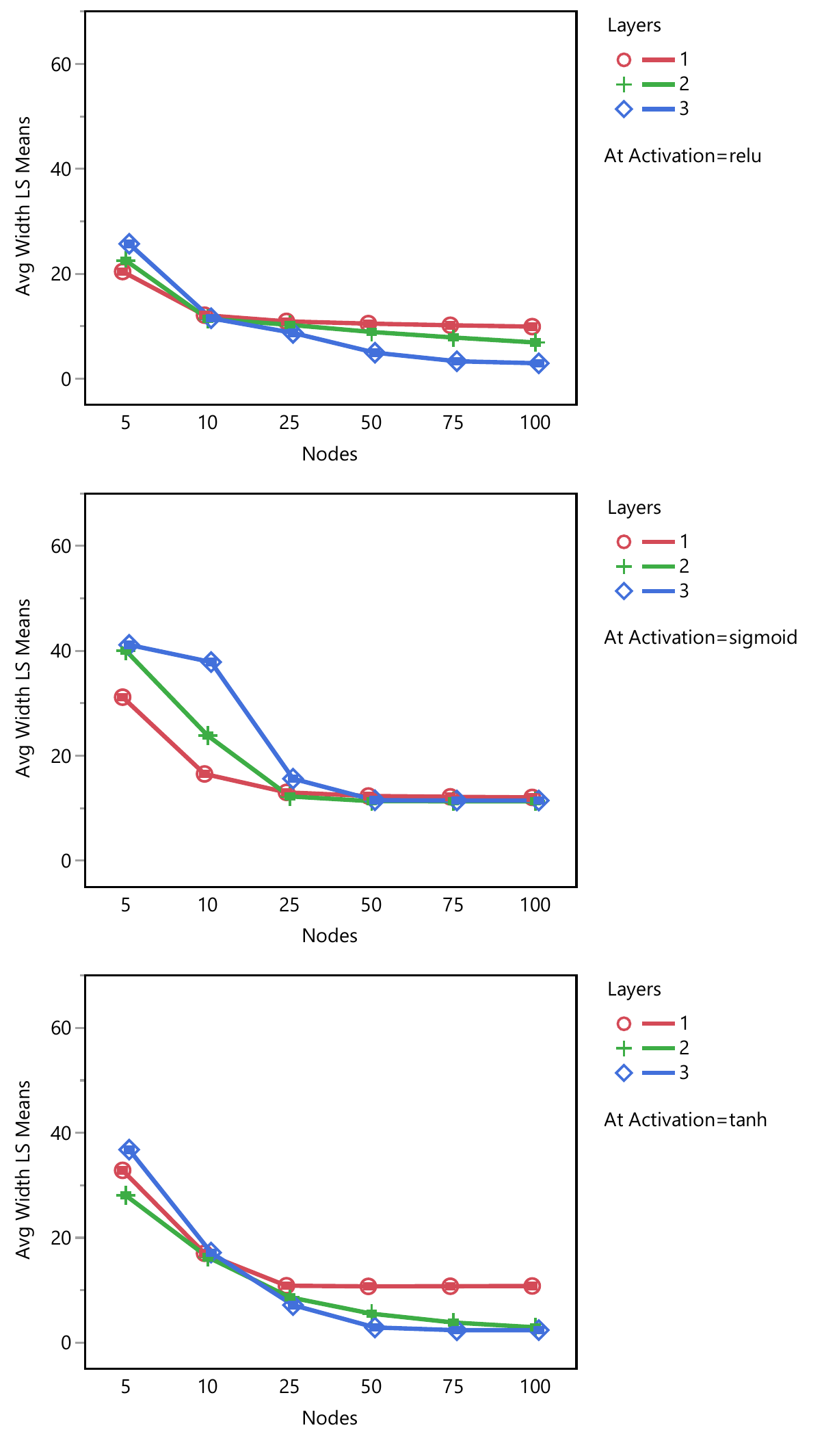}} & %
  \resizebox{2.8in}{!}{\includegraphics{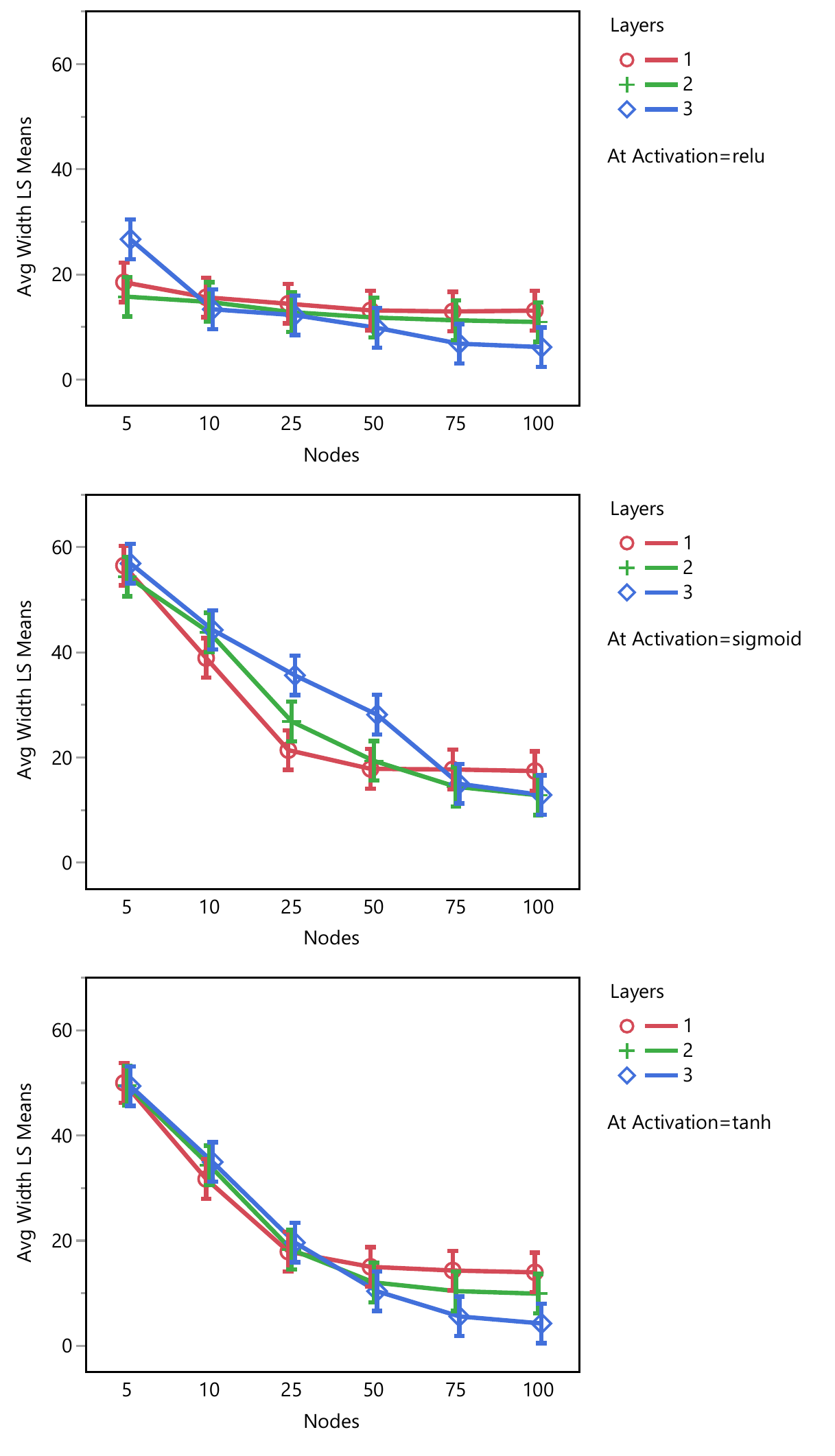}} \\
  \hline
\end{tabular}
\caption{PI Average Width in the Energy Efficiency Data Set.}
\label{figure:EnergyEfficiency_PIAvgW}
\end{center}
\end{figure}

\begin{figure}[H]
\begin{center}
\begin{tabular}{|c|c|}
  \hline \textbf{Pivot Bootstrap ($\bm{1,000}$ Resamples) }& \textbf{Split Conformal Inference} \\
  \hline \resizebox{2.8in}{!}{\includegraphics{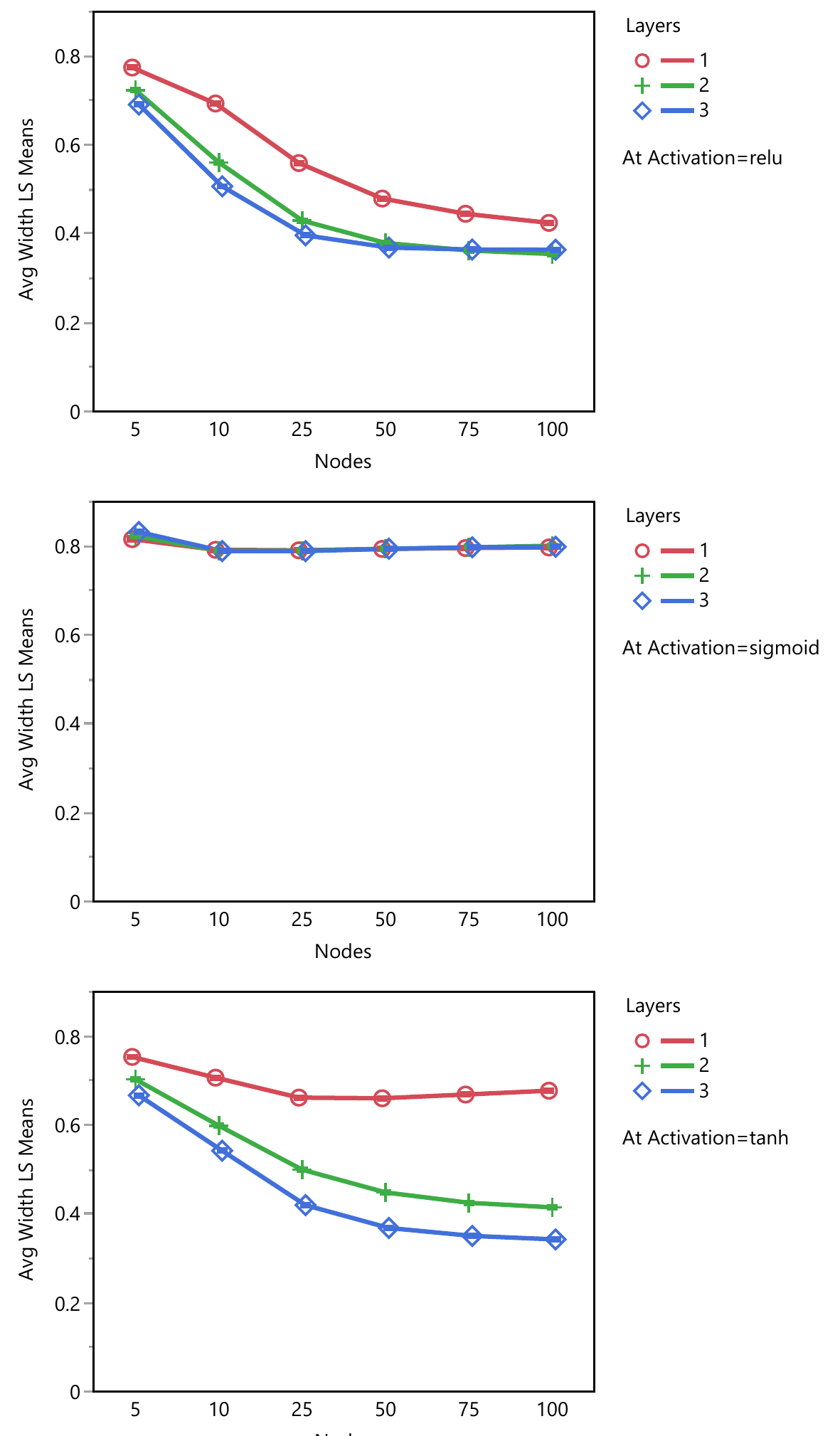}} & %
  \resizebox{2.8in}{!}{\includegraphics{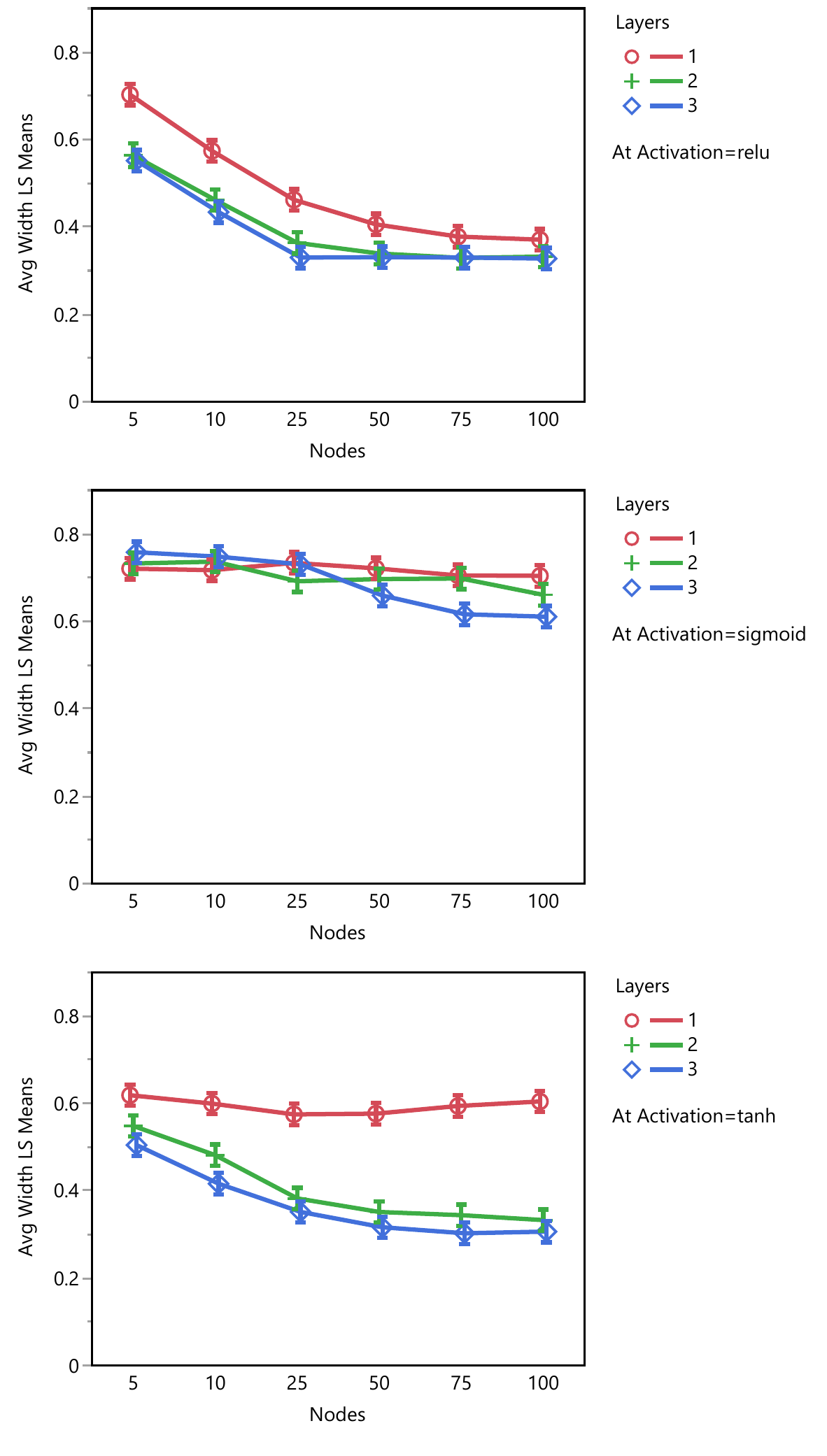}} \\
  \hline
\end{tabular}
\caption{PI Average Width in the Kinematics Data Set.}
\label{figure:Kinematics_PIAvgW}
\end{center}
\end{figure}

\begin{figure}[H]
\begin{center}
\begin{tabular}{|c|c|}
  \hline \textbf{Pivot Bootstrap ($\bm{1,000}$ Resamples) }& \textbf{Split Conformal Inference} \\
  \hline \resizebox{2.8in}{!}{\includegraphics{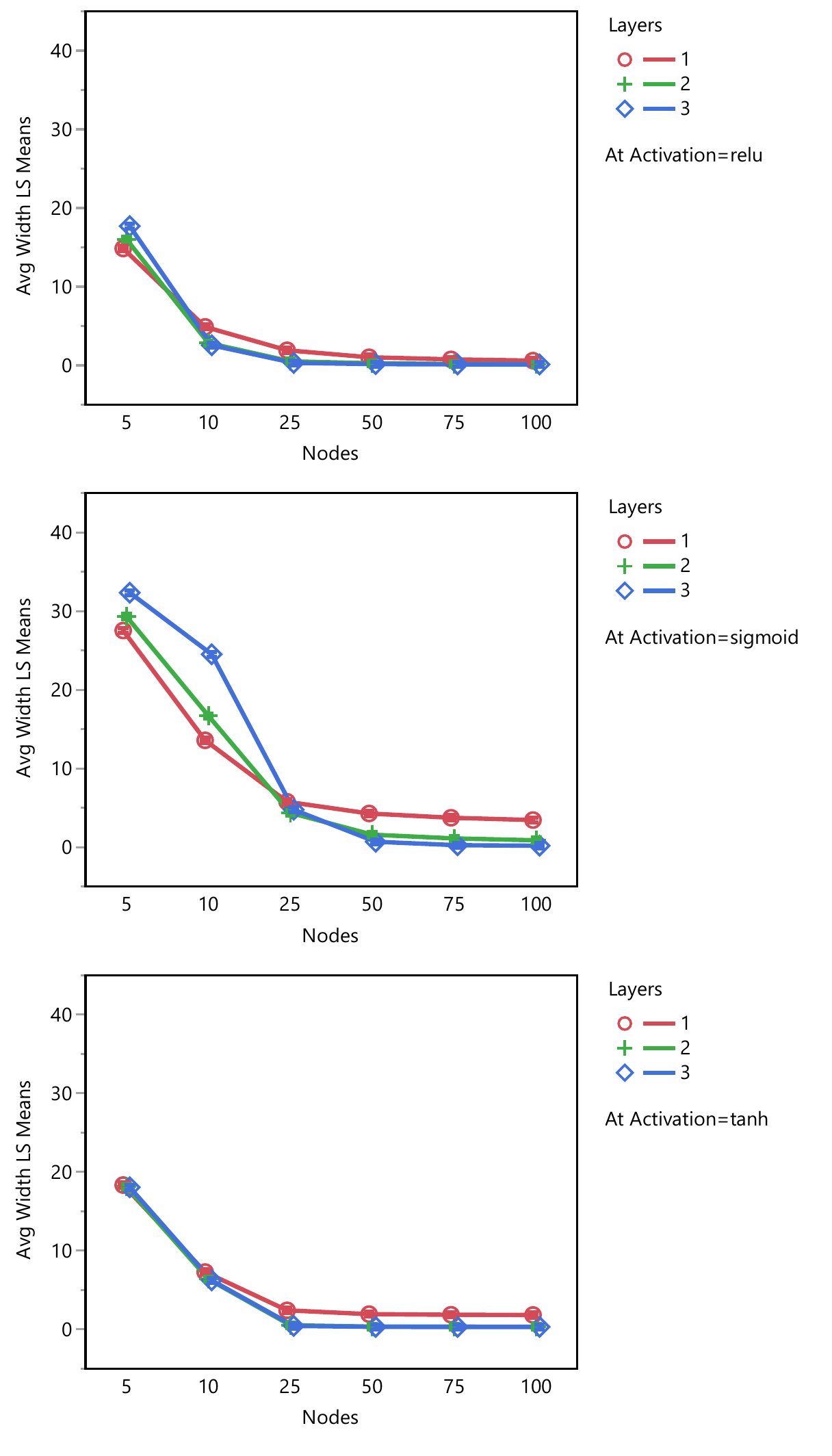}} & %
  \resizebox{2.8in}{!}{\includegraphics{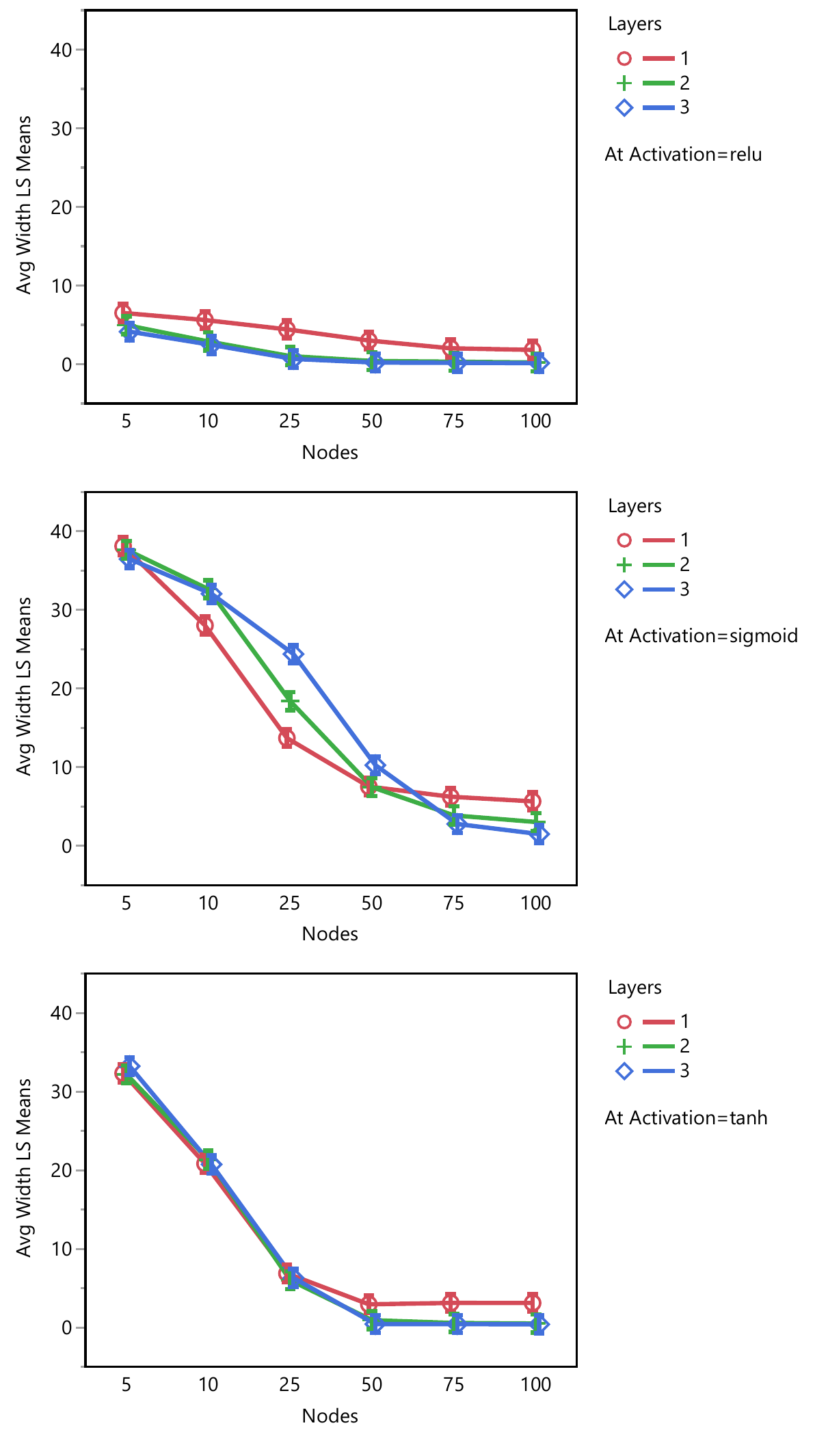}} \\
  \hline
\end{tabular}
\caption{PI Average Width in the Naval Propulsion Data Set.}
\label{figure:NavalPropulsion_PIAvgW}
\end{center}
\end{figure}

\begin{figure}[H]
\begin{center}
\begin{tabular}{|c|c|}
  \hline \textbf{Pivot Bootstrap ($\bm{1,000}$ Resamples) }& \textbf{Split Conformal Inference} \\
  \hline \resizebox{2.8in}{!}{\includegraphics{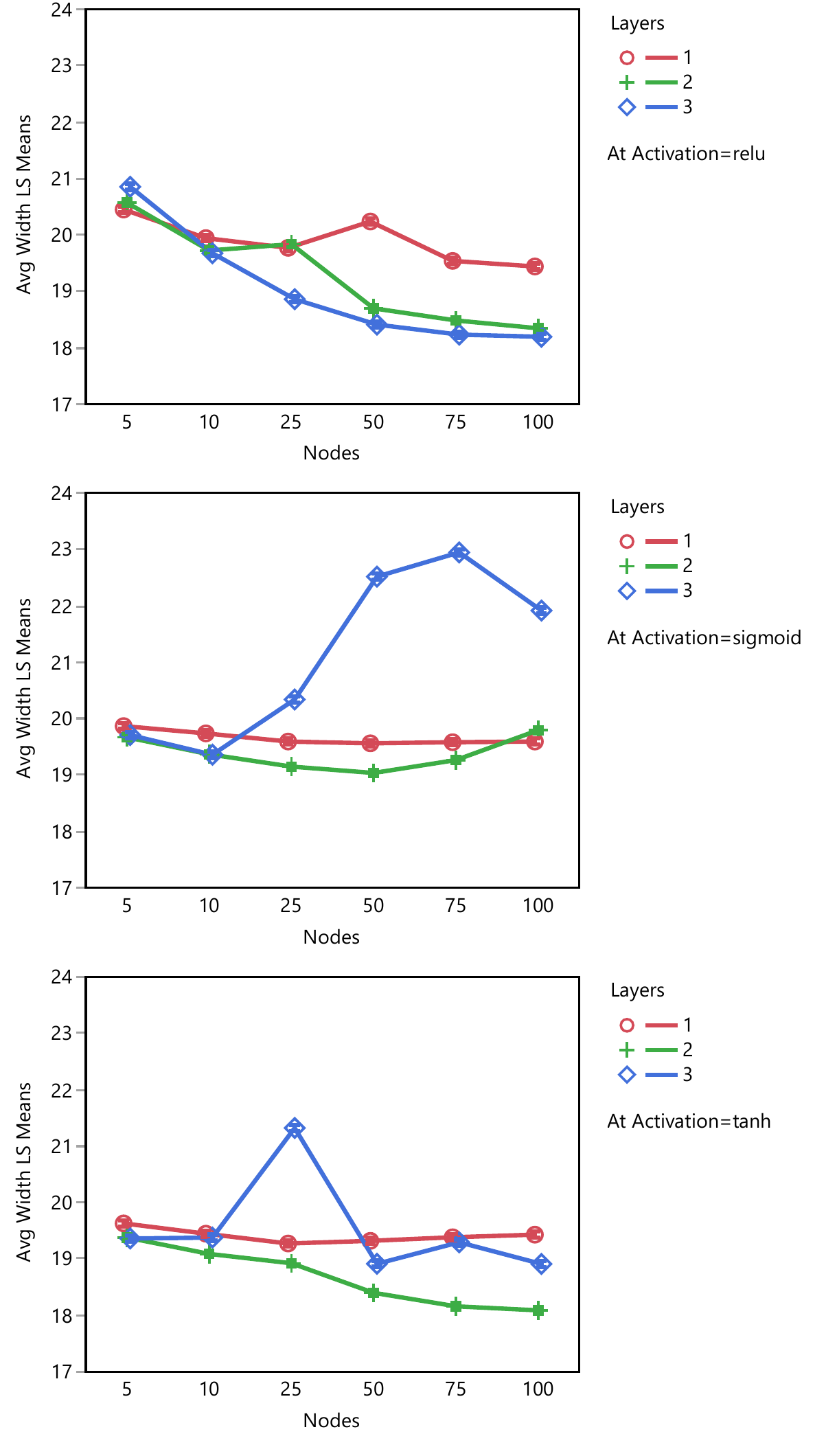}} & %
  \resizebox{2.8in}{!}{\includegraphics{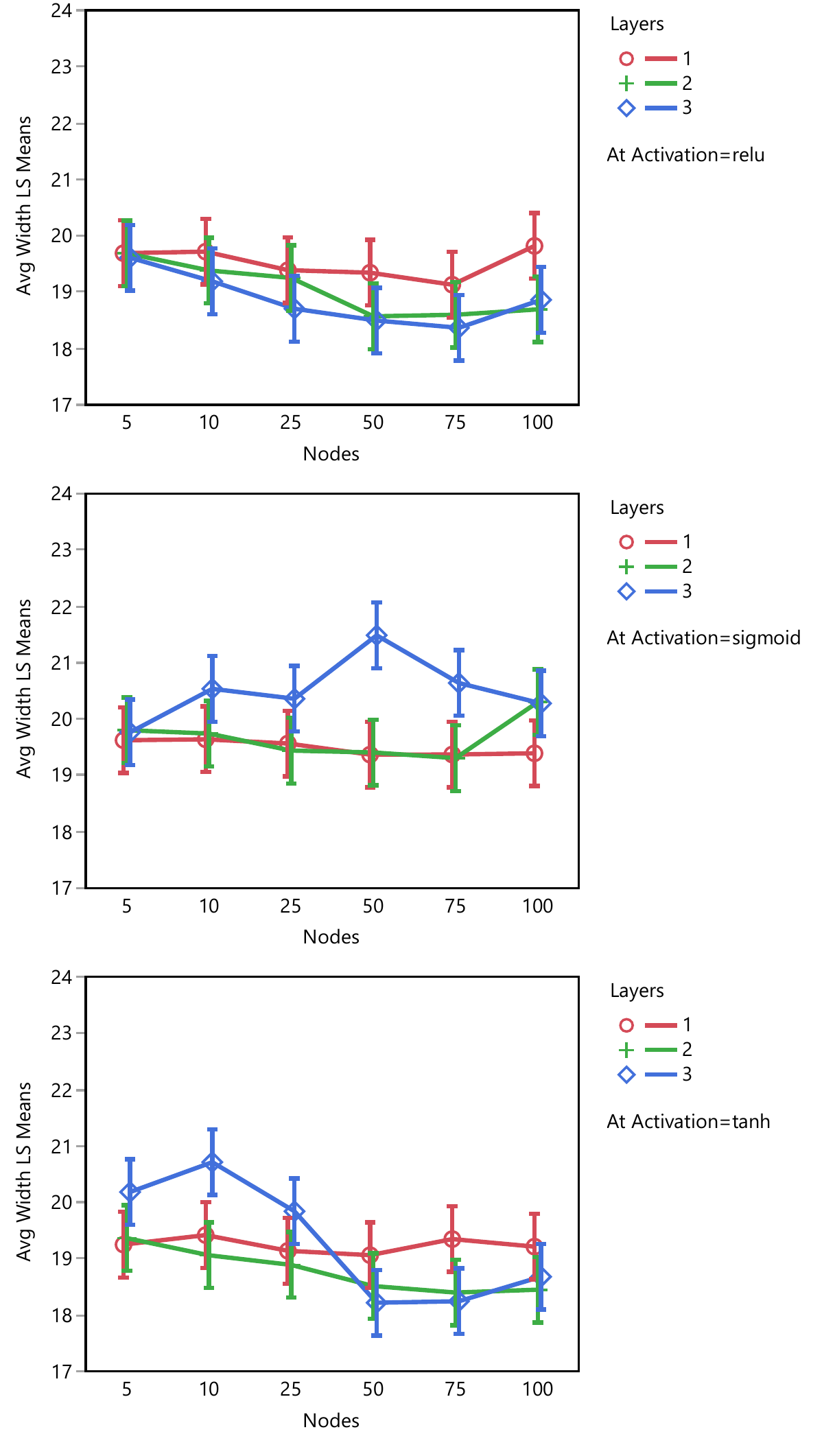}} \\
  \hline
\end{tabular}
\caption{PI Average Width in the Protein Structure Data Set.}
\label{figure:ProteinStructure_PIAvgW}
\end{center}
\end{figure}

\begin{figure}[H]
\begin{center}
\begin{tabular}{|c|c|}
  \hline \textbf{Pivot Bootstrap ($\bm{1,000}$ Resamples) }& \textbf{Split Conformal Inference} \\
  \hline \resizebox{2.8in}{!}{\includegraphics{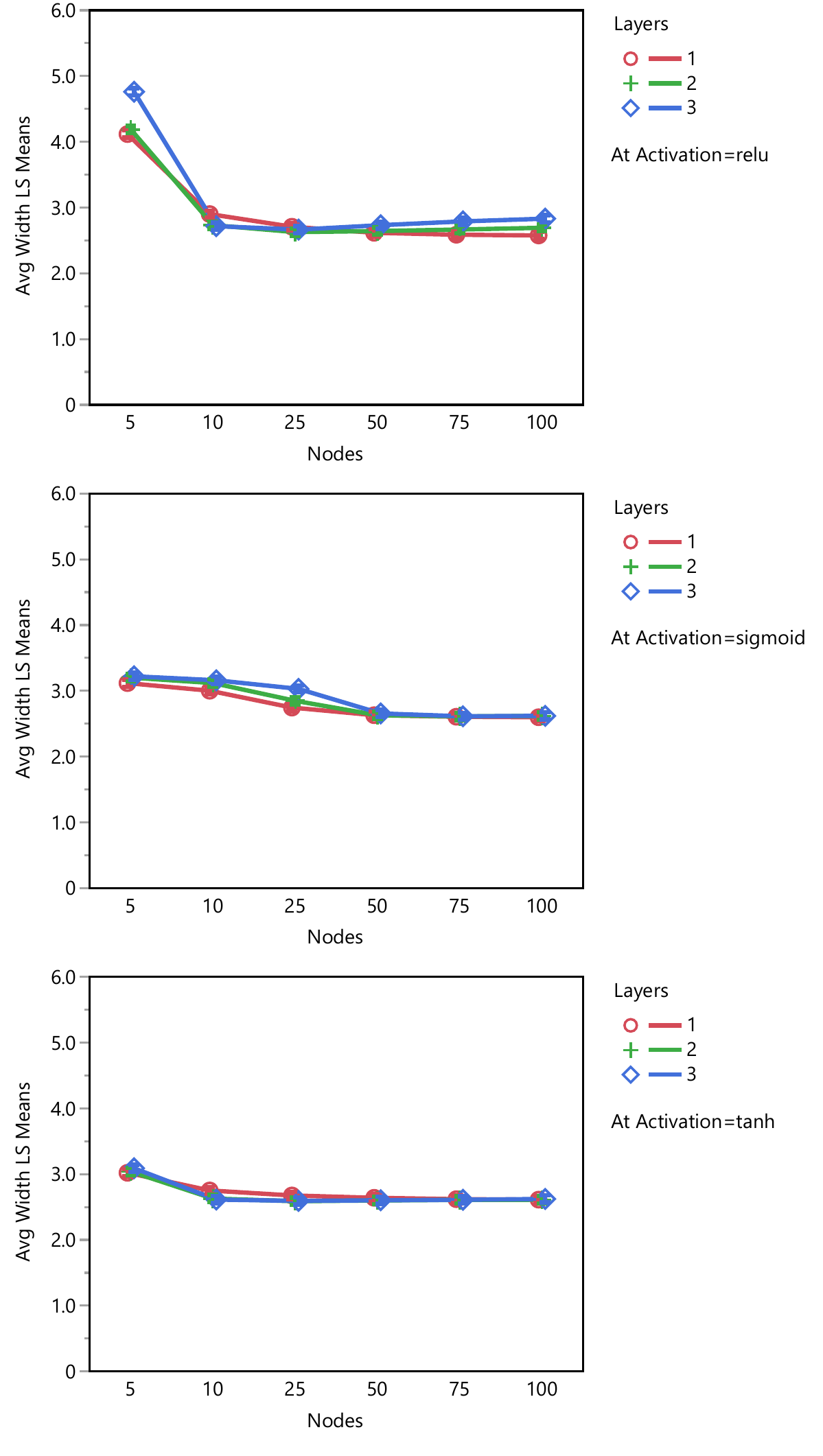}} & %
  \resizebox{2.8in}{!}{\includegraphics{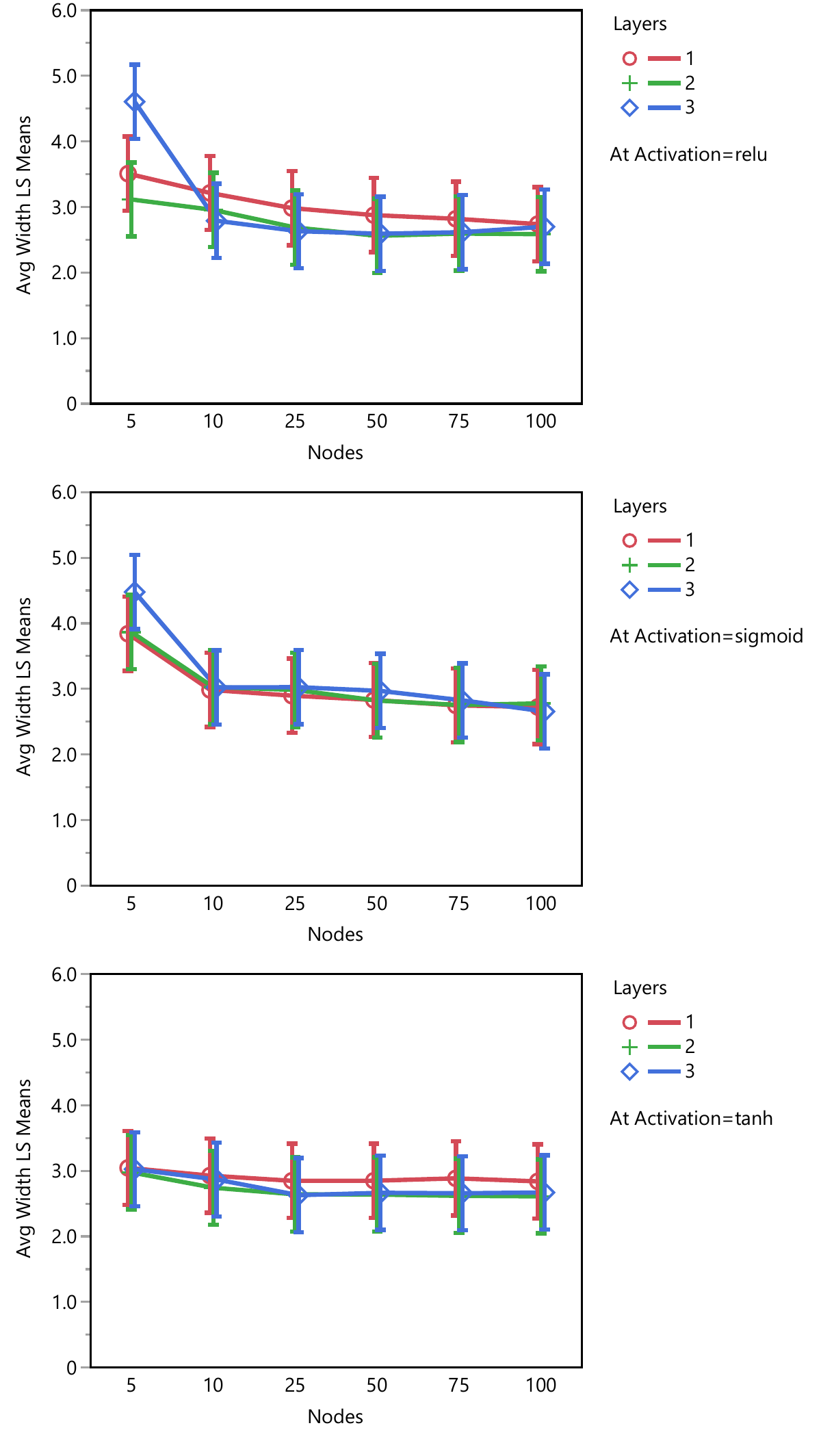}} \\
  \hline
\end{tabular}
\caption{PI Average Width in the Wine Quality Data Set.}
\label{figure:WineQuality_PIAvgW}
\end{center}
\end{figure}

\begin{figure}[H]
\begin{center}
\begin{tabular}{|c|c|}
  \hline \textbf{Pivot Bootstrap ($\bm{1,000}$ Resamples) }& \textbf{Split Conformal Inference} \\
  \hline \resizebox{2.8in}{!}{\includegraphics{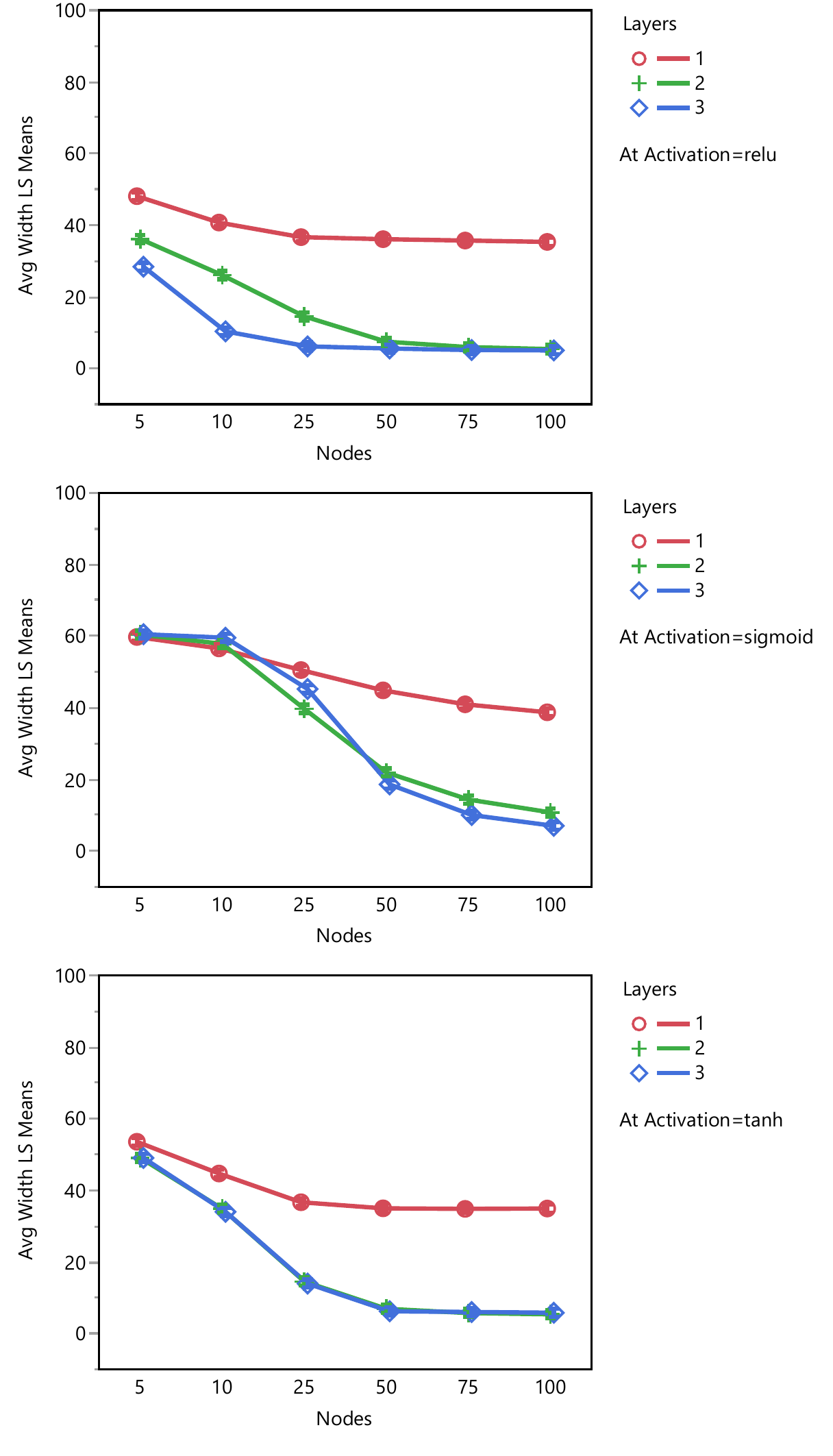}} & %
  \resizebox{2.8in}{!}{\includegraphics{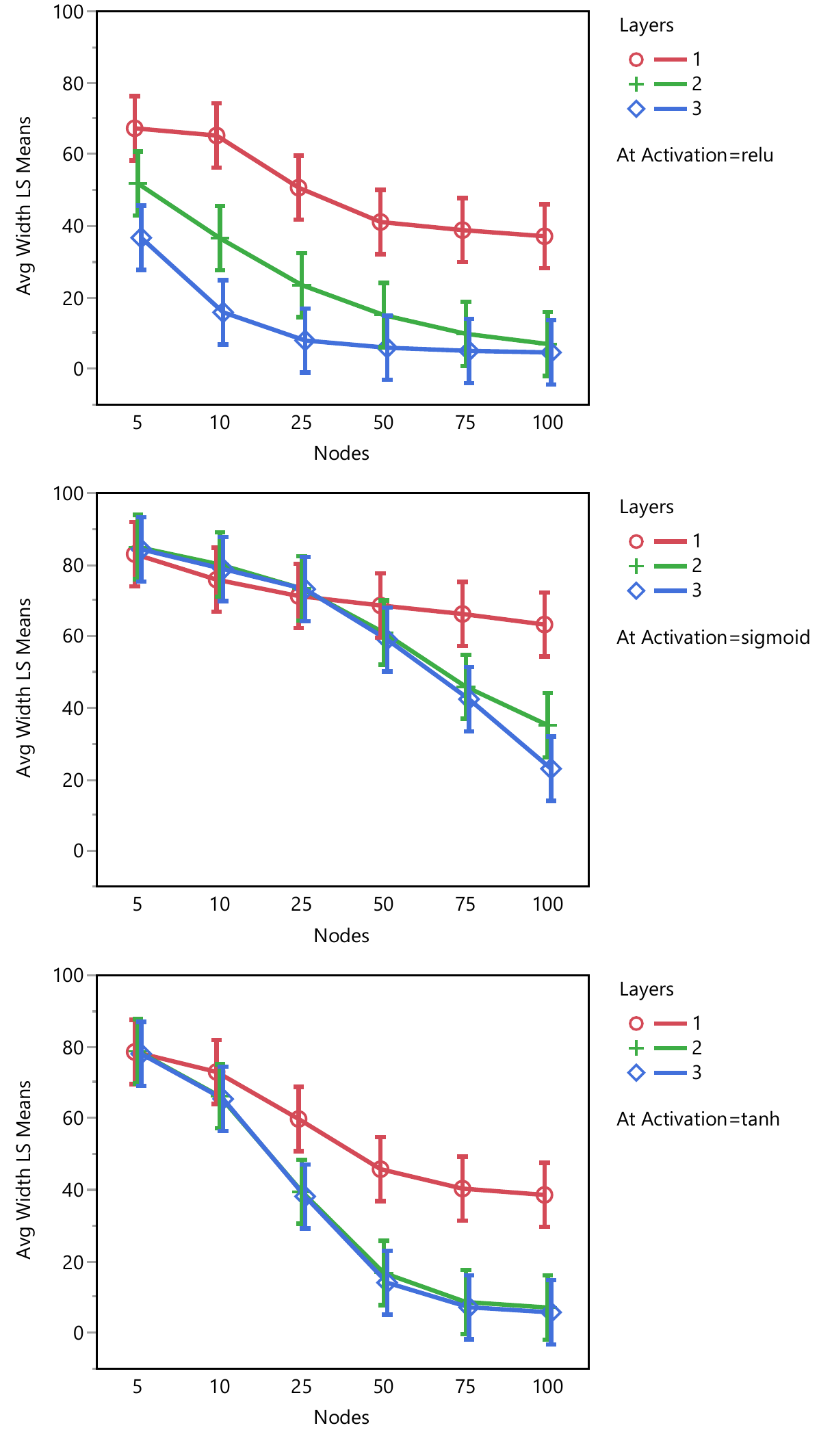}} \\
  \hline
\end{tabular}
\caption{PI Average Width in the Yacht Hydrodynamics Data Set.}
\label{figure:YachtHydro_PIAvgW}
\end{center}
\end{figure}

\begin{figure}[H]
\begin{center}
\begin{tabular}{|c|c|}
  \hline \textbf{Pivot Bootstrap ($\bm{1,000}$ Resamples) }& \textbf{Split Conformal Inference} \\
  \hline \resizebox{2.8in}{!}{\includegraphics{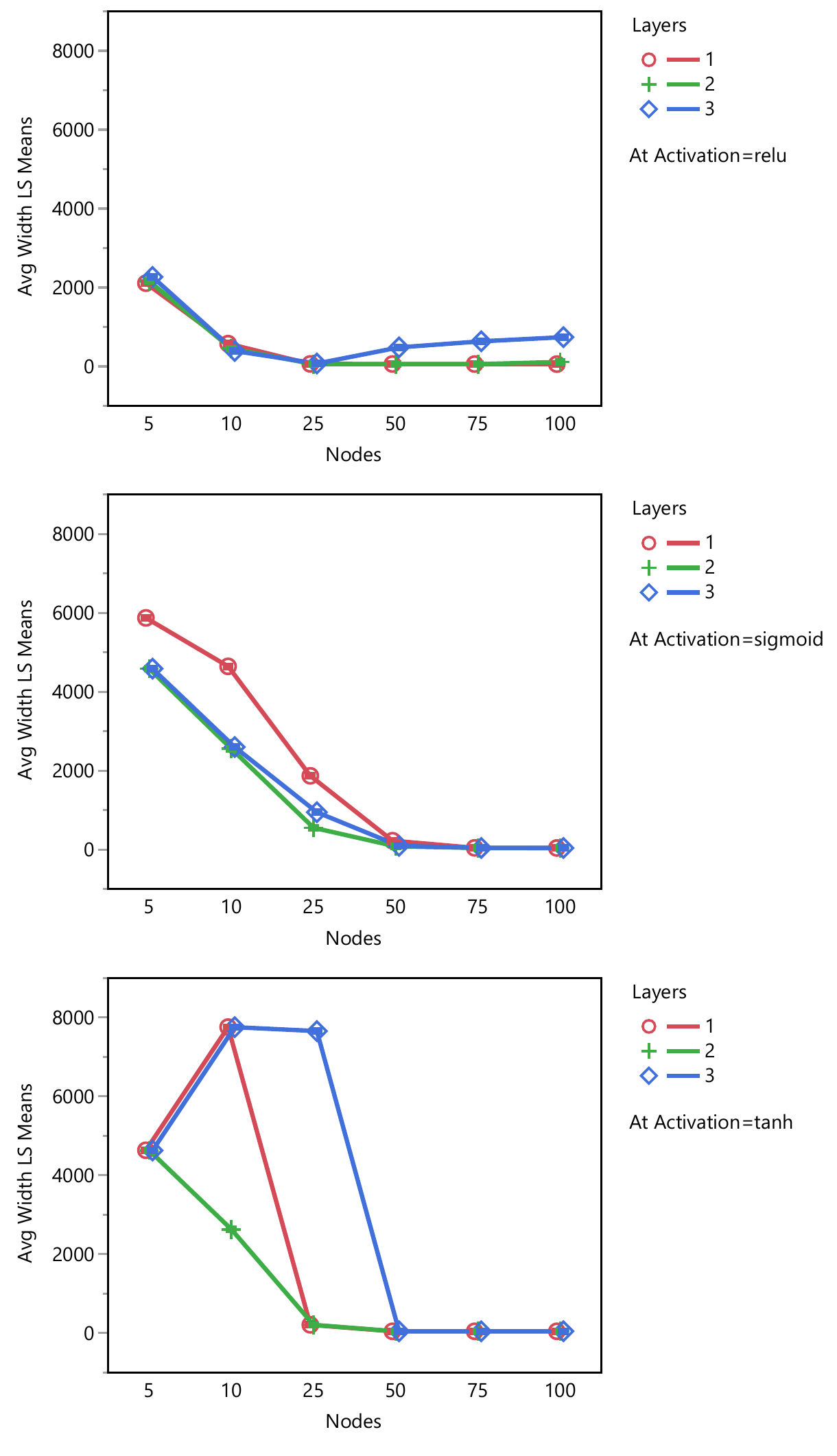}} & %
  \resizebox{2.8in}{!}{\includegraphics{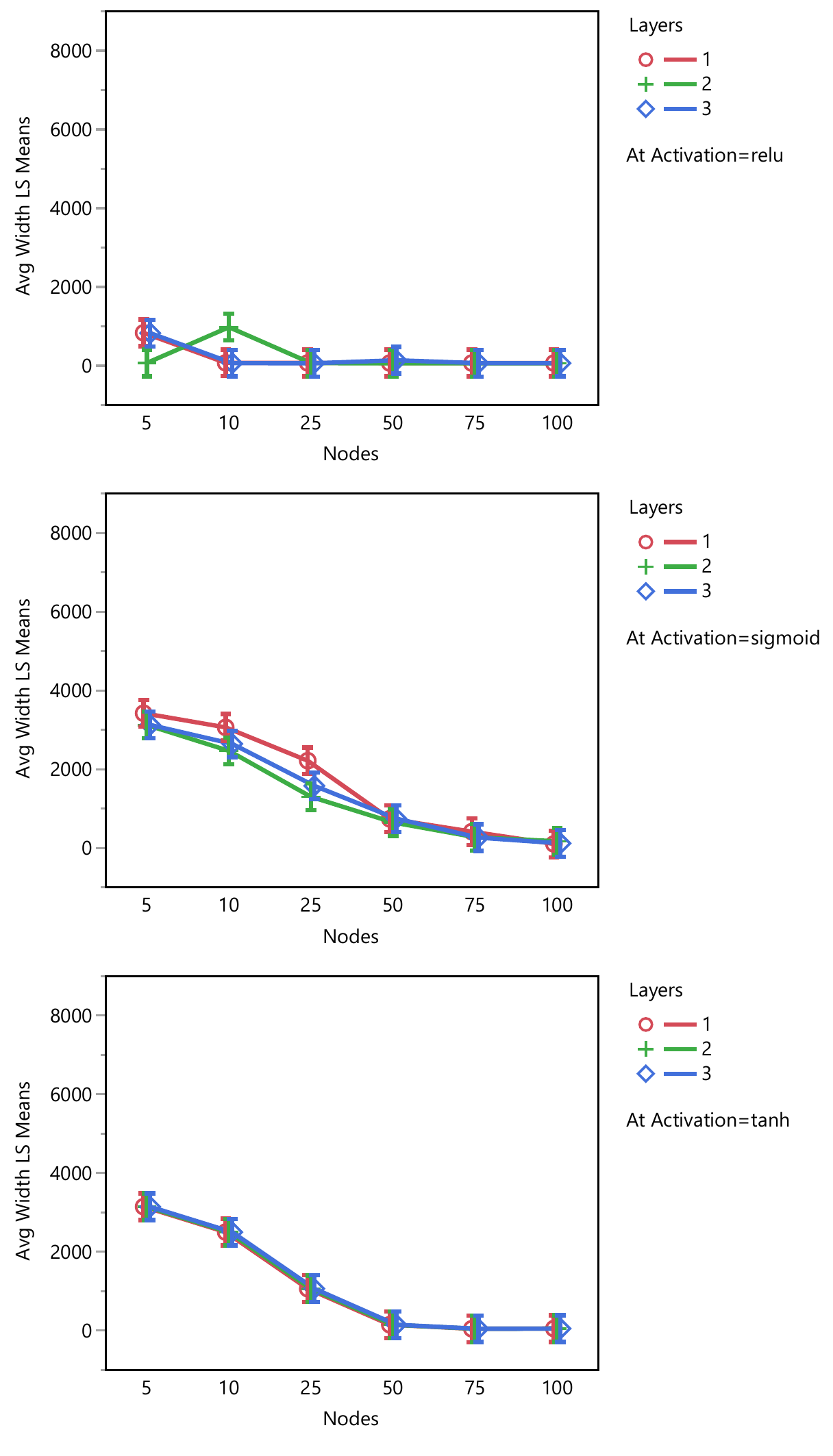}} \\
  \hline
\end{tabular}
\caption{PI Average Width in the Year Prediction Data Set.}
\label{figure:YearPrediction_PIAvgW}
\end{center}
\end{figure}

\section{Bootstrapped PI Coverage vs RMSE Plots} \label{appendix_c}

\begin{figure}[H]
 \centering
 \resizebox{6in}{!}{\includegraphics{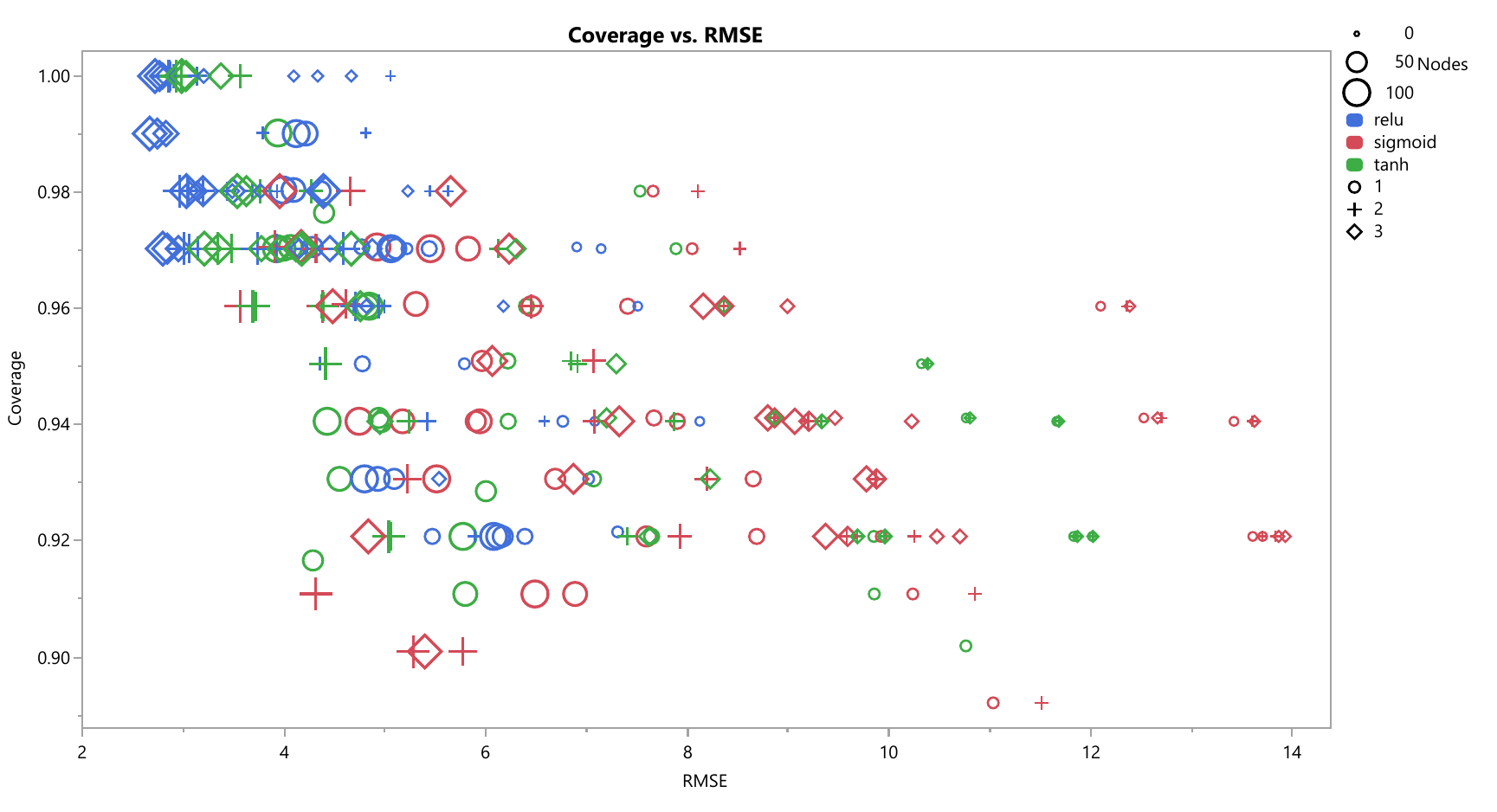}}
 \caption{PI Coverage versus RMSE for the Boston Housing Data Set.}
 \label{figure:BostonHousing_Cov_vs_RMSE}
\end{figure}

\begin{figure}[H]
 \centering
 \resizebox{6in}{!}{\includegraphics{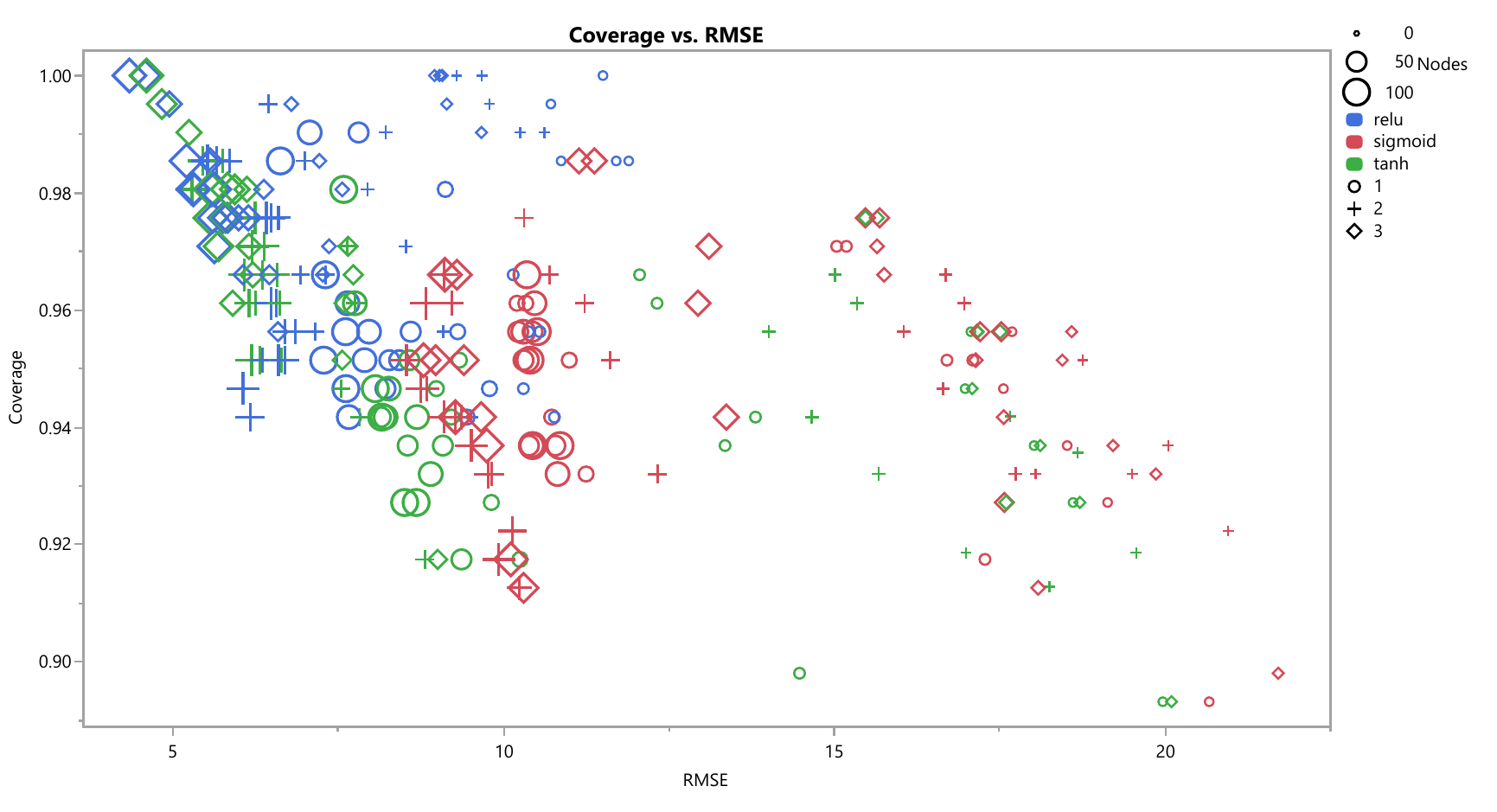}}
 \caption{PI Coverage versus RMSE for the Concrete Strength Data Set.}
 \label{figure:ConcreteStrength_Cov_vs_RMSE}
\end{figure}

\begin{figure}[H]
 \centering
 \resizebox{6in}{!}{\includegraphics{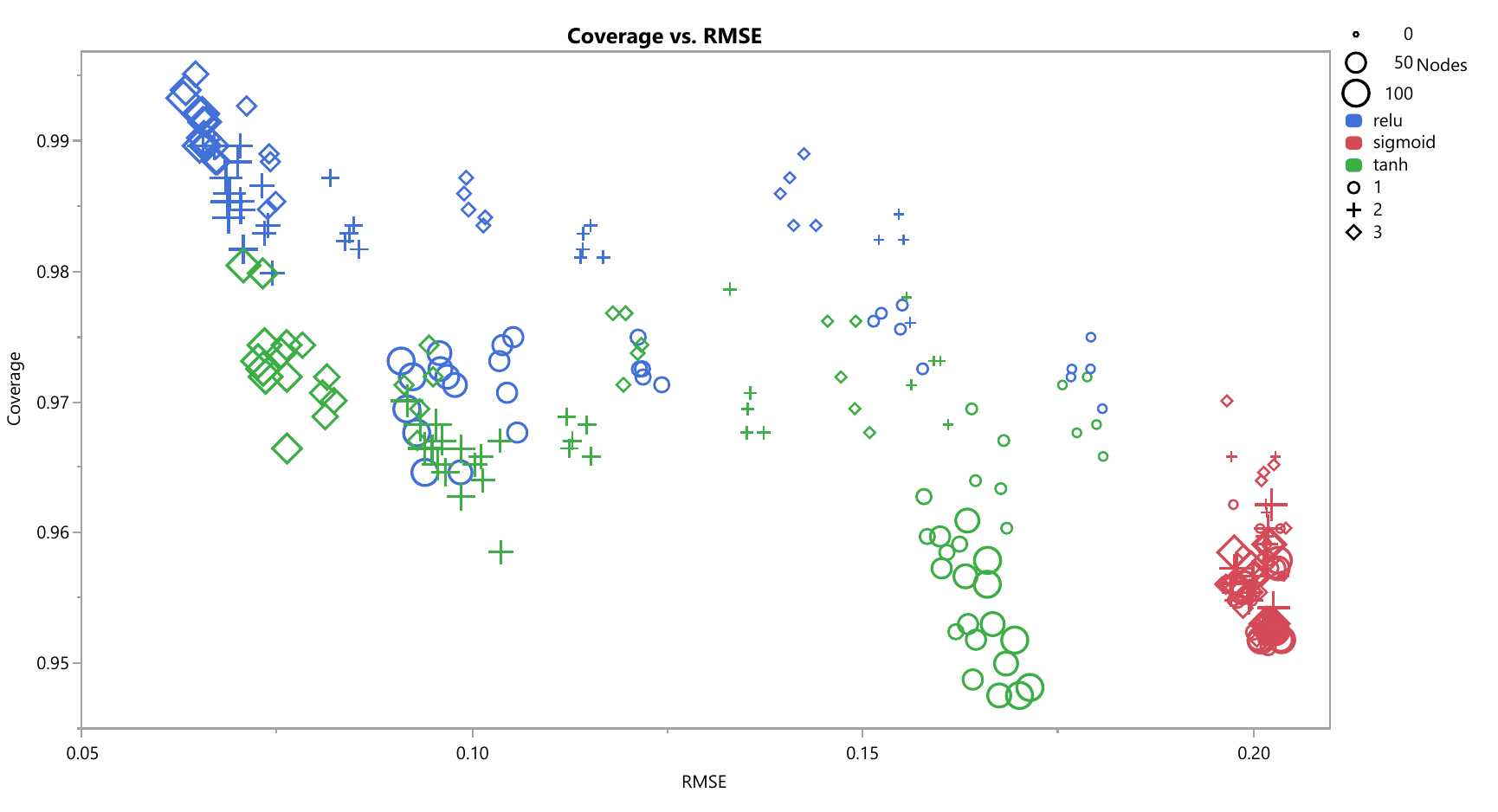}}
 \caption{PI Coverage versus RMSE for the Kinematics Data Set.}
 \label{figure:Kinematics_Cov_vs_RMSE}
\end{figure}

\begin{figure}[H]
 \centering
 \resizebox{6in}{!}{\includegraphics{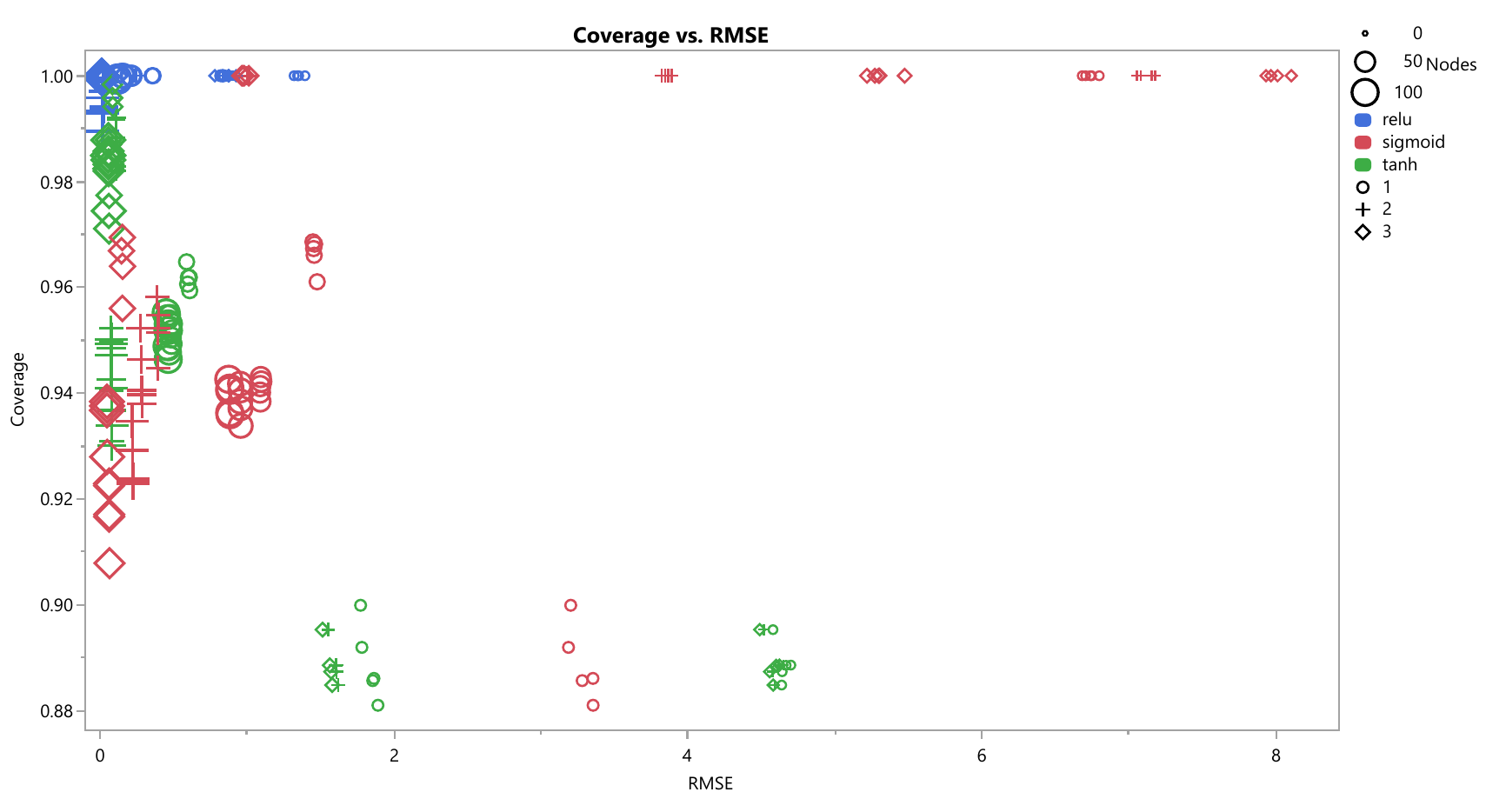}}
 \caption{PI Coverage versus RMSE for the Naval Propulsion Data Set.}
 \label{figure:NavalPropulsion_Cov_vs_RMSE}
\end{figure}

\begin{figure}[H]
 \centering
 \resizebox{6in}{!}{\includegraphics{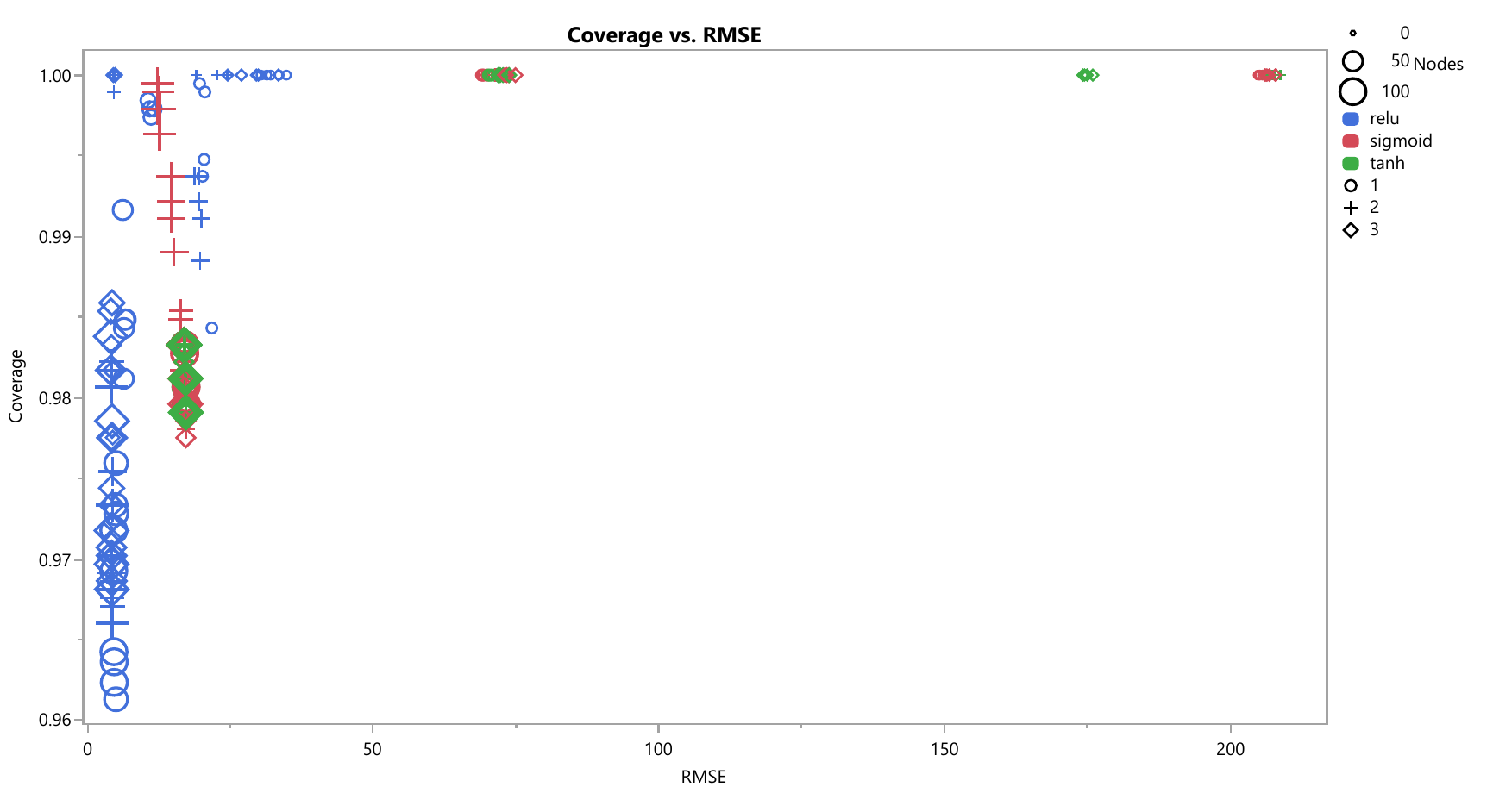}}
 \caption{PI Coverage versus RMSE for the Power Plant Data Set.}
 \label{figure:PowerPlant_Cov_vs_RMSE}
\end{figure}

\begin{figure}[H]
 \centering
 \resizebox{6in}{!}{\includegraphics{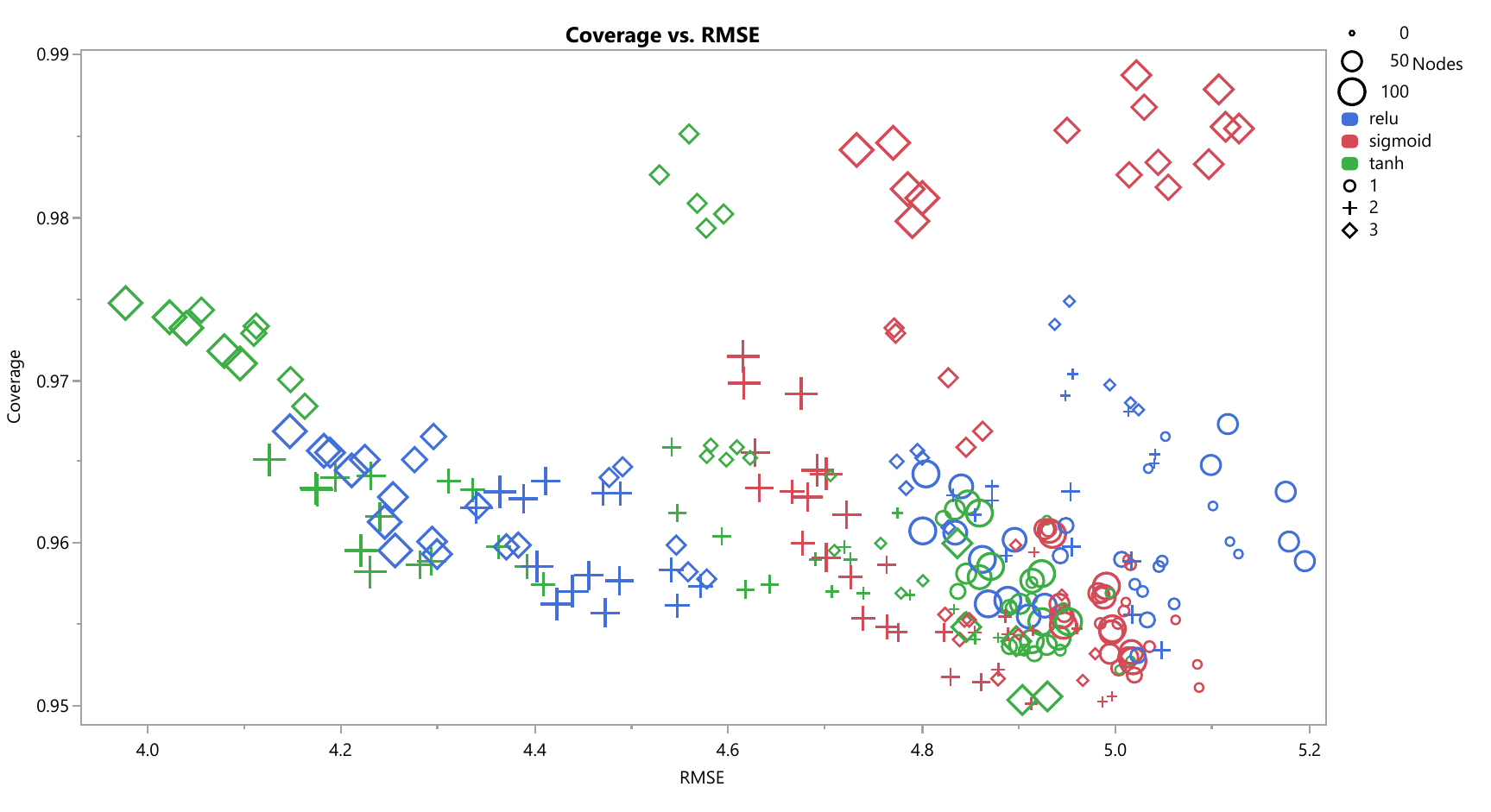}}
 \caption{PI Coverage versus RMSE for the Protein Structure Data Set.}
 \label{figure:ProteinStructure_Cov_vs_RMSE}
\end{figure}

\begin{figure}[H]
 \centering
 \resizebox{6in}{!}{\includegraphics{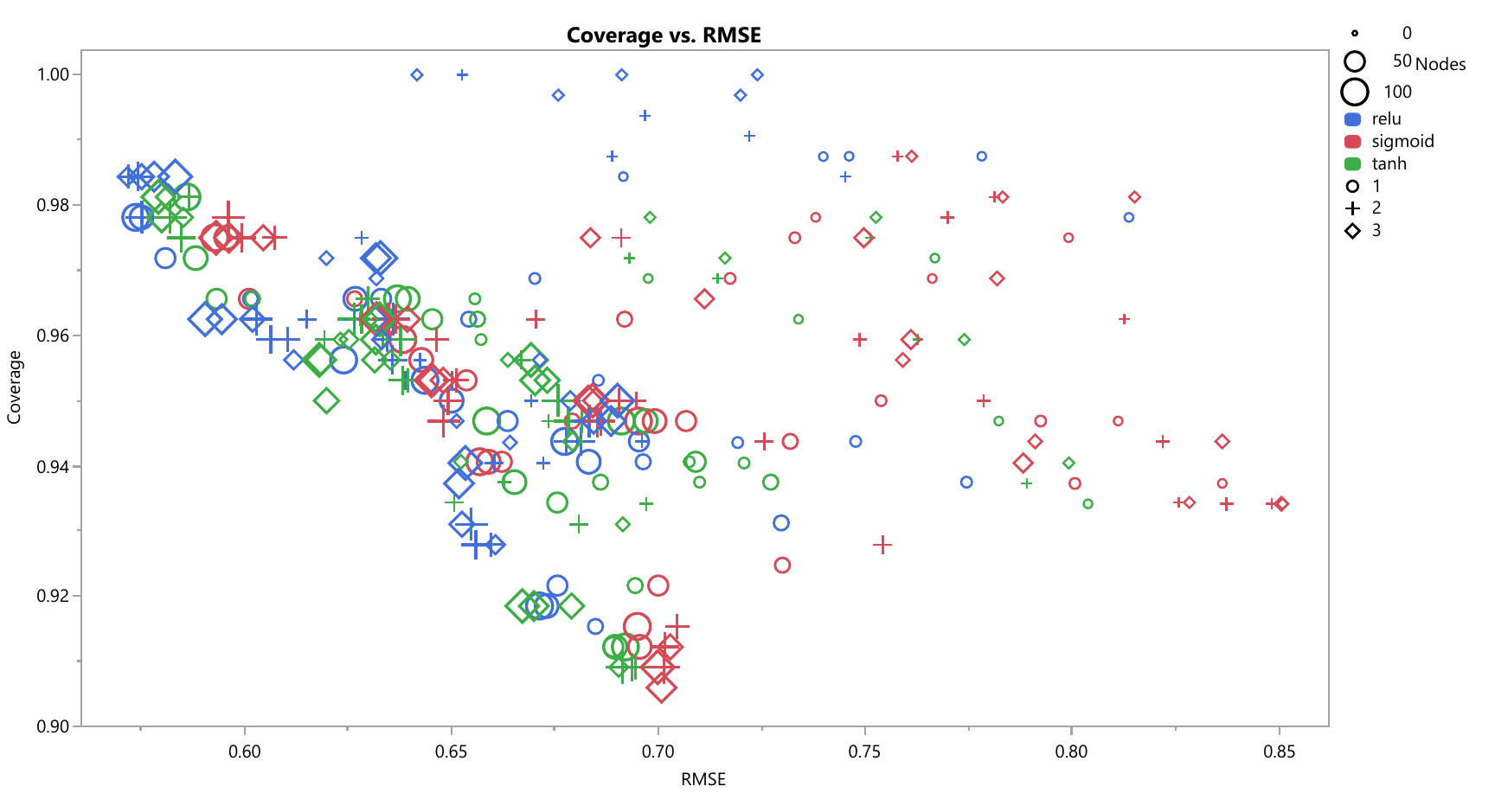}}
 \caption{PI Coverage versus RMSE for the Wine Quality Data Set.}
 \label{figure:WineQuality_Cov_vs_RMSE}
\end{figure}

\begin{figure}[H]
 \centering
 \resizebox{6in}{!}{\includegraphics{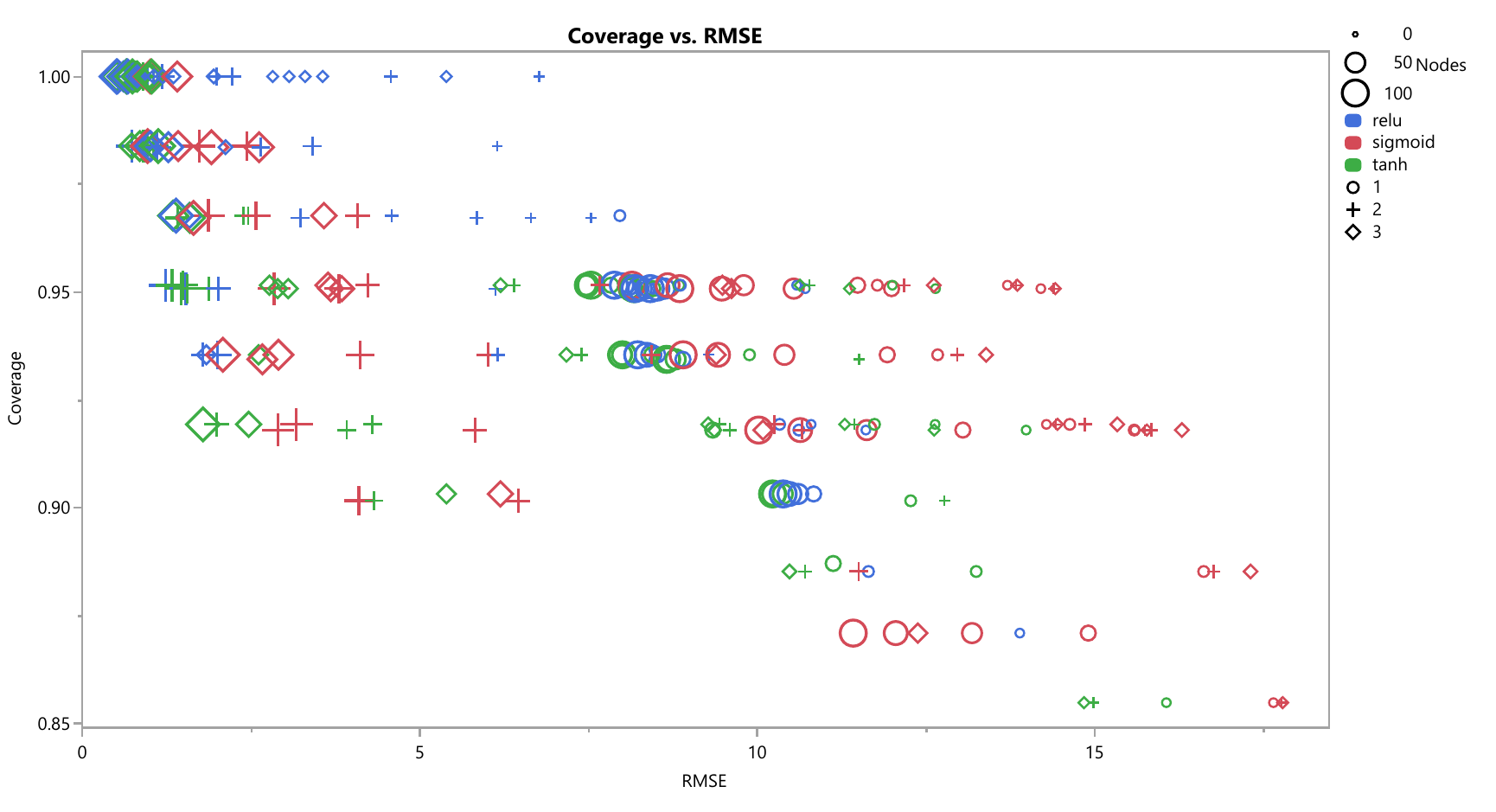}}
 \caption{PI Coverage versus RMSE for the Yacht Hydrodynamics Data Set.}
 \label{figure:YachtHydro_Cov_vs_RMSE}
\end{figure}

\begin{figure}[H]
 \centering
 \resizebox{6in}{!}{\includegraphics{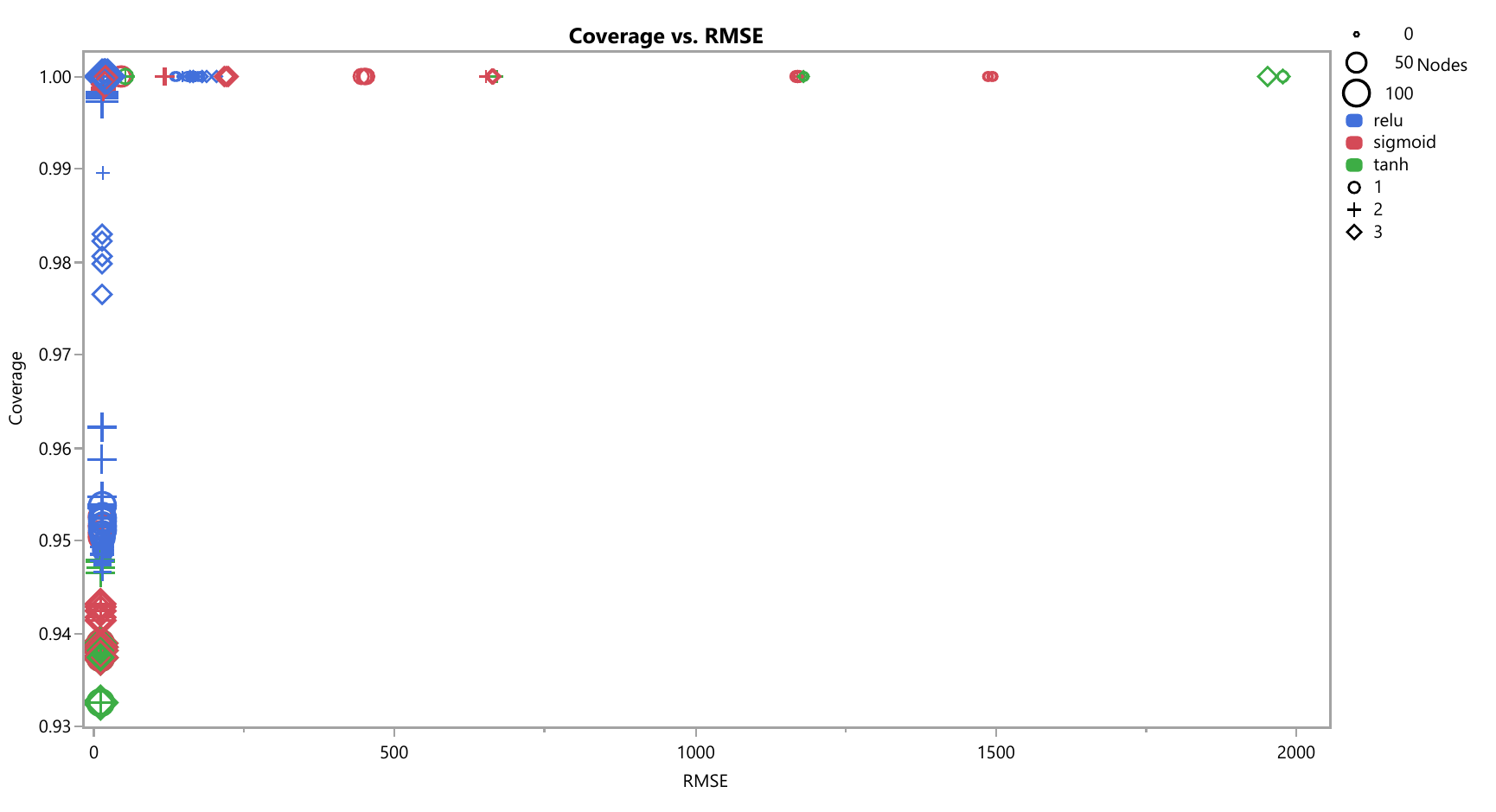}}
 \caption{PI Coverage versus RMSE for the Year Prediction Data Set.}
 \label{figure:YearPrediction_Cov_vs_RMSE}
\end{figure}

\begin{figure}[H]
 \centering
 \resizebox{6in}{!}{\includegraphics{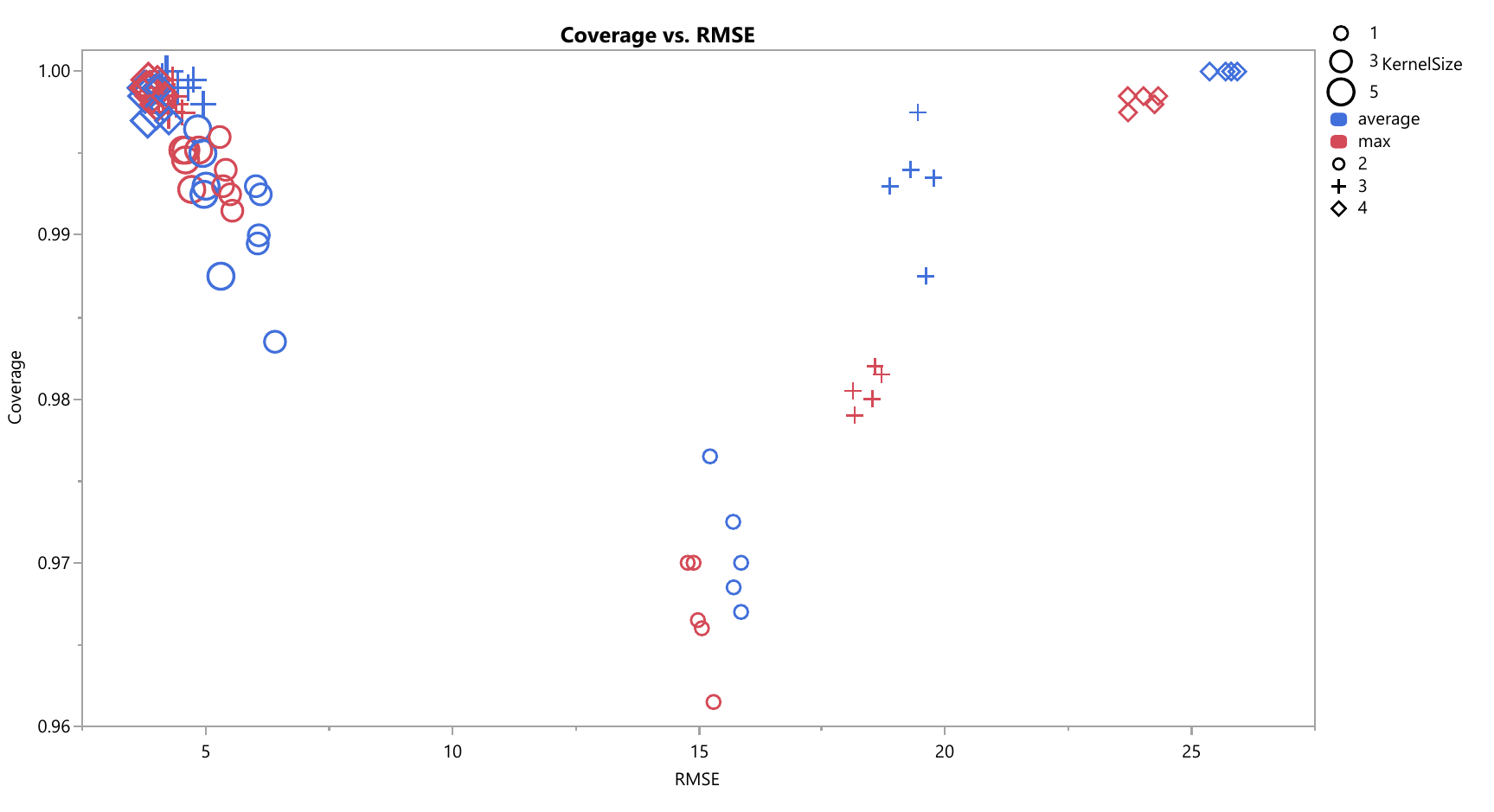}}
 \caption{PI Coverage versus RMSE for the RotNIST Data Set.}
 \label{figure:RotNIST_Cov_vs_RMSE}
\end{figure}

\end{document}